\documentclass{article}

\usepackage{microtype}
\usepackage{graphicx}
\usepackage{booktabs} 

\usepackage{hyperref}

\usepackage[accepted]{icml2023}

\usepackage{amsmath}
\usepackage{amssymb}
\usepackage{mathtools}
\usepackage{amsthm}

\usepackage{multirow}
\usepackage{subcaption}

\usepackage[export]{adjustbox}

\usepackage{tabularx}

\usepackage[capitalize,noabbrev]{cleveref}

\theoremstyle{plain}
\newtheorem{theorem}{Theorem}[section]

\newtheorem{lemma}[theorem]{Lemma}
\newtheorem{corollary}[theorem]{Corollary}
\theoremstyle{definition}

\theoremstyle{remark}

\icmltitlerunning{Synthetic Data for Model Selection}

\usepackage{xspace}
\makeatletter
\DeclareRobustCommand\onedot{\futurelet\@let@token\@onedot}
\def\@onedot{\ifx\@let@token.\else.\null\fi\xspace}
\def\eg{\emph{e.g}\onedot} 
\def\ie{\emph{i.e}\onedot}

\def\wrt{w.r.t\onedot}

\begin{document}

\twocolumn[
\icmltitle{Synthetic Data for Model Selection}




\begin{icmlauthorlist}
\icmlauthor{Alon Shoshan}{comp}
\icmlauthor{Nadav Bhonker}{comp}
\icmlauthor{Igor Kviatkovsky}{comp}
\icmlauthor{Matan Fintz}{comp}
\icmlauthor{G\'{e}rard Medioni}{comp}

\end{icmlauthorlist}

\icmlaffiliation{comp}{Amazon}

\icmlcorrespondingauthor{Alon Shoshan}{alonshos@amazon.com}

\icmlkeywords{Machine Learning, ICML}

\vskip 0.3in
]



\printAffiliationsAndNotice{}  

\begin{abstract}
Recent breakthroughs in synthetic data generation approaches made it possible to produce highly photorealistic images which are hardly distinguishable from real ones.
Furthermore, synthetic generation pipelines have the potential to generate an unlimited number of images.
The combination of high photorealism and scale turn synthetic data into a promising candidate for improving various machine learning (ML) pipelines.
Thus far, a large body of research in this field has focused on using synthetic images for training, by augmenting and enlarging training data.
In contrast to using synthetic data for training, in this work we explore whether synthetic data can be beneficial for model selection.
Considering the task of image classification, we demonstrate that when data is scarce, synthetic data can be used to replace the held out validation set, thus allowing to train on a larger dataset.
We also introduce a novel method to calibrate the synthetic error estimation to fit that of the real domain.
We show that such calibration significantly improves the usefulness of synthetic data for model selection.

\end{abstract}


\section{Introduction}
\label{sec:introduction}
Traditionally, in supervised ML pipelines, the data used to train a model is divided into two sets: the \emph{training set} and the \emph{validation set}.
The former is used to train various models, while the latter is used for ranking and selecting the best performing one, \ie, the best architecture and hyper-parameters.
Eventually, a final model with the selected configuration is trained on the entire data, including the training and the validation sets.
Test data is inaccessible to the training pipeline, especially for model selection.
The training-validation split provides means to estimate the models' error rate, which can be used for ranking.
However, it is not helpful for selection of models that were trained on the entire data.
As the optimal hyper-parameters depend on the number of training data samples and since models sharing the exact same hyper-parameters may exhibit large variance in accuracy, this may eventually lead to selecting a sub-optimal model.


In this work we propose to substitute the held out validation set with synthetic data, allowing for model selection even when training on the entire dataset.
The synthetic validation set is created using generative models trained on the immediately available data, and without reliance on external knowledge or tools (\eg, additional data sources, 3D rendering engines).
This makes our approach self-sufficient and applicable to a wide range of problems.   

    
Recent advances in the quality of synthetic data generation pipelines~\citep{Karras2019stylegan2,karras2020training,Karras2021,peng2018visda, dhariwal2021diffusion, saharia2022photorealistic} have reduced the synthetic-to-real domain gap enough to successfully utilize the generated data for training deep models~\citep{besnier2020dataset}. 
Other works have focused on analyzing and quantifying various characteristics of the domain gap~\citep{sajjadi2018assessing,kynkaanniemi2019improved}.
That said, to the best of our knowledge, the specific task of model selection with synthetic data was not addressed. 
When using synthetic data for training a model, one's goal is to minimize the generalization gap \wrt the real domain~\citep{ben2010theory}. 
Solving for generalization in presence of a synthetic-to-real domain gap is challenging. 
However, for model selection, one's goal is to use synthetic data for ranking a set of trained models, while requiring rank preservation in the real domain.
In this work, we introduce a sufficient condition for cross-domain rank preservation and empirically validate its value for model selection.
We perform extensive experiments on the CIFAR10~\citep{krizhevsky2009learning} dataset showing that indeed we are able to improve overall accuracy by selecting better models using synthetic data.
Furthermore, we introduce a novel calibration approach to improve the ranking capabilities of synthetic data on rich visual domains such as ImageNet~\citep{imagenet_cvpr09}.

We summarize our contributions as following:
\begin{itemize}
    \item We show that the error rank of models evaluated on synthetic data mostly preserves the rank of their performance with respect to a held out test set.
    \item We demonstrate the value of this observation by allowing to train on the entire dataset and to select the best model using synthetic data rather than a held-out validation set.
    We show that this method, on average, selects a better model.
    \item We introduce a novel calibration approach to improve the ranking capabilities of synthetic datasets by re-weighing its samples.
    Such re-weighting enables a high level of ranking in challenging visual domains, even when high-quality synthetic generators are not available.
    \item We provide a sufficient condition for which the rank preservation will be maintained across domains.


\end{itemize}





\section{Related Work}
\label{sec:related_work}
Model selection is a challenging task traditionally done by comparing the model estimation errors via cross-validation.
This venture is expensive as it requires to train each model a number of times.
Multiple methods exist to improve the efficiency of cross-validation ~\citep{liu2018fast, ghosh2020approximate, wilson2020approximate}.
These are limited to specific data types or models and were not demonstrated on complex tasks such as image classification.
A common alternative approach is to use a single train-validation split.
In order to leverage all the available data, both approaches require to retrain the selected model again on the entire dataset and do not allow to select a specific trained model.
Furthermore, additional training data may change the optimal model hyper-parameters.
Other methods exist in which a validation set is not required altogether.
To avoid using a held out validation set,~\citet{corneanu2020computing} employed persistent homology to estimate the performance gap between training and testing error without a test dataset.
While this method can generalize well across datasets, it does not allow to compare classifiers of different architectures.
In~\citet{neyshabur2017exploring} different measures based on norms of weight matrices are proposed for quantifying and guaranteeing generalization in deep models.
~\citet{li2020leave} uses the training set with augmentations to be used instead of a validation set.
This method was only demonstrated on simple classification datasets using simple non deep learning classifiers.


In the recent years generative models have improved significantly and are at the point where state-of-the-art generative models, such as GANs~\citep{Karras2019stylegan2, karras2019style,karras2017progressive,karras2021alias,esser2021taming} and diffusion models~\citep{dhariwal2021diffusion,saharia2022photorealistic, ho2020denoising,song2020score,rombach2022high}, are able to produce high-resolution synthetic images that are indistinguishable from real images.  
The potential of producing unlimited amount of training data using generative models has prompted the exploration of using synthesized data to augment downstream tasks.

Although promising, the use of synthetic data for training is not trivial.
\citet{ravuri2019classification} show that training a classifier using only synthetic data results in sub-optimal accuracy in both CIFAR10 and ImageNet datasets.
\citet{eilertsen2021ensembles} address the problem of training with synthetic data by using an ensemble of GANs rather than a single one. 



\section{Synthetic Data for Model Selection}
\label{sec:synthetic_data_for_evaluation}
\begin{figure*}[t]
    \centering
    \small
    \begin{tabular}{c c c}
        Train50K & Train30K & Train10K \\
        \includegraphics[width=0.3\linewidth]{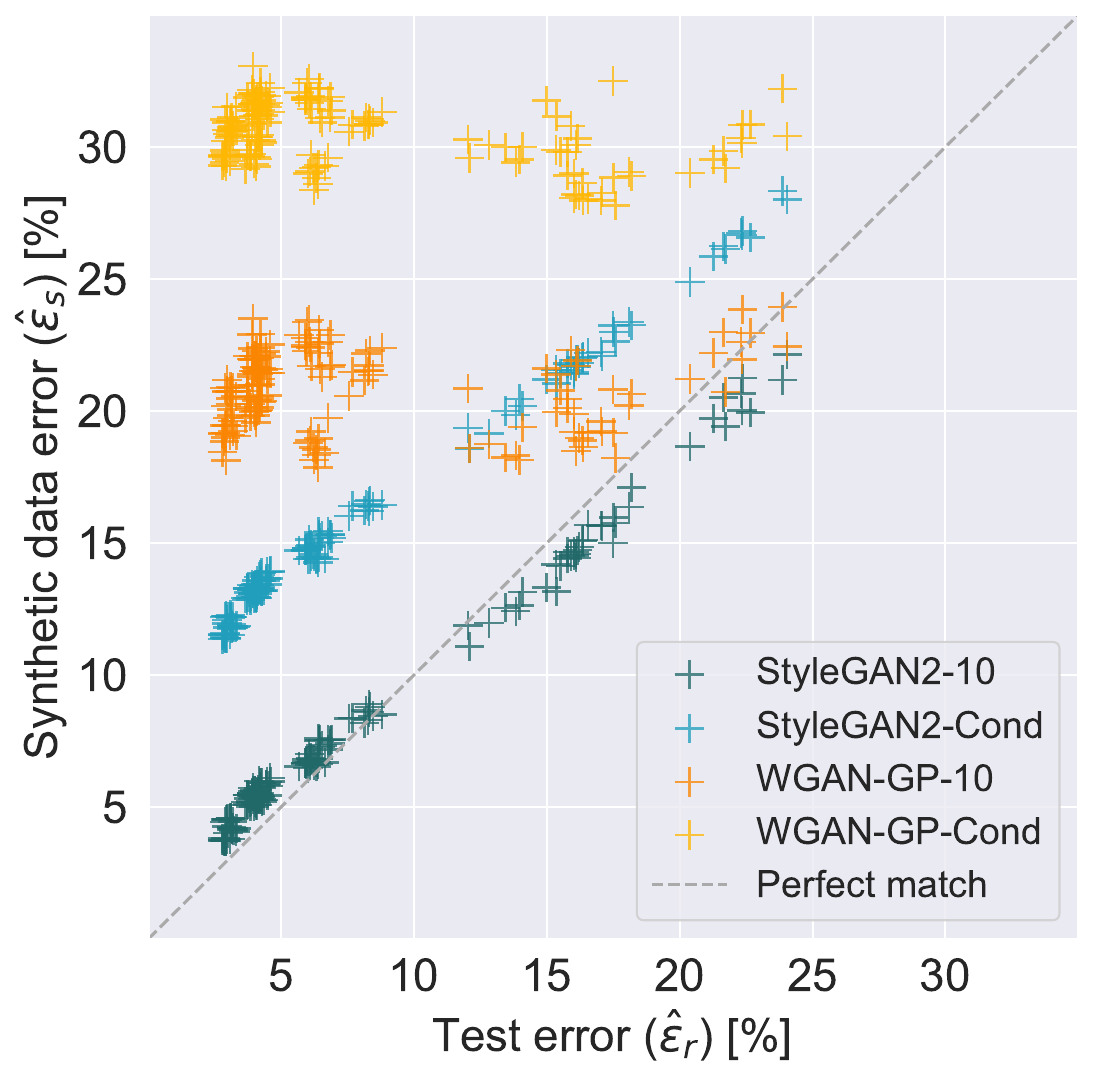} &
        \includegraphics[width=0.3\linewidth]{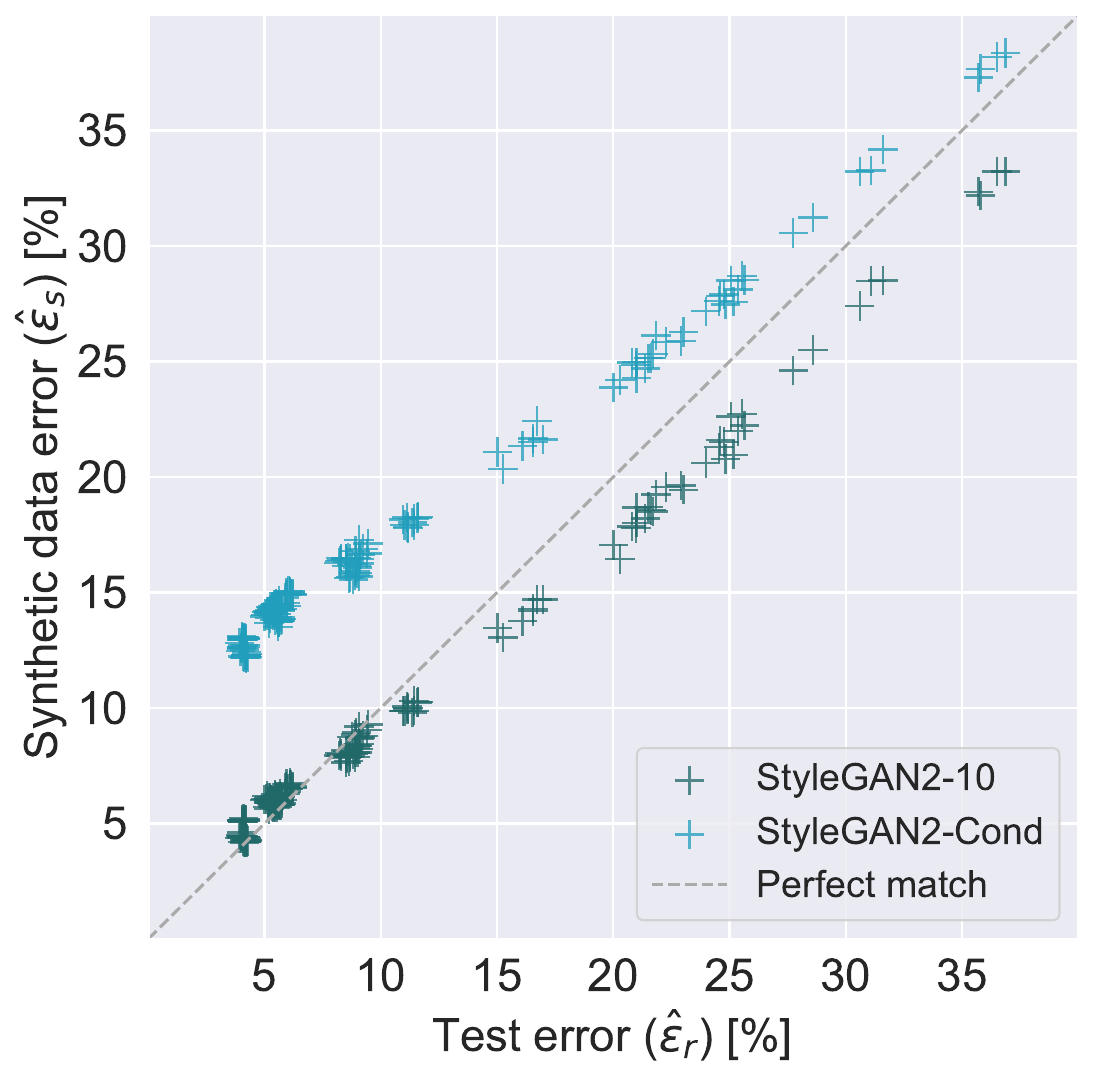} &
        \includegraphics[width=0.3\linewidth]{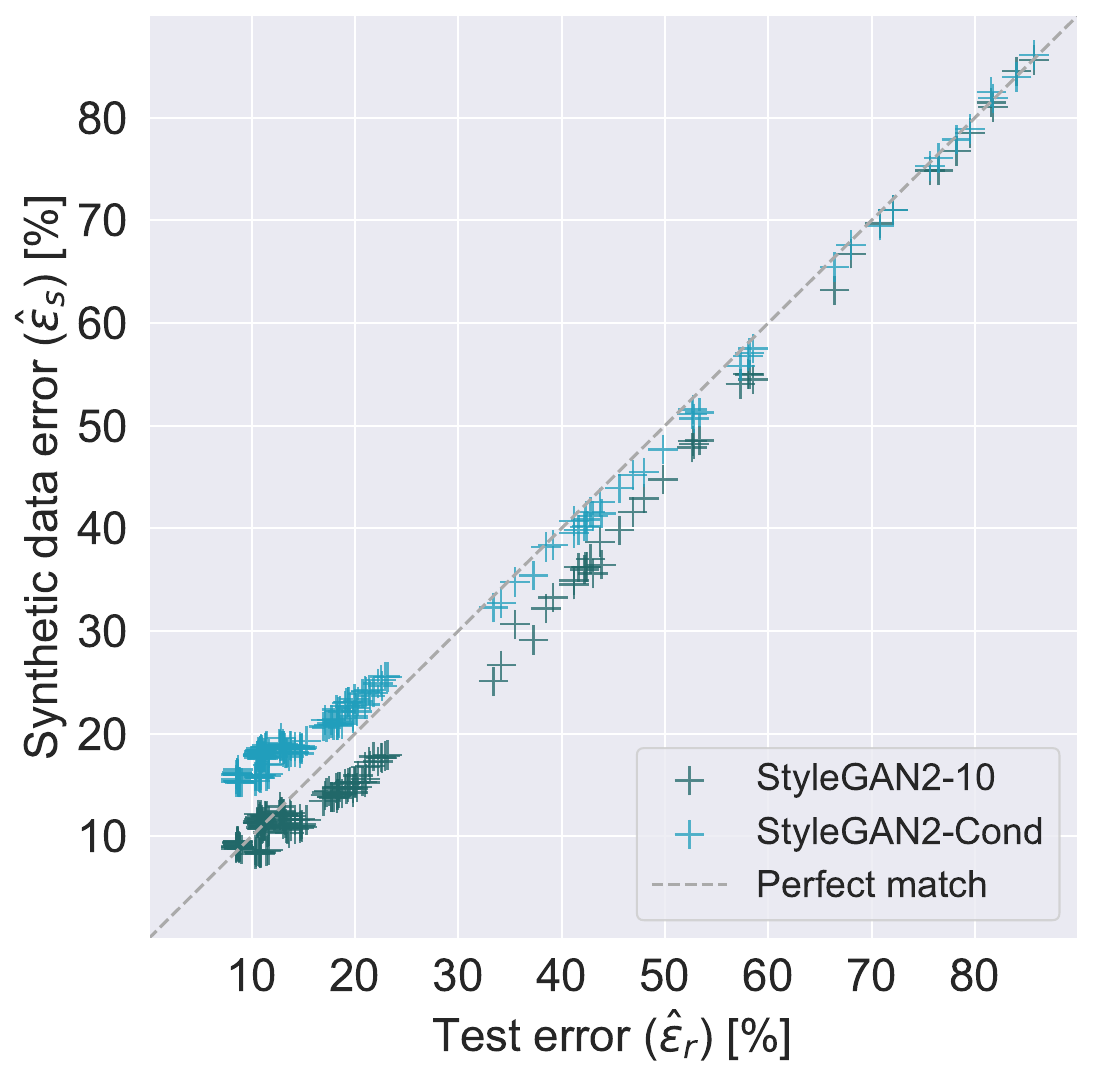} \\
    \end{tabular}  
\vspace{-0.08in} 
\caption{\textbf{Synthetic vs. test errors:} Each point represents the synthetic data error (y axis), $\hat{\epsilon}_s$, and test error (x axis), $\hat{\epsilon}_r$, of a single model. A total of 170 models were trained per CIFAR10 subset. 
Each model was evaluated by multiple synthetic datasets represented by different colors.}
\vspace{-0.18in}
\label{fig:standard_correletion_graph}
\end{figure*}
We follow notations similar to~\citet{ben2010theory}.
A domain is defined as a pair consisting of a distribution $\mathcal{D}=\langle\Omega,\mu \rangle$, where $\Omega$ is the sample domain and $\mu$ is the probability density function, and a labeling function $f: \Omega \rightarrow \mathcal{Y}$, where $\mathcal{Y}$ represents possible classes.
We consider a particular pair of domains, where one is the original real domain, denoted by $\langle \mathcal{D}_r,f_r \rangle$, and the second one is synthetic, denoted by $\langle \mathcal{D}_s,f_s \rangle$, specifically tuned to mimic the real one.
A hypothesis (model) is a function $h: \Omega \rightarrow \mathcal{Y}$. 
The risk 
or the probability that a hypothesis $h$ disagrees with a labeling function $f$, according to the distribution $\mathcal{D}_r$ 
is defined as:
$ \epsilon_r(h,f) = E_{\mathbf{x} \sim \mathcal{D}_r}[ h(\mathbf{x}) \ne f(\mathbf{x}) ]. $
We neglect the difference between $f_r$ and $f_s$ and use the shorthand notation, $\epsilon_r(h)=\epsilon_r(h,f)$.
Let $\Delta\epsilon$ denote the risk difference between two hypotheses, $h_1, h_2 \in \mathcal{H}$, measured over a probabilistic distribution $\mathcal{D}$, \emph{\ie}, $\Delta\epsilon=\epsilon(h_2)- \epsilon(h_1)$.

A common approach for model selection is the holdout method, where two datasets are sampled from $\langle \mathcal{D}_r,f_r \rangle$: the training set, $D_r^{trn}= \{\mathbf{x}_i, y_i \}_{i=1}^{N_{trn}}$ and the validation set, $D_r^{val}= \{\mathbf{x}_i, y_i \}_{i=1}^{N_{val}}$.
A model (hypothesis) is trained using empirical risk minimization on $D_r^{trn}$.
Thereafter, the model's risk is estimated using the validation set:
$\hat{\epsilon}_r(h) = \frac{1}{N_{val}} \sum_{i=1}^{N_{val}} I ( h(\mathbf{x}_i) \ne y_i )$.
This allows to compare different models with different hyperparameters and to select those that minimize $\hat{\epsilon}_r(h)$.
Other approaches such as cross-validation and bootstrap also exist~\citep{kohavi1995study}.
Since increasing the number of samples in the training set almost always increases the accuracy of the model, a common final step is to re-train the model, using the hyperparameters found in the previous step, on the entire dataset, $D_r^{trn} \cup D_r^{val}$.
However, without a held-out dataset, it is no longer possible to compare models.

We propose to replace this two-step approach with a single step where we train a model on the entire dataset, then rather than estimating $\hat{\epsilon}_r(h)$, we instead generate a new dataset $D_s= \{\mathbf{x}_i, y_i \}_{i=1}^{N_s}$ via a generative model, then we estimate the error $\hat{\epsilon}_s(h)$, where the domain $\langle D_s,f_s \rangle$ approximates the original domain $\langle D_r,f_r \rangle$.
Although it may often be impossible to guarantee highly accurate error estimation due to the synthetic-real domain gap, below we present a sufficient condition for hypotheses' error rank preservation across domains.

\begin{lemma}
\label{lemma_1}
Let $\Delta\epsilon$ denote the risk difference between two hypotheses, $h_1, h_2 \in \mathcal{H}$, measured over a probability distribution $\mathcal{D}=\langle\Omega, \mu\rangle$, \emph{\ie}, $\Delta\epsilon=\epsilon(h_2)- \epsilon(h_1)$. Let $f$ denote the labeling function. Let $\Omega_1=\left\{\mathbf{x} \in \Omega | h_1(\mathbf{x}) \neq f(\mathbf{x}) \land h_2(\mathbf{x}) = f(\mathbf{x}) \right\}$ and $\Omega_2=\left\{\mathbf{x} \in \Omega | h_2(\mathbf{x}) \neq f(\mathbf{x}) \land h_1(\mathbf{x}) = f(\mathbf{x}) \right\}$.
Then, 
\begin{align*}
\Delta\epsilon=\int_{\Omega_2} \mu(\mathbf{x}) d\mathbf{x} - \int_{\Omega_1} \mu(\mathbf{x}) d\mathbf{x}.
\end{align*}
\end{lemma}
Proof is provided in Appendix~\ref{sec:appendix_proof_of_lemma}.
Informally, Lemma~\ref{lemma_1} states that the error gap between two hypotheses depends only on the area where they disagree.

\begin{theorem}
\label{th:main}
Let $\Delta\epsilon_r$ and $\Delta\epsilon_s$ denote the risk difference between two hypotheses, $h_1, h_2 \in \mathcal{H}$, measured over the real and the synthetic probability distributions $\mathcal{D}_r=(\Omega, \mu_r)$ and $\mathcal{D}_s=(\Omega, \mu_s)$, respectively, \emph{\ie}, $\Delta\epsilon_r=\epsilon_r(h_2)- \epsilon_r(h_1)$ and $\Delta\epsilon_s=\epsilon_s(h_2)- \epsilon_s(h_1)$.
Let $f$ denote the labeling function. Then, for any $h_1, h_2 \in \mathcal{H}:$
\begin{align*}
\Delta \epsilon_s - \Delta \epsilon_r \leq \delta_{h_1 \oplus h_2}(\mu_r, \mu_s),
\end{align*}
where $\delta_{h_1 \oplus h_2}$ is the total variation computed over the subset of the domain $\Omega$, where the hypotheses $h_1$ and $h_2$ do not agree. 
\end{theorem}

\begin{proof}
\begin{align*}
\Delta \epsilon_s &- \Delta \epsilon_r = \int_{\Omega_2} \mu_s(\mathbf{x}) d\mathbf{x} -\int_{\Omega_1} \mu_s(\mathbf{x}) d\mathbf{x} \\
                  &~~~~~~~~~~~~ -\int_{\Omega_2} \mu_r(\mathbf{x}) d\mathbf{x} + \int_{\Omega_1} \mu_r(\mathbf{x}) d\mathbf{x} \\
                  &= \int_{\Omega_2} \mu_s(\mathbf{x}) - \mu_r(\mathbf{x}) d\mathbf{x} - \int_{\Omega_1} \mu_s(\mathbf{x}) - \mu_r(\mathbf{x}) d\mathbf{x} \\
                  &\leq \int_{\Omega_2} |\mu_s(\mathbf{x}) - \mu_r(\mathbf{x})| d\mathbf{x} + \int_{\Omega_1} |\mu_s(\mathbf{x}) - \mu_r(\mathbf{x})| d\mathbf{x} \\
                  &=
                  \int_{\Omega_1 \cup \Omega_2} |\mu_s(\mathbf{x}) - \mu_r(\mathbf{x})| d\mathbf{x} 
                  \leq \delta_{h_1 \oplus h_2}(\mu_r, \mu_s)
\end{align*}
\end{proof}
The last line follows from the fact that $\Omega_1 \cap \Omega_2=\emptyset$.
Theorem~\ref{th:main} provides a condition for error rank preservation between two hypotheses.
In order to reach a bound that is true for any two hypotheses, we upper bound it by the total variation.


\begin{corollary}
\label{cor:main}
Given the definitions above, let $\delta(\mu_r, \mu_s)$ denote the total variation between the two distributions, $\mathcal{D}_r, \mathcal{D}_s$. Then, 
\begin{align*}
\Delta\epsilon_s \geq \delta(\mu_r, \mu_s) \Rightarrow \Delta\epsilon_r \geq 0.
\end{align*}
\end{corollary}

Informally, Corollary~\ref{cor:main} indicates that if the total variation between the real and synthetic distributions is not larger than the synthetic risk difference between a pair of hypotheses, then their error ranking is preserved across  domains.

We note that the total variation bound is quite loose.
Theoretically, a tighter bound can be achieved following an approach presented in \citet{ben2010theory}.
We present this connection in Appendix~\ref{sec:appendix_proof_of_bound}.
However, we are not aware of any practical estimation method to estimate this divergence.
Therefore, we resolve to measuring the total variation since it has a practical measurement method.

\section{Experiments}
\label{sec:experiments}
In Sections~\ref{sec_rank_preservation} and ~\ref{sec:model_selection_exp}, we perform experiments on CIFAR10~\citep{krizhevsky2009learning}.
In these sections, to evaluate the impact of the training set size, we use the following train-test splits: 10K-50K (Train10K), 30K-30K (Train30K), 50K-10K (Train50K). We emphasize that in these experiments the following set of rules hold:
\begin{enumerate}
    \item Only the training portion of the data is available for any training purposes.
    \item The test portion of the images is never used for model selection and is treated as non-existent for any training purposes.  
    \item In each experiment, GANs for generating synthetic datasets are trained only on the training portion of the images, \eg, for experiments with the Train10K dataset, the GANs are trained only on the 10K images.
\end{enumerate}

In Section~\ref{sec_imagenet} we demonstrate rank preservation on ImageNet~\citep{imagenet_cvpr09} and in Section~\ref{sec:imagenet_calib} we introduce a novel calibration method to improve the ranking.


\subsection{Rank Preservation}
\label{sec_rank_preservation}
In our first experiment we focus on several commonly used deep model architectures.
For each architecture, we select a number of variants. 
In total we experiment with 17 distinct architectures (see Appendix~\ref{sec:appendix_models_description} for details). 
For each architecture, 10 models were trained on each of the three datasets.
In Figure~\ref{fig:standard_correletion_graph}, we plot the empirical test errors, $\hat{\epsilon}_r$, vs. the empirical synthetic errors, $\hat{\epsilon}_s$, measured on datasets generated by four different GAN methods: (a) StyleGAN2-10, (b) StyleGAN2-Cond, (c) WGAN-GP-10, and (d) WGAN-GP-Cond (see details in Appendix~\ref{sec:appendix_data_generation_description}).
We can observe that, while in general $\hat{\epsilon}_r \neq \hat{\epsilon}_s$, for the StyleGAN2 based models, we are able to produce datasets that preserve the error ranking of different classification models.
We measure this using Spearman's rank correlation coefficient. 
For different GANs, we have measured the following ranking coefficients: 0.97 (a), 0.98 (b), -0.19 (c) and 0.14 (d).
In~\citet{sajjadi2018assessing} the connection between total variation ($\delta(\mu_r,\mu_s)$) and precision and recall for distributions (PRD) was established, and an empirical method for estimating it was suggested.
We use this method to empirically validate Corollary~\ref{cor:main}.
For the GANs above, we measured the following values: 3.5\% (a), 8.7\% (b), 43\% (c) and 34\% (d).
Indeed we see matching behaviors where the two models with high Spearman correlation have low total variation, and vice versa.
For example, it follows from Corollary~\ref{cor:main} that for GAN (a), if two hypotheses have $\Delta\epsilon_s(h_1,h_2) \ge 3.5\%$, then their rank on real data will be preserved.


\subsection{Model Selection}
\label{sec:model_selection_exp}
We consider three model selection scenarios where synthetic data can be used:
\begin{enumerate}
    \item \textbf{Early stopping (ES):} Given a training schedule of a single model, select an epoch from which to take the model weights from.
    \item \textbf{Random seed selection (RSS):} Given the same architecture and hyper-parameters, select a model instance out of $N$ trained models where the 
    difference between the models is the 
    randomness of the training process, \eg, weight initialization and dataset sampling order.
    \item \textbf{Hyper-parameter search (HPS):} Select a model out of a set of models trained with different hyper-parameters.
    Possible hyper-parameters are: learning rate, batch size, number of layers and network depth.
\end{enumerate}



\begin{figure}
\centering

\begin{tabular}{c c}
    \includegraphics[height=0.345\linewidth]{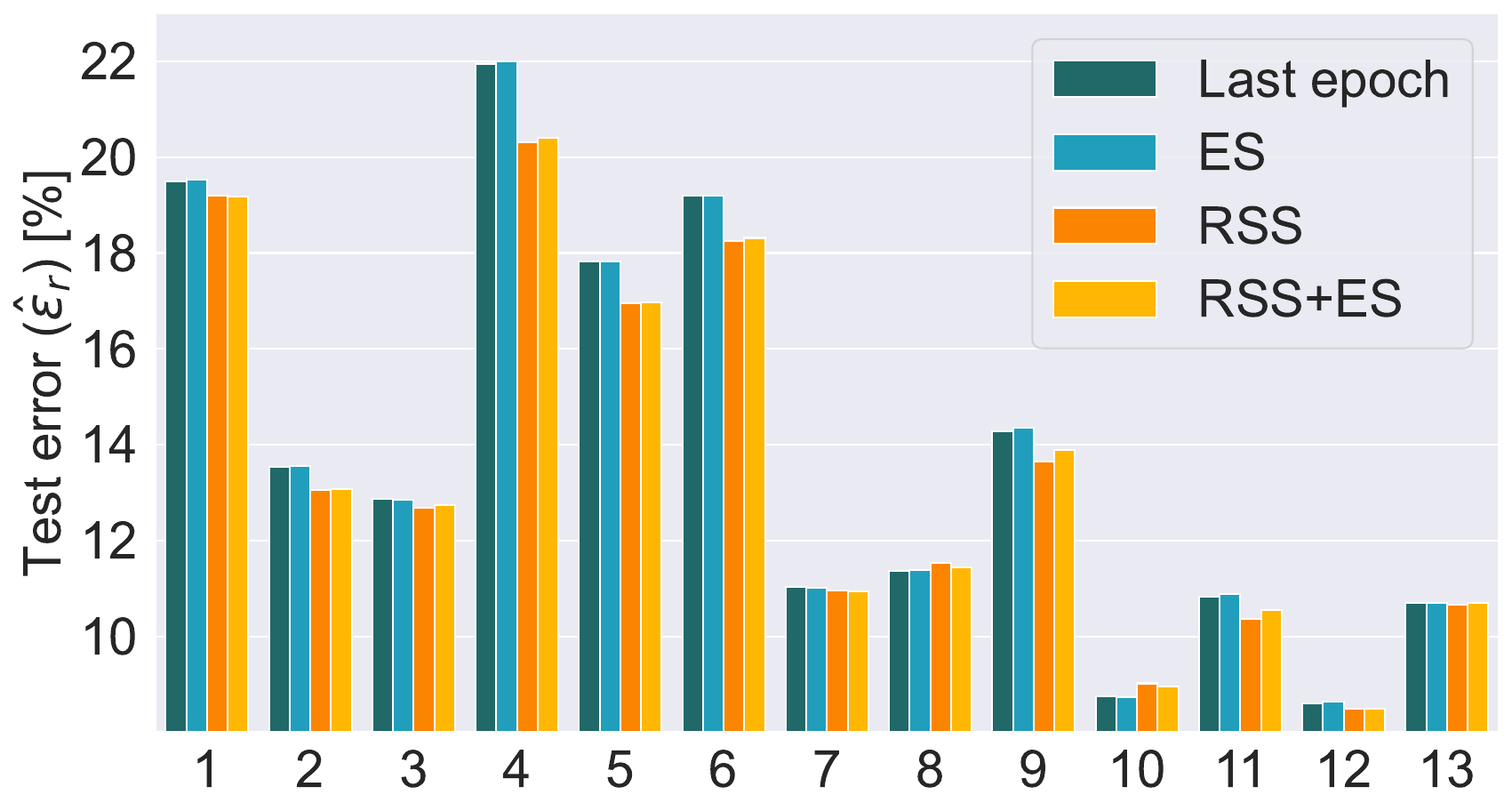} &
    \includegraphics[height=0.345\linewidth]{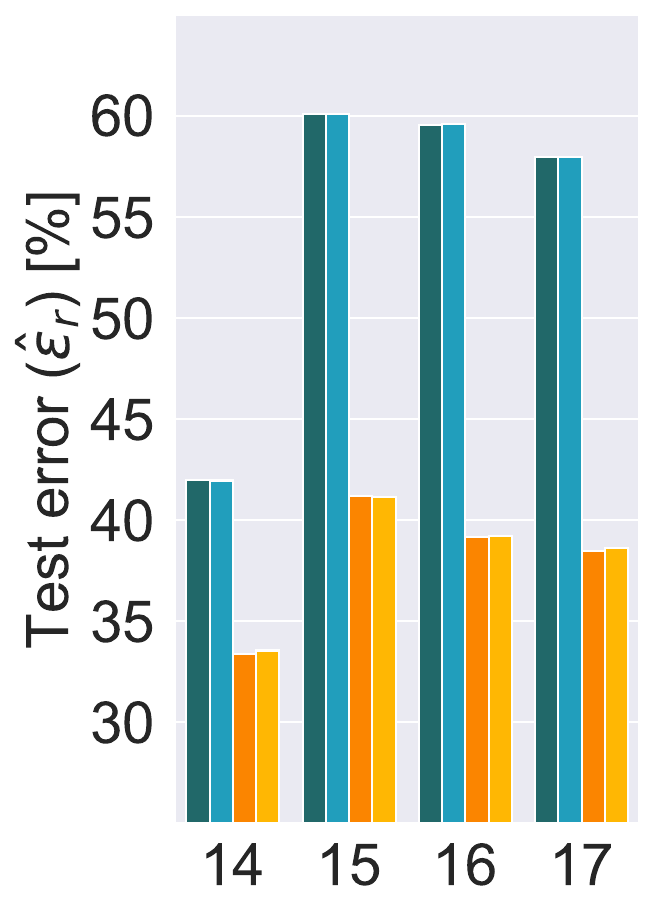}
\end{tabular}
\vspace{-0.1in}
\caption{\textbf{ES and RSS for model selection (standard architectures)$\downarrow$:} Test errors of the 17 architectures (x-axis corresponds to architectures described in Appendix~\ref{sec:appendix_models_description}) trained on the Train10K dataset. ``Last epoch'' and ES show the average error of the 10 models for each architecture. RSS and RSS+ES show the results of the selected model out of the 10 models.}
\label{fig:arch_tr10_barplots}
\end{figure}

\begin{table}[t]
    \centering
    \caption{\textbf{ES and RSS for model selection (randomly wired networks)$\downarrow$:}
    Average test error $\pm$ $95\%$ confidence intervals and standard deviation in parentheses for several model selection scenarios.
    Baseline -- all 640 models at the last epoch.
    ES -- all 640 models at the best synthetic set epoch.
    RSS -- 64 models at the best synthetic set epoch, where each of the 64 models was selected out of the 10 trained models for each architecture (by the best synthetic set error).
    ES+RSS -- 64 models at the best synthetic set epoch and selected by RSS.
    10K, 30K and 50K refer to the Train10K, Train30K and Train50K datasets respectively.
    }
    \label{tab:rwnn_es_rs_esrs}
    \small
    \renewcommand{\arraystretch}{1.2}
    \begin{tabular}{ l  c  c  c  c}
        \toprule
           & Baseline & ES & RSS & ES + RSS \\ 
        \midrule
        \raisebox{-3pt}{10K} & $\underset{(1.50)}{19.38\scriptstyle{\pm0.12}}$ & $\underset{(1.50)}{19.36\scriptstyle{\pm0.12}}$ & \textbf{$\underset{\pmb{(1.17)}}{\pmb{18.79}\scriptstyle{\pmb{\pm0.29}}}$} & $\underset{(1.15)}{18.88\scriptstyle{\pm0.28}}$ \\ 
        \raisebox{-3pt}{30K} & $\underset{(0.45)}{9.19\scriptstyle{\pm0.03}}$ & $\underset{(0.44)}{9.19\scriptstyle{\pm0.03}}$ & \textbf{$\underset{\pmb{(0.37)}}{\pmb{9.09}\scriptstyle{\pmb{\pm0.09}}}$} & $\underset{(0.39)}{9.1\scriptstyle{\pm0.10}}$ \\ 
        \raisebox{-3pt}{50K} & $\underset{(0.28)}{7.09\scriptstyle{\pm0.02}}$ & $\underset{(0.27)}{7.1\scriptstyle{\pm0.02}}$ & $\underset{(0.31)}{7.02\scriptstyle{\pm0.08}}$ & \textbf{$\underset{\pmb{(0.30)}}{\pmb{7.01}\scriptstyle{\pmb{\pm0.08}}}$} \\
        \bottomrule
    \end{tabular}
    
\vspace{-0.05in}
\end{table}
\subsubsection{Early stopping and random seed selection on standard architectures}
\label{sec_es_rss_standard}
We explore the impact of synthetic data on the ES and RSS model selection scenarios and their combination. 
We highlight that both of these scenarios require a held-out dataset. 
Therefore in the standard pipeline they cannot be used when training the model on the entire dataset.
Using synthetic data one is able to utilize these model selection scenarios.
For ES, the best synthetic epoch was selected for every training run. 
For RSS, per architecture, the model that performed the best on synthetic data at the last epoch was selected.
For RSS + ES, per architecture, the model that performed the best on synthetic data at its best epoch was selected.
We first experiment with the same standard model architectures as in Section~\ref{sec_rank_preservation}.
Figure~\ref{fig:arch_tr10_barplots} shows the results on Train10K, demonstrating that for nearly all architectures RSS improves accuracy. 
On the other hand, ES demonstrates only marginal impact on accuracy. 
This might be because the models' accuracy hardly changes across the last epochs, where the model has already converged (see Appendix~\ref{sec:appendix_convergence_plots} for convergence plot examples).
In Appendix~\ref{sec:appendix_full_barplots} we show the results for all datasets, where RSS shows comparable or better performance.

\subsubsection{Early stopping and random seed selection on similar architectures}
\label{sec:es_rso_rwnn}
Next, we evaluate the impact of ES and RSS across multiple models with similar architectures.
For each dataset we constructed 64 architectures.
For generating the architectures we used randomly wired neural networks (RWNN) framework~\citep{Xie_2019_ICCV} with WS(2,0.25), resulting in 64 unique but similar architectures per dataset.
Each architecture is trained 10 times on each dataset (a total of 1920 models were trained). 
Table~\ref{tab:rwnn_es_rs_esrs} concludes the experiment.
Since the errors of the models are of the same scale, we report the average performance and 95\% confidence intervals.
RSS has a significant impact on model selection with an average improvement over the baseline of $0.59$/$0.1$/$0.07$ (corresponding to: Train10K, Train30K, Train50K).
Similarly to the previous experiment ES has no significant impact.

\begin{table}
    \centering
    \caption{\textbf{Architecture hyper-parameters search results$\downarrow$:}
    Average test error $\pm$ $95\%$ confidence intervals and standard deviation in parentheses on held out test set.
    From left to right: 10 best models selected using synthetic data; 10 best architectures selected by real validation set and retrained on the entire training set; average error of all trained models.
    }
    \label{tab:architecture_search_conclusion}

    \small 
    \renewcommand{\arraystretch}{1.2}
    \begin{tabular}{ l c c c}
    \toprule
      & Synthetic & Standard & All models \\ 
    \midrule
    \raisebox{-3pt}{Train10K} \quad\quad\quad & 
    \textbf{$\underset{\pmb{(0.28)}}{\pmb{17.74}\scriptstyle{\pmb{\pm0.20}}}$} &
    $\underset{(0.38)}{18.06\scriptstyle{\pm0.08}}$ &
    $\underset{(1.50)}{19.39\scriptstyle{\pm0.12}}$ \\
    \raisebox{-3pt}{Train30K} \quad &
    \textbf{$\underset{\pmb{(0.12)}}{\pmb{8.64}\scriptstyle{\pmb{\pm0.09}}}$} &
    $\underset{(0.17)}{8.81\scriptstyle{\pm.03}}$ &
    $\underset{(0.45)}{9.17\scriptstyle{\pm.03}}$ \\
    \raisebox{-3pt}{Train50K} \quad & 
    \textbf{$\underset{\pmb{(0.19)}}{\pmb{6.78}\scriptstyle{\pmb{\pm0.14}}}$} &
    $\underset{(0.20)}{6.85\scriptstyle{\pm0.02}}$ &
    $\underset{(0.28)}{7.09\scriptstyle{\pm0.02}}$ \\
    \bottomrule
    \end{tabular}

\end{table}
\begin{figure*}
    \centering
    \begin{tabular}{c c c}
        Train50K & Train30K & Train10K \\
        \includegraphics[width=0.3\linewidth]{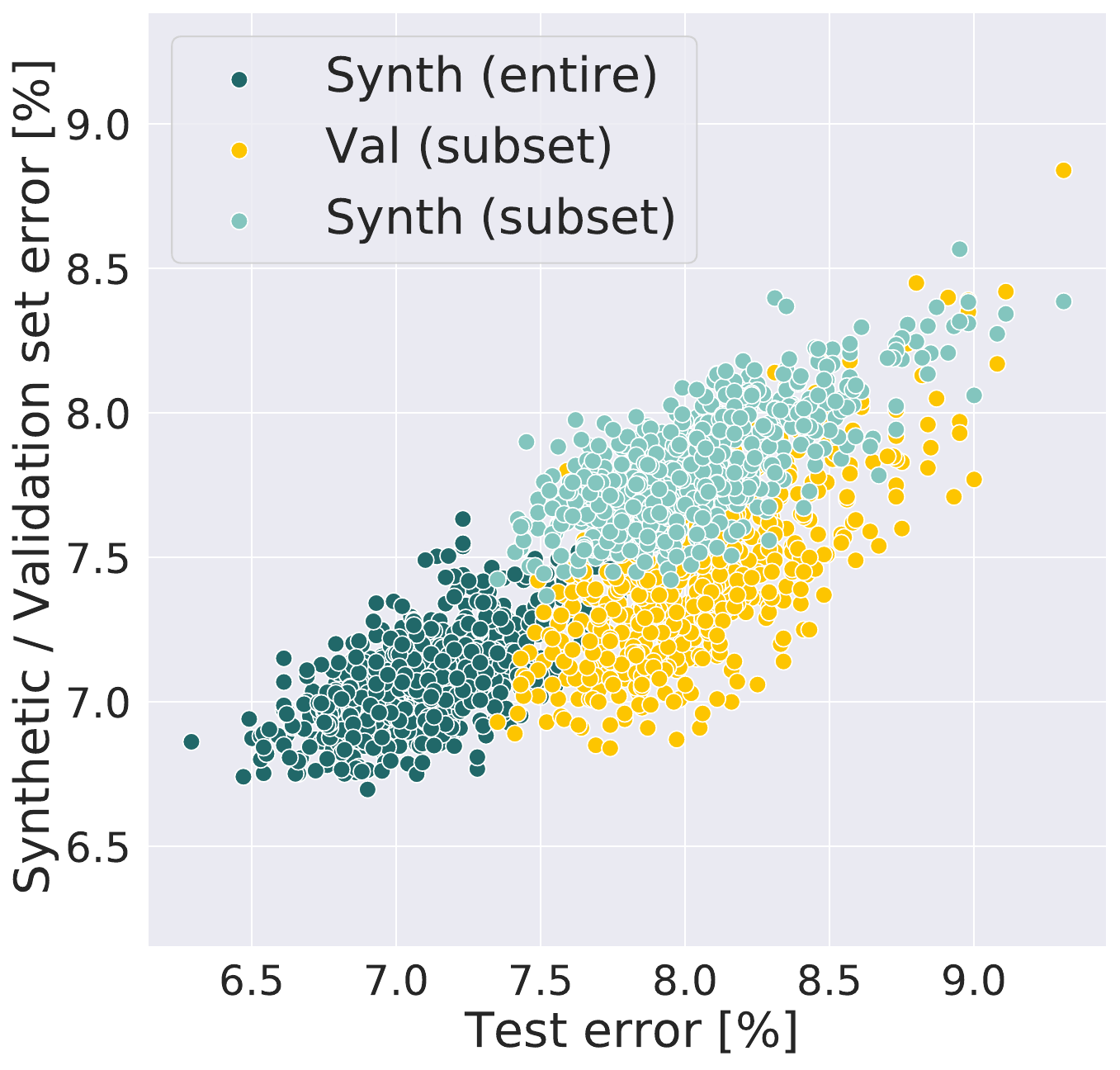} &
        \includegraphics[width=0.3\linewidth]{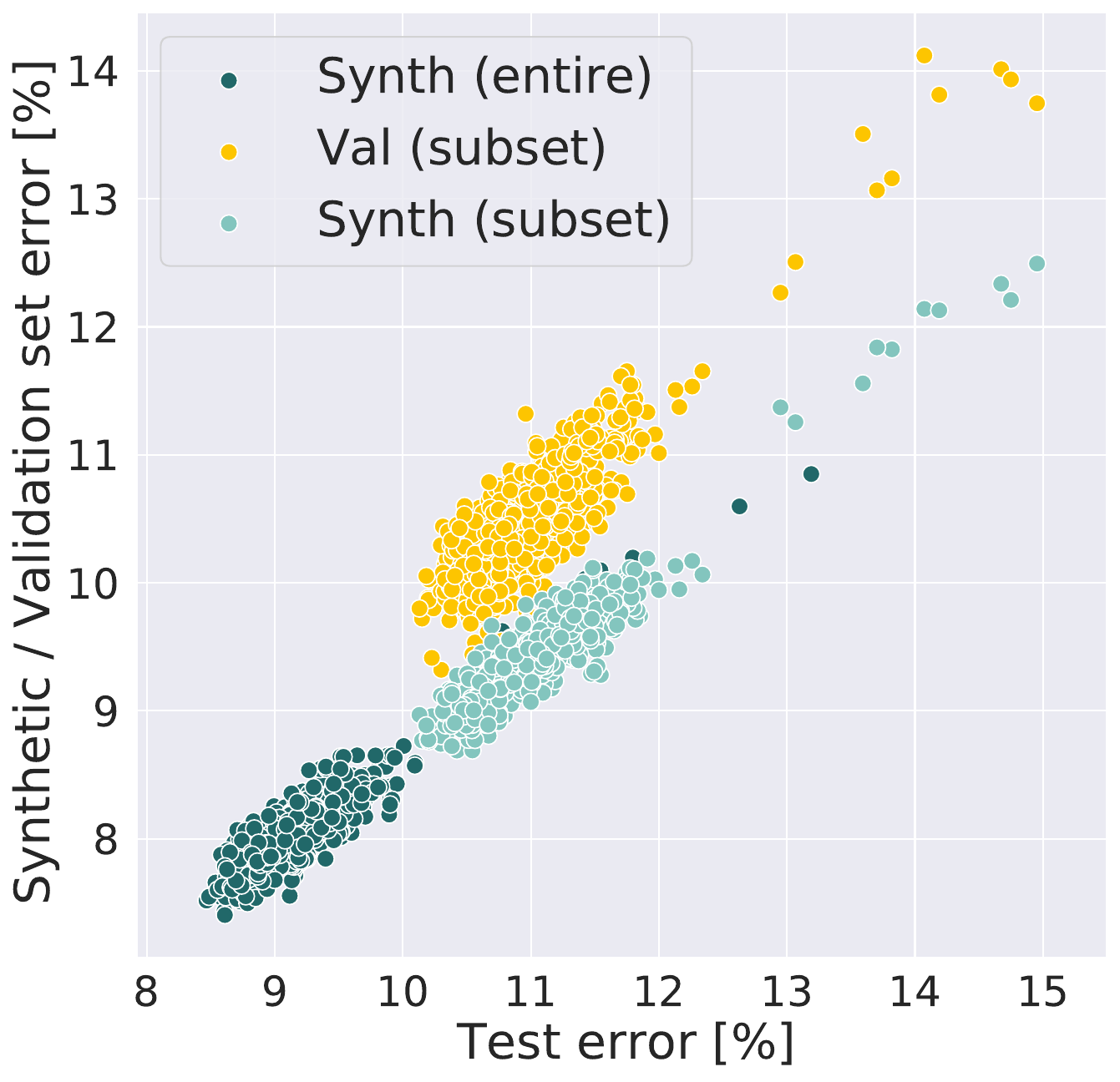} &
        \includegraphics[width=0.3\linewidth]{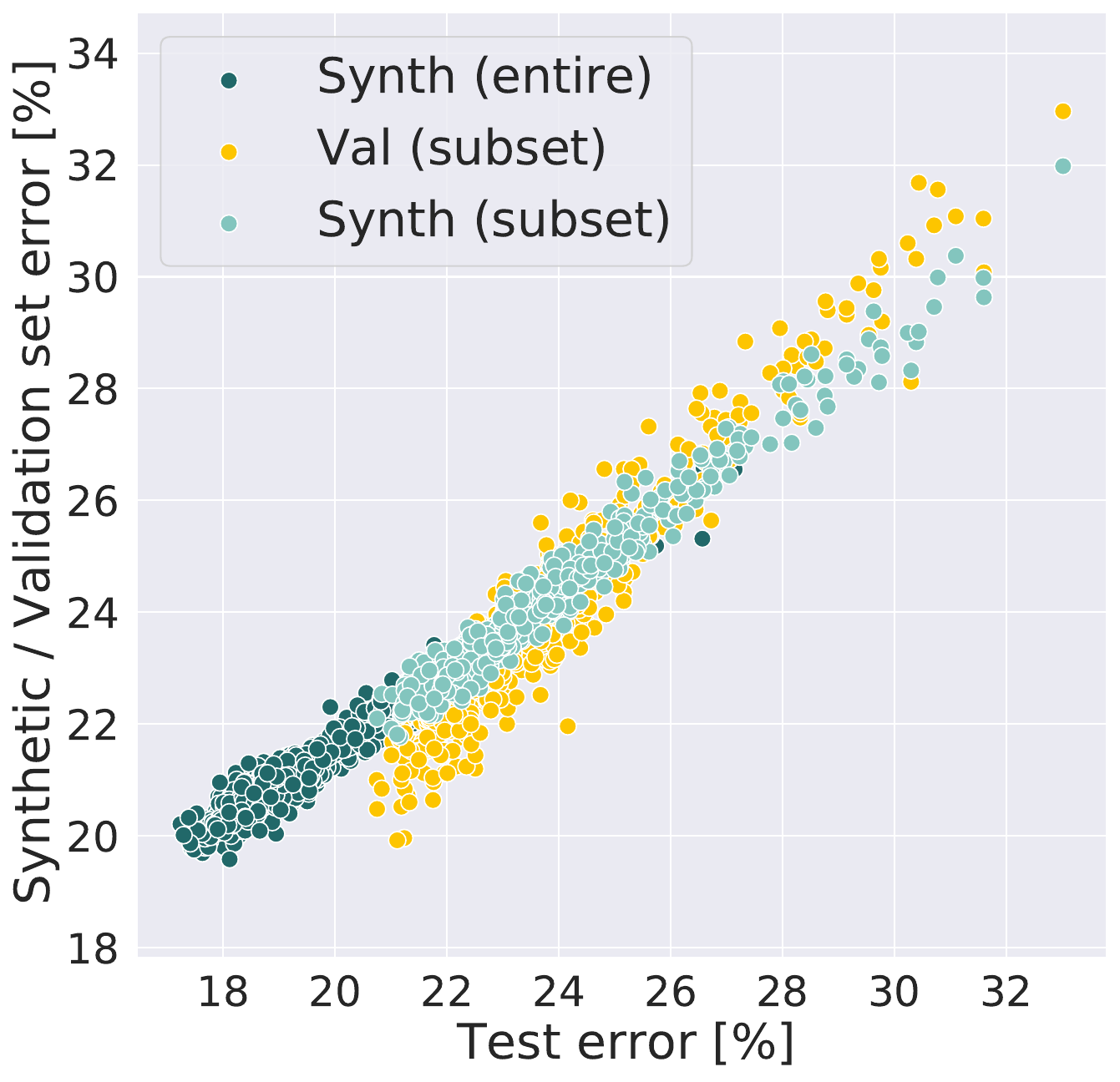} \\
    \end{tabular}
    \includegraphics[width=0.95\linewidth]{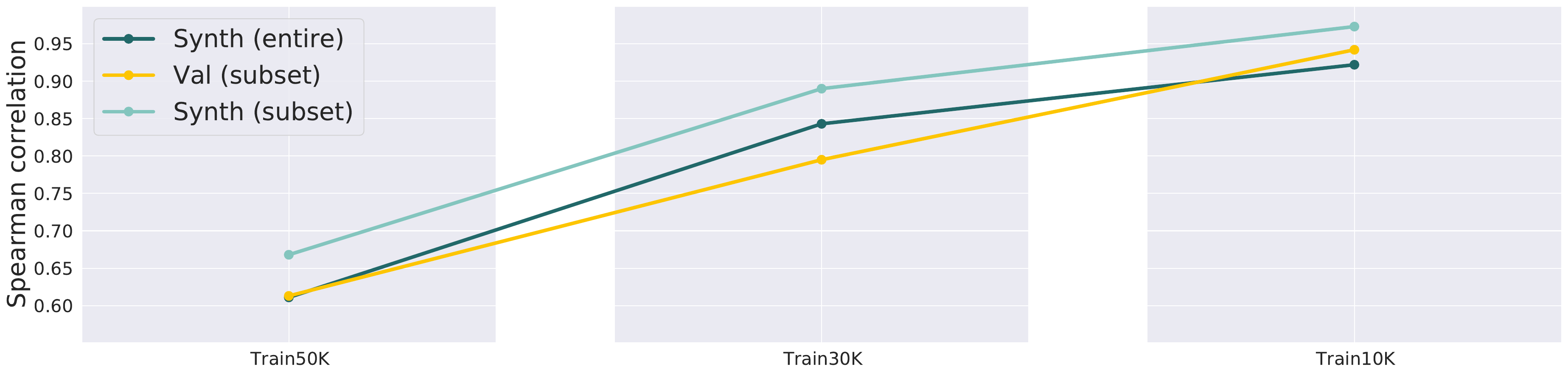} \\
    \begin{tabular}{c c c}
        \includegraphics[width=0.3\linewidth]{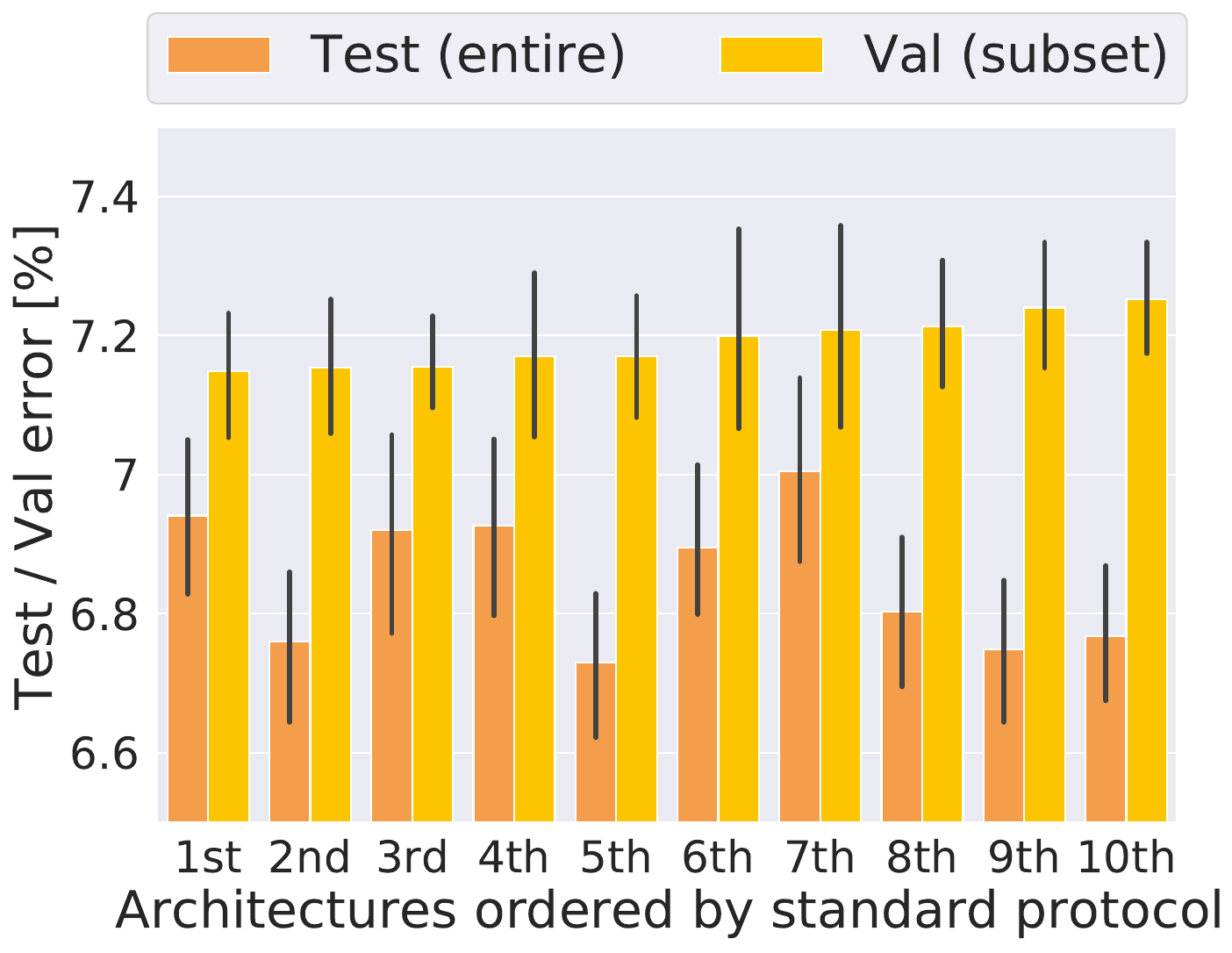} &
        \includegraphics[width=0.3\linewidth]{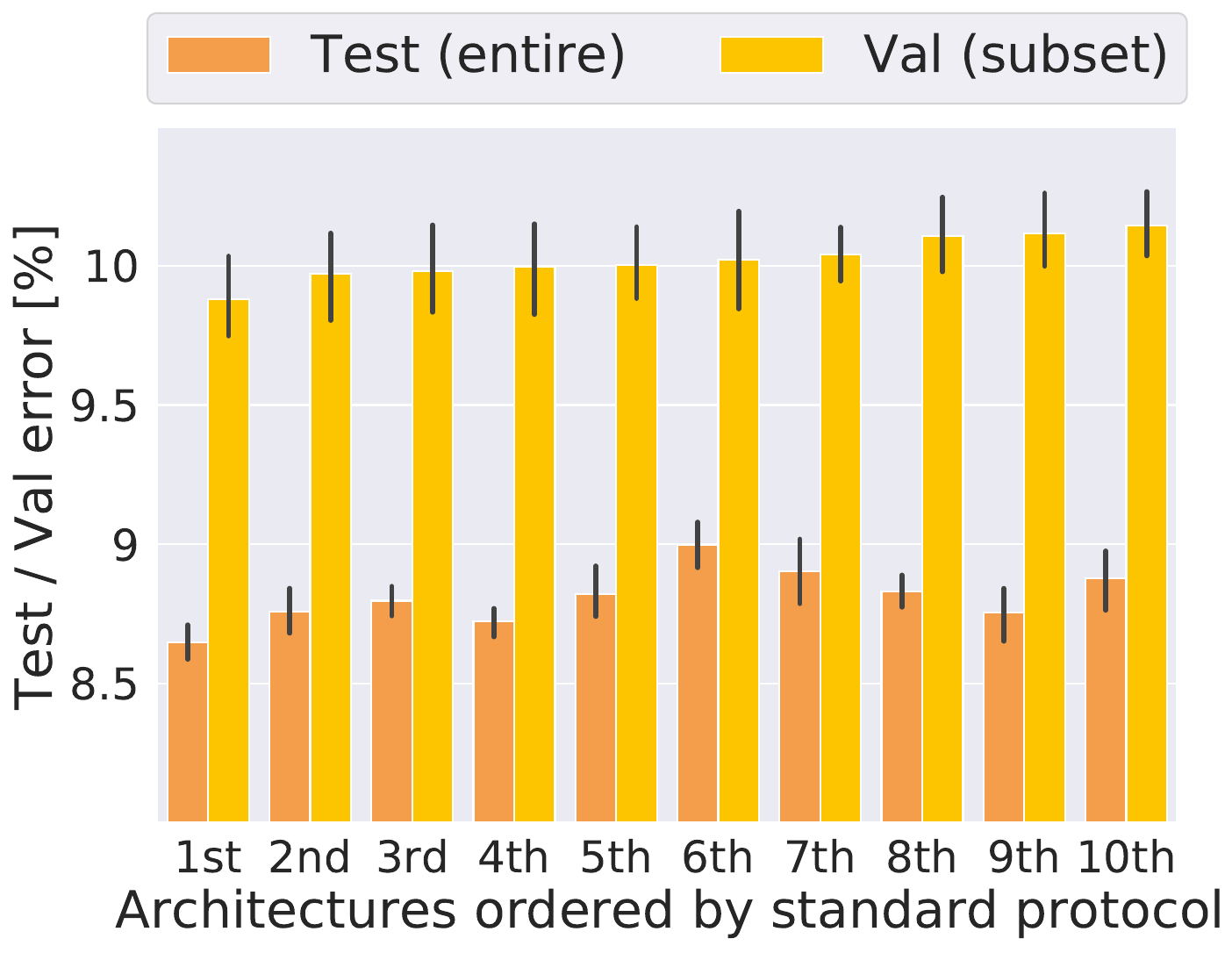} &
        \includegraphics[width=0.3\linewidth]{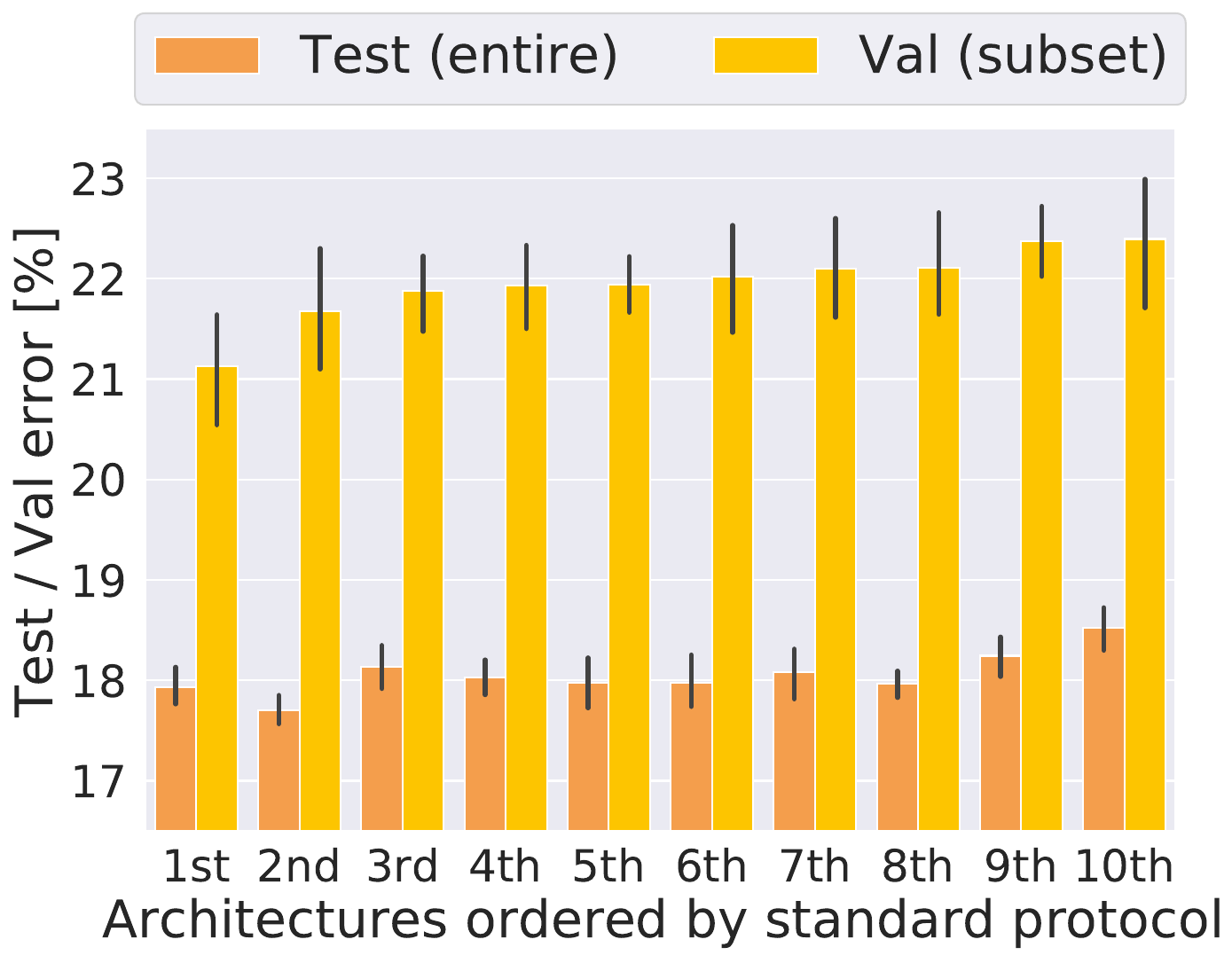} \\
    \end{tabular}
    \includegraphics[trim={0cm 17.8cm 0cm 0cm},clip,width=0.85\linewidth]{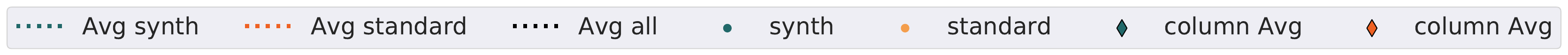} \\
    \begin{tabular}{c c c}
        \includegraphics[width=0.3\linewidth]{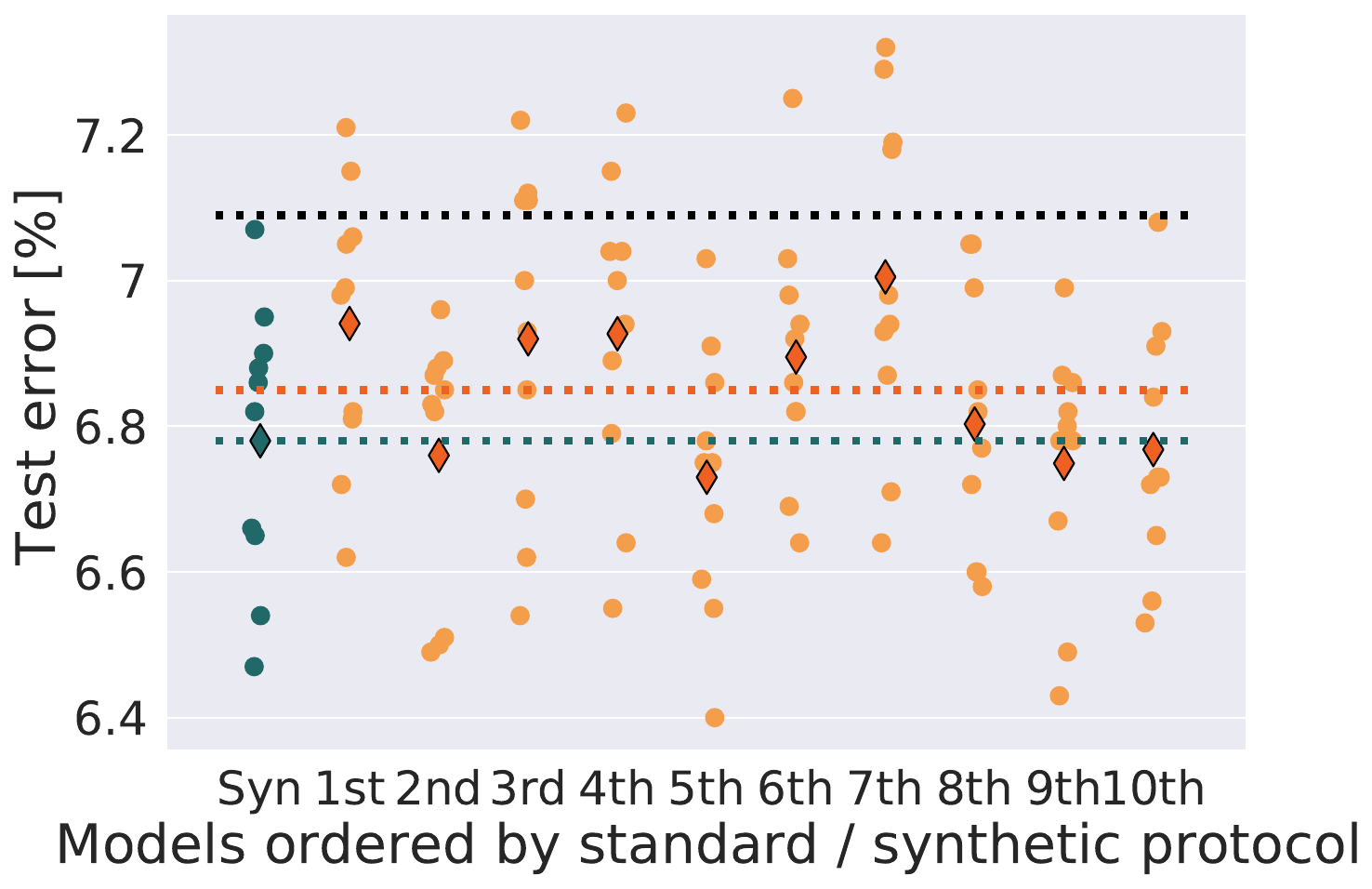} &
        \includegraphics[width=0.3\linewidth]{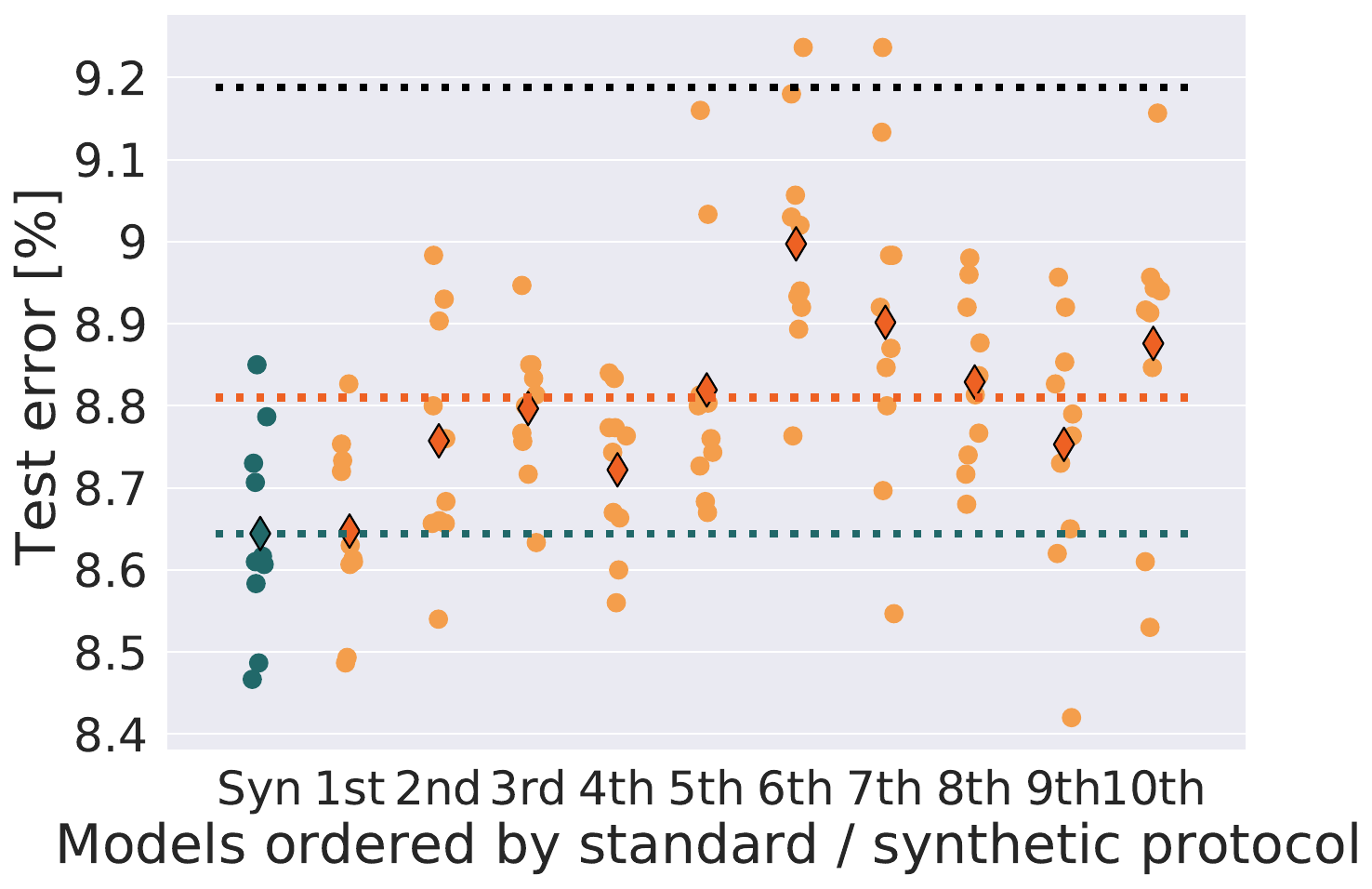} &
        \includegraphics[width=0.3\linewidth]{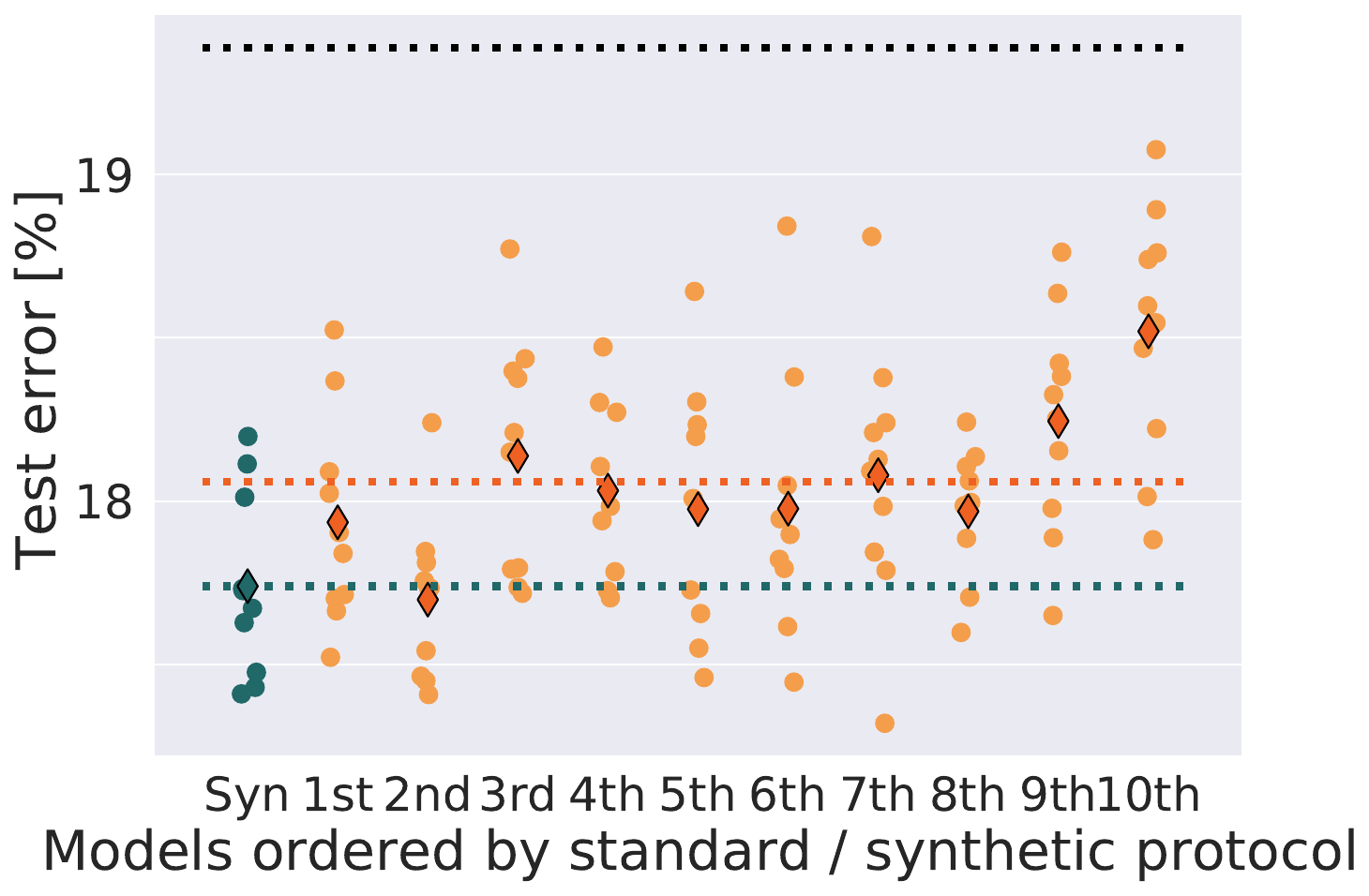} \\
    \end{tabular}
    
\caption{\textbf{Synthetic data for architecture hyper-parameter search:} 
\textbf{(1st row)} Each point represents the validation/synthetic set error (y-axis) and test error (x-axis) of a specific model.
``Entire'' corresponds to models that were trained on the entire dataset (Train50K, Train30K, Train10K). 
``Subset'' corresponds to models that were trained on the training subset. 
\textbf{(2nd row)} Spearman correlation between the validation/synthetic set errors and the test set errors.
\textbf{(3rd row)} 
Yellow bars represent the 10 architectures that performed the best (average of 10 training runs) on the validation set (ranked from the best to 10th best).
Orange bars represent the same architectures, trained on the entire dataset and their performance (average of 10 training runs) on the test set.
Black lines represents the 95\% confidence interval.
\textbf{(4th row)} 
The points in the first column, ``Syn'', correspond to the test errors of the 10 best performing models selected using the synthetic protocol.
The rest of the columns (1st-10th) correspond to the test error results of all trained models out of the 10 best architectures selected by the standard protocol (same architectures as row 3).
Horizontal lines represent average test error rates of: 10 best synthetic models (Avg synth), average of the 10 best models selected by the standard protocol (Avg standard), and the average of all 640 models (Avg all).
}

\label{fig:arc_search_synth_val_test_correlation}
\end{figure*}
\subsubsection{Synthetic data for architecture hyper-parameter search}
Next, we explore the contribution of using synthetic datasets for selecting a model out of multiple possible architectures and training instances (HPS). 
We consider three model selection protocols:
\begin{enumerate}
    \item \textbf{Selecting a random model:} The na\"ive baseline of selecting a random model. 
    \item \textbf{Standard protocol:} Split the dataset into training and validation subsets. 
    Then: (1) train each architecture $N$ times with the training subset; 
    (2) select the architecture that on average performed the best on the validation subset; 
    (3) Train the selected architecture on the entire dataset. 
    This methods allows for selecting a ``promising'' architecture without the ability to select a specific trained model instance (architecture and weights). 
    \item \textbf{Synthetic protocol:} 
    (1) Train each architecture $N$ times on the entire dataset; 
    (2) evaluate the accuracy at each training epoch on the synthetic dataset; 
    (3) select the model that preformed the best in step 2. 
    This method allows for selecting a ``promising'' model instance.
\end{enumerate}
Given a training set and a held-out test set (not available for model selection), we compared the three model selection protocols.
The standard protocol requires a validation set, to this end we split each of the datasets into training and validation subsets (train/val): Train10K was split into 7.5K/2.5K, Train30K was split into 22.5K/7.5K and Train50K was split into 40K/10K (see Appendix~\ref{sec:appedix_architecture_search} for more details).
In each experiment 64 architectures were evaluated using the different protocols.
For generating similar architectures, we sampled RWNN architectures with the same parameters WS(2, 0.25) (same architectures as in~\ref{sec:es_rso_rwnn}).
In both the standard and synthetic protocols we trained each architecture 10 times.
For the standard protocol we train on the training subset (step 1) and for the synthetic protocol we use the entire dataset.
Note that for the standard protocol, it is not possible to select a model instance out of the 10 trained instances of each architecture that was trained on the entire data (step 3).
Therefore, we use the average test error of the 10 trained models of each architecture (on the entire dataset) as a data point for comparisons.

Figure~\ref{fig:arc_search_synth_val_test_correlation} shows different analyses of the experimental results.
From the first two rows of the figure we can infer that there is a strong correlation between the error on synthetic data, $\hat{\epsilon_s}$, and the error on the test set, $\hat{\epsilon_r}$. 
We evaluate this correlation using Spearman's rank correlation coefficient, as it is appropriate for measuring rank preservation.
From the Spearman correlation plot (second row) we learn that the ranking capability of the synthetic data is comparable to that of the real data validation set. 
This strengthens our premise that using synthetic data for model selection is appropriate.
It can be seen that the correlation improves when the training set is smaller and the errors are larger.
This result coincides with Corollary~\ref{cor:main} where for larger gaps in synthetic error, there is a lower chance for a flip in model ranking.
From the third row we can infer that the ranking of architectures might change when moving from training on a smaller training set and evaluating on a validation set to training on the entire dataset end evaluating on the test set. 
This implies that a potential gain in accuracy could be achieved by selecting a model out of the models that were directly trained on the entire dataset.
From the last row we can learn that, on average, model selection using synthetic data improves over the standard method.
Again, the impact of synthetic data increases as the training dataset size decreases.
Given that a synthetic dataset is available, training the models directly on the entire dataset is simpler than training on a subset and re-training on the entire dataset.
Table~\ref{tab:architecture_search_conclusion} summarizes the experiments results and shows that the synthetic protocol achieves, on average, better results compared to the standard protocol.

\begin{figure}
\centering

\includegraphics[width=0.98\linewidth]{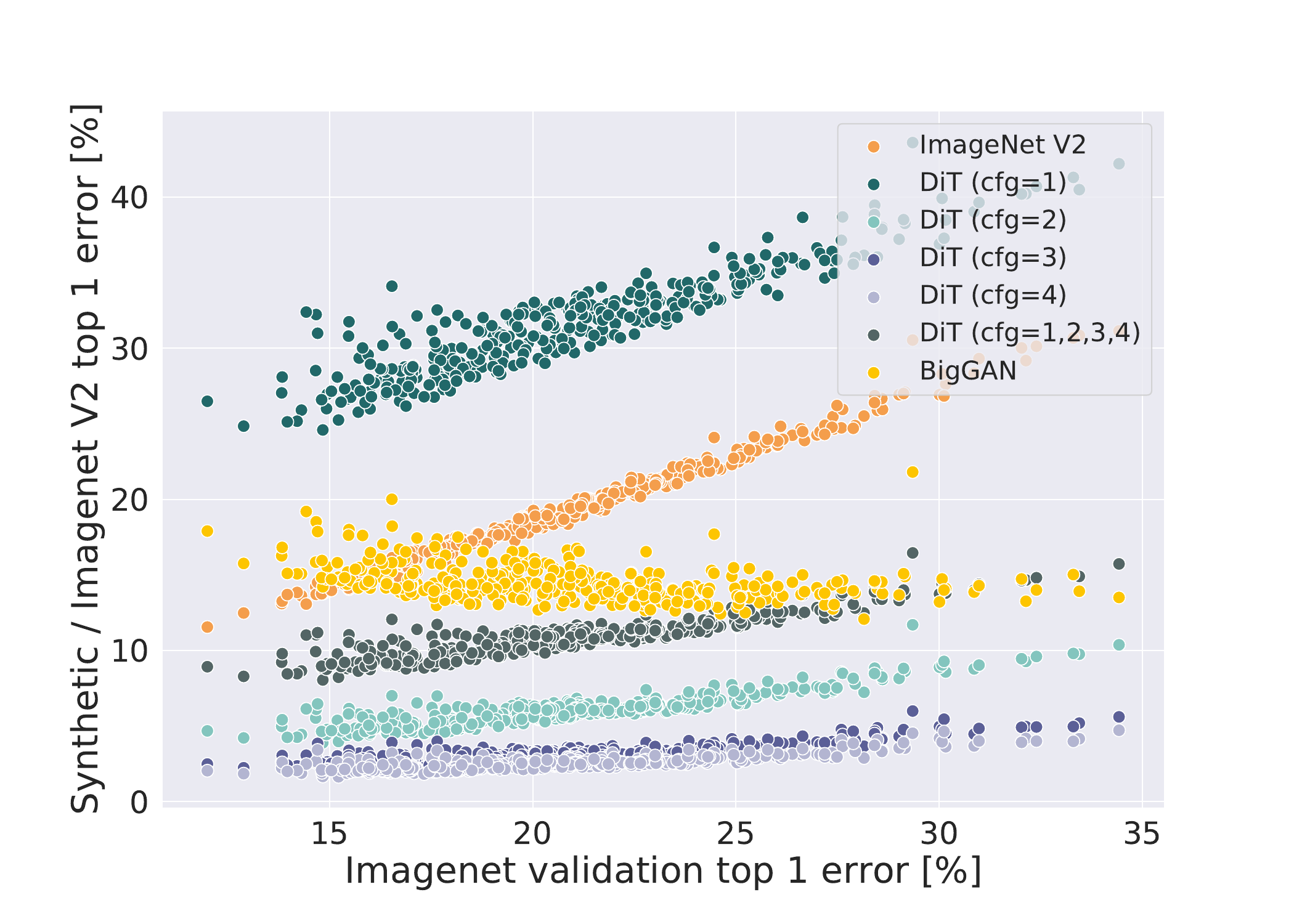}
\vskip -0.15in
\caption{\textbf{Synthetic vs. ImageNet errors:}
Each point represents the synthetic data error (y axis), $\hat{\epsilon}_s$ and ImageNet validation error (x axis), $\hat{\epsilon}_r$, of a single model. A total of 390 models trained on ImageNet were used.
Each model was evaluated by multiple synthetic datasets represented by different colors.
Additionally, we evaluated the models on real ImageNet V2 images as a baseline (orange points).}

\label{fig:imagenet}
\end{figure}
\begin{table}[h!]
    \centering
    \caption{\textbf{Ranking models trained on ImageNet using synthetic data$\uparrow$:}
    Top1 and Top 5 columns show the Spearman correlation between the \textit{top 1} error of 390 evaluation models on each dataset to the \textit{top 1} and \textit{top 5} error of the same models on the ImageNet validation set.
    The three first rows show the correlation on variants of ImageNetV2.
    Rows 4-9 show the correlation on synthetic datasets. 
    The last six rows show the correlation on calibrated synthetic datasets.} 
    \label{tab:imagenet_corr}
    \small
    \renewcommand{\arraystretch}{1.2}
    \begin{tabular}{ l  c c  c}
        \toprule
        Dataset 
        & Calibrated & Top 1   & Top 5 \\
        \midrule
        ImageNetV2 (format a)  & & $0.994$    & $0.994$ \\
        ImageNetV2 (format b)  & & $0.996$    & $0.994$ \\
        ImageNetV2 (format c)  & & $0.994$    & $0.995$ \\
        \midrule
        BigGAN                 & & $-0.363$   & $-0.357$ \\
        DiT (cfg $=1$)         & & $0.907$    & $0.900$ \\
        DiT (cfg $=2$)         & & $0.874$    & $0.862$ \\
        DiT (cfg $=3$)         & & $0.819$    & $0.801$ \\
        DiT (cfg $=4$)         & & $0.788$    & $0.768$ \\
        DiT (cfg $=1,2,3,4$)   & & $0.893$    & $0.880$ \\
        \midrule
        BigGAN                 & \checkmark & $0.978$ & $0.970$ \\
        DiT (cfg $=1$)         & \checkmark & $0.984$ & $0.977$ \\
        DiT (cfg $=2$)         & \checkmark & $0.980$ & $0.971$ \\
        DiT (cfg $=3$)         & \checkmark & $0.974$ & $0.963$ \\
        DiT (cfg $=4$)         & \checkmark & $0.966$ & $0.954$ \\
        DiT (cfg $=1,2,3,4$)   & \checkmark & \pmb{$0.986$} & \pmb{$0.978$} \\
        \bottomrule
    \end{tabular}


\end{table}
\subsection{Rank Preservation on ImageNet}
\label{sec_imagenet}
\begin{figure*}[]
\centering

\begin{subfigure}[b]{0.49\linewidth}
    \includegraphics[width=0.99\linewidth]{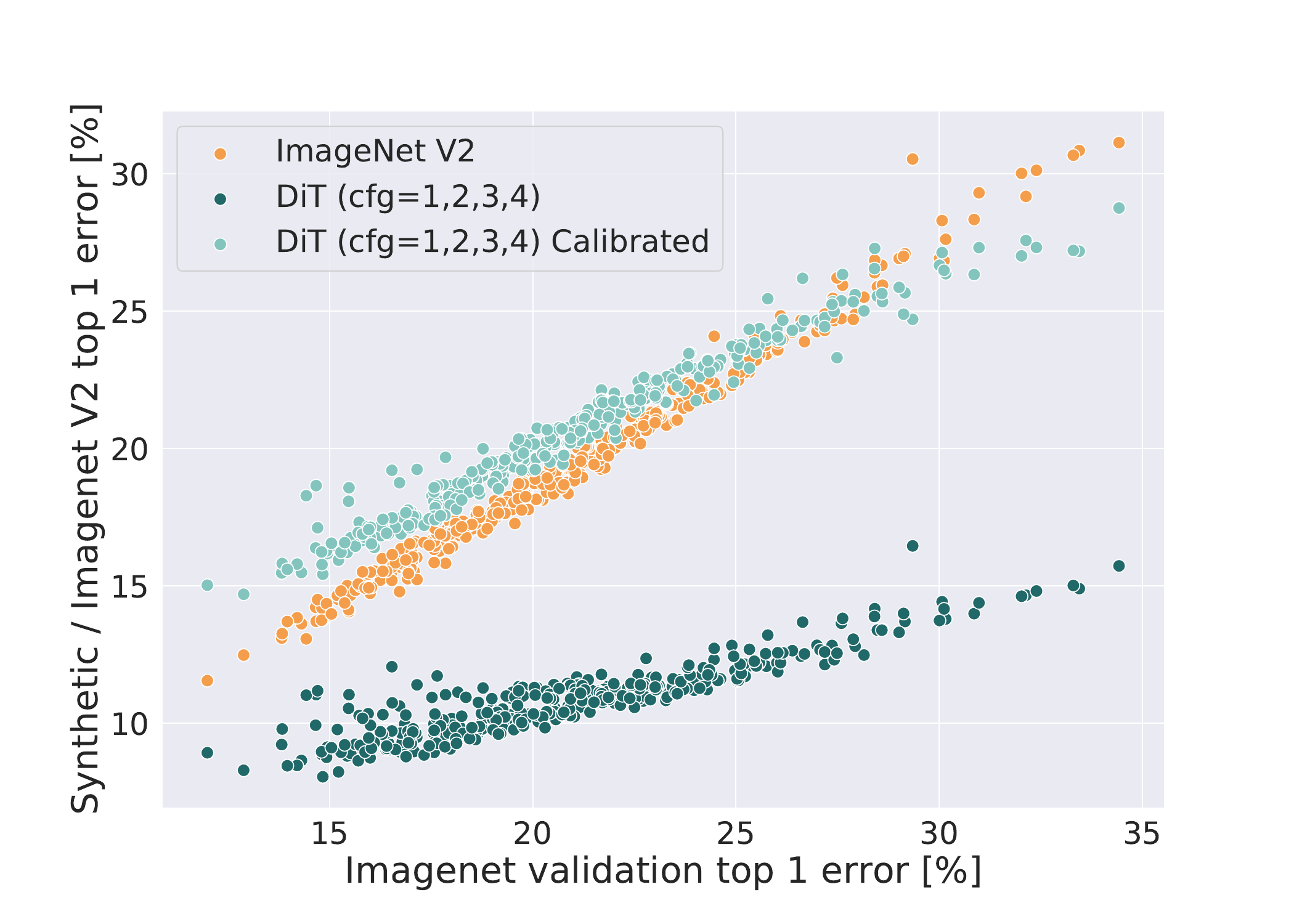}
    \vskip -0.1in
    \caption{DiT (cfg $=1,2,3,4$)}
    \label{fig:a}
\end{subfigure}
\begin{subfigure}[b]{0.49\linewidth}
    \includegraphics[width=0.99\linewidth]{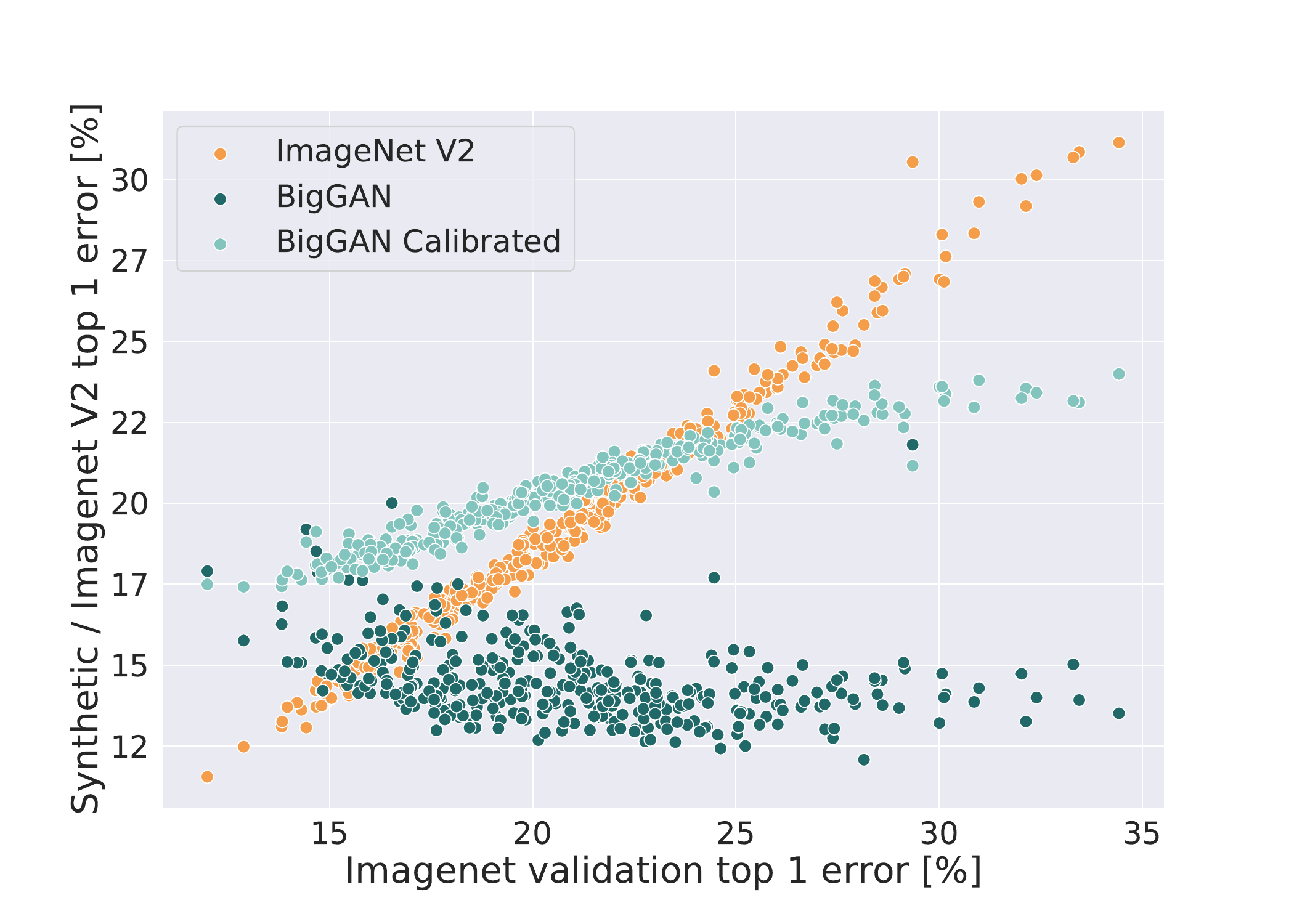}
    \vskip -0.1in
    \caption{BigGAN}
    \label{fig:a}
\end{subfigure}
    
\vskip -0.05in
\caption{\textbf{Synthetic data calibration:}
We demonstrate improved rank preservation on 390 held-out models after the calibration process, which utilized the remaining 50 models. 
(a) shows results for DiT, (b) shows the results for BigGAN.} 


\vskip -0.1in
\label{fig:imagenet_calib}
\end{figure*}
\begin{table}[t]
    \centering
    \caption{\textbf{Spearman correlation vs. number of calibration models (M)$\uparrow$.}
    }
    \label{tab:imagenet_calib_train_n}
    \small
    \renewcommand{\arraystretch}{1.2}
    \begin{tabular}{l c c c c}
        \toprule
        & M $=5$ & M $=10$ & M $=30$ & M $=50$ \\
        \midrule
        BigGAN            & 0.931 & 0.940 & 0.973 & 0.978 \\
        DiT (cfg=1,2,3,4) & 0.965 & 0.966 & 0.979 & 0.986 \\
        \bottomrule
    \end{tabular}
    \vskip -0.1in
\end{table}
In this section we explore the potential of synthetic data to be used for model selection in a more challenging domain.
To this end we choose the ImageNet~\citep{imagenet_cvpr09} 1,000 classification task.
The main differences between CIFAR10 and ImageNet is that the former has 5,000 low-res images for each of the 10 classes, while the latter has 732--1,300, high resolution images for each of its 1,000 classes.

We use BigGAN~\citep{brock2018large} and DiT~\citep{Peebles2022DiT} as our synthetic image generation models.
These models were selected for two reasons: both are conditioned on the 1,000 ImageNet classes and do not use a pre-trained classifier as conditional guidance. 
Additionally, the generation code of both models was publicly released by the authors.
We have generated the following six synthetic datasets: 
a dataset containing 100 images per class, generated by BigGAN with truncation $=0.7$;
four datasets, each containing 100 images per class, generated by DiT with classifier free guidance~\citep{ho2021classifierfree} of $1$, $2$, $3$ or $4$; and a dataset that is the combination of the four previous DiT datasets (cfg $=1,2,3,4$).
In our experiments we used ImageNetV2~\citep{recht2019imagenet} as a baseline, which is a test set attempting to mimic the distribution of ImageNet.
For rank preservation analysis, we utilized all the models from TIMM~\citep{rw2019timm} and pretrained-models\footnote{\scriptsize{\url{https://github.com/cadene/pretrained-models.pytorch}.}} that were trained on images of 224x224 resolution.
This lead to a total of 440 models, that were divided randomly into 390 models for rank preservation analysis and 50 models that were set aside for the calibration process (to be explained in Section~\ref{sec:imagenet_calib}).
 
Both Figure~\ref{fig:imagenet} and Table~\ref{tab:imagenet_corr} demonstrate that rank preservation can be achieved to some extent using images generated by DiT (Spearman correlation of $0.893$).
However, the best synthetic dataset still has a gap in rank correlation compared to the real ImageNetV2 dataset, which has a Spearman correlation of $0.996$. 
The dataset generated by BigGAN is irrelevant for model ranking ($-0.363$).
We suspect that the weak ranking capability of BigGAN compared to DiT, is due to the gap in image quality and image variability (we present generated samples in Appendix~\ref{sec:appendix_imagenet_generated}). 



\subsection{Synthetic Dataset Calibration}
\label{sec:imagenet_calib}
In this section we describe a novel approach for reducing the error estimation gap between the synthetic and the real data. 
We propose to calibrate the synthetic data error estimation using a held out set of classifiers and show that this results in improved rank preservation across the domains.
We provide the following intuition behind the calibration process.
According to Theorem~\ref{th:main}, the ability to use synthetic data for ranking models depends only on the probability density gap between the synthetic and real distribution in the area of disagreement, $\delta_{h_1 \oplus h_2}(\mu_r, \mu_s)$. 
Thus, reducing the disagreement effectively improves the ranking.
Towards this goal, we re-weight the contribution of each synthetic sample to the estimated error.
The re-weighted error no longer corresponds to the original synthetic distribution, rather it corresponds to a distribution that is closer to the real one.
We assign weights such that the calculated weighted errors of a small set of calibration models, will be close to their corresponding real errors.
If the synthetic images contain information that allows discrimination between models, and the discrimination correlates with model ranking, then we expect the ranking to generalize to unseen models.

To achieve this, we construct the following regression problem. 
Given a set of $M$ models and $C$ classes, where each class has $N_C$ synthetic images, $\{ \{x_{i}^{c}\}_{i=1}^{N_c} \}_{c=1}^{C}$.
We denote $\hat{\epsilon}_{r,m}^c$ as the empirical risk of model $m$ over real images belonging to class $c$. 
We also denote $\hat{\mathbf{\epsilon}}_{r}^c = [ \hat{\epsilon}_{r,m}^c]_{m=1}^M$, \ie the real empirical risks of all models on class $c$.
$\mathbf{Q}_c \in \mathbb{Z}_2^{ N_c \times M }$ is a binary matrix where each column, $\mathbf{q}_{m}^c$, indicates the prediction correctness of model $m$ on images $\{x_i^c\}_{i=1}^{N_c}$.
We wish to find a set of weights, $\mathbf{w}_c\in \mathbb{R}^{N_c}$, that solves the linear ridge regression problem:
\begin{equation}
\mathbf{w}_c = \arg\!\min_{\mathbf{w}} \biggl\{|| \hat{\mathbf{\epsilon}}^c_{r} - {\mathbf{Q}_c}^T \mathbf{w} ||_2^2 + \lambda ||\mathbf{w}||_2^2\biggr\},
\end{equation}
where $\lambda$ is the Lagrange multiplier\footnote{We set $\lambda=0.5$ and include an intercept (bias) term.}. 
Intuitively, the weight penalty ``spreads'' the error influence across many synthetic images, preventing a single image to dominate the ranking.
This prevents ``overfitting'' the solution to the models used for calibration.
For each class, $c$, we solve an independent optimization problem using a closed-form solution\footnote{$\mathbf{w}_c = (\mathbf{Q}_c^T\mathbf{Q}_c+\lambda \mathbf{I})^{-1}\mathbf{Q}_c^T \hat{\mathbf{\epsilon}}^c_{r}$.}.
Once optimization is done, we estimate the error of a new model $m'$, on the calibrated dataset by $\hat{\epsilon}_{s,m'} = \sum_{c=1}^C  (\mathbf{q}_{m'}^c)^T \mathbf{w}_c$.

For this experiment, we use the $M=50$ models that were set aside for calibration.
Figure~\ref{fig:imagenet_calib} and Table~\ref{tab:imagenet_corr} demonstrate a drastic improvement in rank preservation.
For the DiT (cfg $=1,2,3,4$) dataset the calibration process improved the Spearman's correlation from $0.893$ to $0.986$, which is comparable to the ImageNetV2 ($0.994$).
Surprisingly, BigGAN's performance improved from total irrelevance ($-0.363$) to be competitive for rank preservation ($0.978$), surpassing all uncalibrated datasets.
We present synthetic images and their corresponding calibration coefficients in Appendix~\ref{sec:appendix_imagenet_calibrated}.
We suspect that while the images are not coherent, they may contain certain features that are correlated with the correct class. 
Models that are able to detect these features will tend to have better performance, and vice versa, some features may be negatively correlated with a certain class, \ie being able to detect it indicates poorer accuracy for this class.

In Table~\ref{tab:imagenet_calib_train_n} we present the Spearman correlation while varying the number of models used for calibration ($M$).
The results indicate that calibrating with five models is enough to achieve a significant improvement.

\section{Conclusions}
\label{sec:conclusion}
In this paper we presented a comprehensive empirical study, evaluating the impact of using synthetic data for model selection.
The empirical evidence suggest that evaluating trained models on synthetic data can outperform the standard methods for model selection that are based solely on the available real images.
In addition, we show that synthetic data can be used to rank models trained on high-resolution images from a diverse set of classes and we propose a novel calibration method that significantly improves the ranking capabilities.  


\newpage
\newpage

\bibliography{main_icml.bib}
\bibliographystyle{icml2023}

\newpage
\appendix
\onecolumn
\newpage

\appendix

\section{Connection to $\mathcal{H}\Delta\mathcal{H}$-divergence~\cite{ben2010theory}}
\label{sec:appendix_proof_of_bound}
In this section we show a connection to another common metric for measuring the difference between domains, the $\mathcal{H}\Delta\mathcal{H}$-divergence.

\begin{lemma}
[Lemma 3 from~\cite{ben2010theory}] For any pair of hypotheses $h_1, h_2\in \mathcal{H}$,
\begin{align*}
|\epsilon_s(h_1, h_2) - \epsilon_r(h_1, h_2)| \leq \frac{1}{2}d_{\mathcal{H}\Delta\mathcal{H}}(\mathcal{D}_s, \mathcal{D}_r).
\end{align*}
\end{lemma}
\begin{proof}
By the definition of $\mathcal{H}\Delta\mathcal{H}$-divergence,
\begin{align*}
d_{\mathcal{H}\Delta\mathcal{H}}(\mathcal{D}_s, \mathcal{D}_r) &= 2\sup_{h, h'\in\mathcal{H}}|Pr_{\mathbf{x}\sim \mathcal{D}_s}[h(\mathbf{x})\neq h'(\mathbf{x})]-Pr_{\mathbf{x}\sim \mathcal{D}_r}[h(\mathbf{x})\neq     h'(\mathbf{x})]| \\
                                                              &= 2\sup_{h, h'\in\mathcal{H}}|\epsilon_s(h, h') - \epsilon_r(h, h')| \geq 2|\epsilon_s(h_1, h_2) - \epsilon_r(h_1, h_2)|
\end{align*}
\end{proof}

\begin{theorem}
\label{th:main_sbd}
Let $\Delta\epsilon_r$ and $\Delta\epsilon_s$ denote the risk difference between two hypotheses, $h_1, h_2 \in \mathcal{H}$, measured over the real and the synthetic probability distributions $\mathcal{D}_r=(\Omega, \mu_r)$ and $\mathcal{D}_s=(\Omega, \mu_s)$, respectively, \emph{\ie}, $\Delta\epsilon_r=\epsilon_r(h_2)- \epsilon_r(h_1)$ and $\Delta\epsilon_s=\epsilon_s(h_2)- \epsilon_s(h_1)$.
Let $f$ denote the labeling function. Then, for any $h_1, h_2 \in \mathcal{H}:$
\begin{align*}
\Delta \epsilon_s - \Delta \epsilon_r \leq d_{\mathcal{H}\Delta\mathcal{H}}(\mathcal{D}_s, \mathcal{D}_r)
\end{align*}

\end{theorem}

\begin{proof}
\begin{align*}
\Delta \epsilon_s - \Delta \epsilon_r &= \epsilon_s(h_2) - \epsilon_s(h_1) - (\epsilon_r(h_2) - \epsilon_r(h_1)) \\
                                      &= \epsilon_s(h_2) - \epsilon_r(h_2) + \epsilon_r(h_1) - \epsilon_s(h_1) \\
                                      &= \epsilon_s(h_2, f) - \epsilon_r(h_2, f) + \epsilon_r(h_1, f) - \epsilon_s(h_1, f) \\
                                      &\leq |\epsilon_s(h_2, f) - \epsilon_r(h_2, f)| + |\epsilon_r(h_1, f) - \epsilon_s(h_1, f)| \\
                                      &\leq \frac{1}{2}d_{\mathcal{H}\Delta\mathcal{H}}(\mathcal{D}_s, \mathcal{D}_r) + \frac{1}{2}d_{\mathcal{H}\Delta\mathcal{H}}(\mathcal{D}_r, \mathcal{D}_s) \\
                                      &= d_{\mathcal{H}\Delta\mathcal{H}}(\mathcal{D}_s, \mathcal{D}_r)
\end{align*}
\end{proof}



\begin{corollary}
Let $\mathcal{D}_r$ and $\mathcal{D}_s$, denote the real and the synthetic (generated) probabilistic distributions, respectively. Let $\Delta\epsilon_r$ and $\Delta\epsilon_s$ denote the risk differences between any two hypotheses, $h_1, h_2 \in \mathcal{H}$. Then,
\begin{align*}
\Delta\epsilon_s \geq d_{\mathcal{H}\Delta\mathcal{H}}(\mathcal{D}_s, \mathcal{D}_r) \Rightarrow \Delta\epsilon_r \geq 0,
\end{align*}
where $d_{\mathcal{H}\Delta\mathcal{H}}(\mathcal{D}_s, \mathcal{D}_r)$ is the $\mathcal{H}\Delta\mathcal{H}$-divergence between the two distributions.

\end{corollary}

\newpage

\newpage

\newpage

\section{Proof of Lemma 3.1}
\label{sec:appendix_proof_of_lemma}

\begin{lemma}
Let $\Delta\epsilon$ denote the risk difference between two hypotheses, $h_1, h_2 \in \mathcal{H}$, measured over a probability distribution $\mathcal{D}=\langle\Omega, \mu\rangle$, \emph{\ie}, $\Delta\epsilon=\epsilon(h_2)- \epsilon(h_1)$. Let $f$ denote the labeling function. Let $\Omega_1=\left\{\mathbf{x} \in \Omega | h_1(\mathbf{x}) \neq f(\mathbf{x}) \land h_2(\mathbf{x}) = f(\mathbf{x}) \right\}$ and $\Omega_2=\left\{\mathbf{x} \in \Omega | h_2(\mathbf{x}) \neq f(\mathbf{x}) \land h_1(\mathbf{x}) = f(\mathbf{x}) \right\}$.
Then, 
\begin{align*}
\Delta\epsilon=\int_{\Omega_2} \mu(\mathbf{x}) d\mathbf{x} - \int_{\Omega_1} \mu(\mathbf{x}) d\mathbf{x}.
\end{align*}
\end{lemma}
\begin{proof}
\begin{align*}
\Delta \epsilon &= \epsilon(h_2) - \epsilon(h_1) \\
                &= E_{\mathbf{x} \sim \mathcal{D}}[ h_2(\mathbf{x}) \ne f(\mathbf{x}) ] - E_{\mathbf{x} \sim \mathcal{D}}[ h_1(\mathbf{x}) \ne f(\mathbf{x}) ] \\
                &= E_{\mathbf{x} \sim \mathcal{D}}[ h_2(\mathbf{x}) \ne f(\mathbf{x}) \land h_1(\mathbf{x}) = f(\mathbf{x})] + E_{\mathbf{x} \sim \mathcal{D}}[ h_2(\mathbf{x}) \ne f(\mathbf{x}) \land h_1(\mathbf{x}) \ne f(\mathbf{x})] \\
                &~ - E_{\mathbf{x} \sim \mathcal{D}}[ h_1(\mathbf{x}) \ne f(\mathbf{x}) \land h_2(\mathbf{x}) = f(\mathbf{x})] - E_{\mathbf{x} \sim \mathcal{D}}[ h_1(\mathbf{x}) \ne f(\mathbf{x}) \land h_2(\mathbf{x}) \ne f(\mathbf{x})] \\ 
                &= E_{\mathbf{x} \sim \mathcal{D}}[ h_2(\mathbf{x}) \ne f(\mathbf{x}) \land h_1(\mathbf{x}) = f(\mathbf{x})] - E_{\mathbf{x} \sim \mathcal{D}}[ h_1(\mathbf{x}) \ne 
                  f(\mathbf{x}) \land h_2(\mathbf{x}) = f(\mathbf{x})] \\
                &= \int_{\Omega_2} \mu(\mathbf{x}) d\mathbf{x} - \int_{\Omega_1} \mu(\mathbf{x}) d\mathbf{x} \\
\end{align*}
\end{proof}

\newpage

\newpage

\section{Synthetic data for architecture hyper-parameter search}
\label{sec:appedix_architecture_search}
In this experiment we explore the contribution of synthetic data for model selection out of a pool of different architectures.
Given a training set and a held-out test set (not available for model selection), we compared the three model selection protocols (selecting a random network, standard protocol, synthetic protocol).
The standard protocol requires a validation set, to this end we split each of the datasets (Train50K, Train30K, Train10K) into training and validation subsets.
For the synthetic protocol a GAN was trained on each dataset to produce a dataset of 100K synthetic images (see Appendix~\ref{sec:appendix_data_generation_description}).
The train/val split and the GANs that where used to create each synthetic dataset are as follows: 
\begin{enumerate}
    \item \textbf{Train10K:} The train/val split is 7.5K/2.5K.
    For the synthetic data protocol, a single StyleGAN2-Cond model trained on the 10K available images was used.  
    \item \textbf{Train30K:} The train/val split is 22.5K/7.5K.
    For the synthetic data protocol, 10 StyleGAN2 models were trained, each on the 3K (per class) available images.
    \item \textbf{Train50K:} The train/val split is 40K/10K.
    For the synthetic data protocol, 10 StyleGAN2 models were trained, each on the 5K (per class) available images.
\end{enumerate}

\newpage

\section{Additional results for early stopping and random seed selection on standard architectures}
\label{sec:appendix_full_barplots}
\begin{figure}[h]

\centering
\begin{tabular}{c c c}
    \rotatebox{90}{
        \quad\quad\quad\quad Train10K
    } &
    \includegraphics[height=0.3\linewidth]{images/es_rs/10K_bars_0.pdf} &
    \includegraphics[height=0.3\linewidth]{images/es_rs/10K_bars_1.pdf} \\
    \rotatebox{90}{
       \quad\quad\quad\quad  Train30K
    } &
    \includegraphics[height=0.3\linewidth]{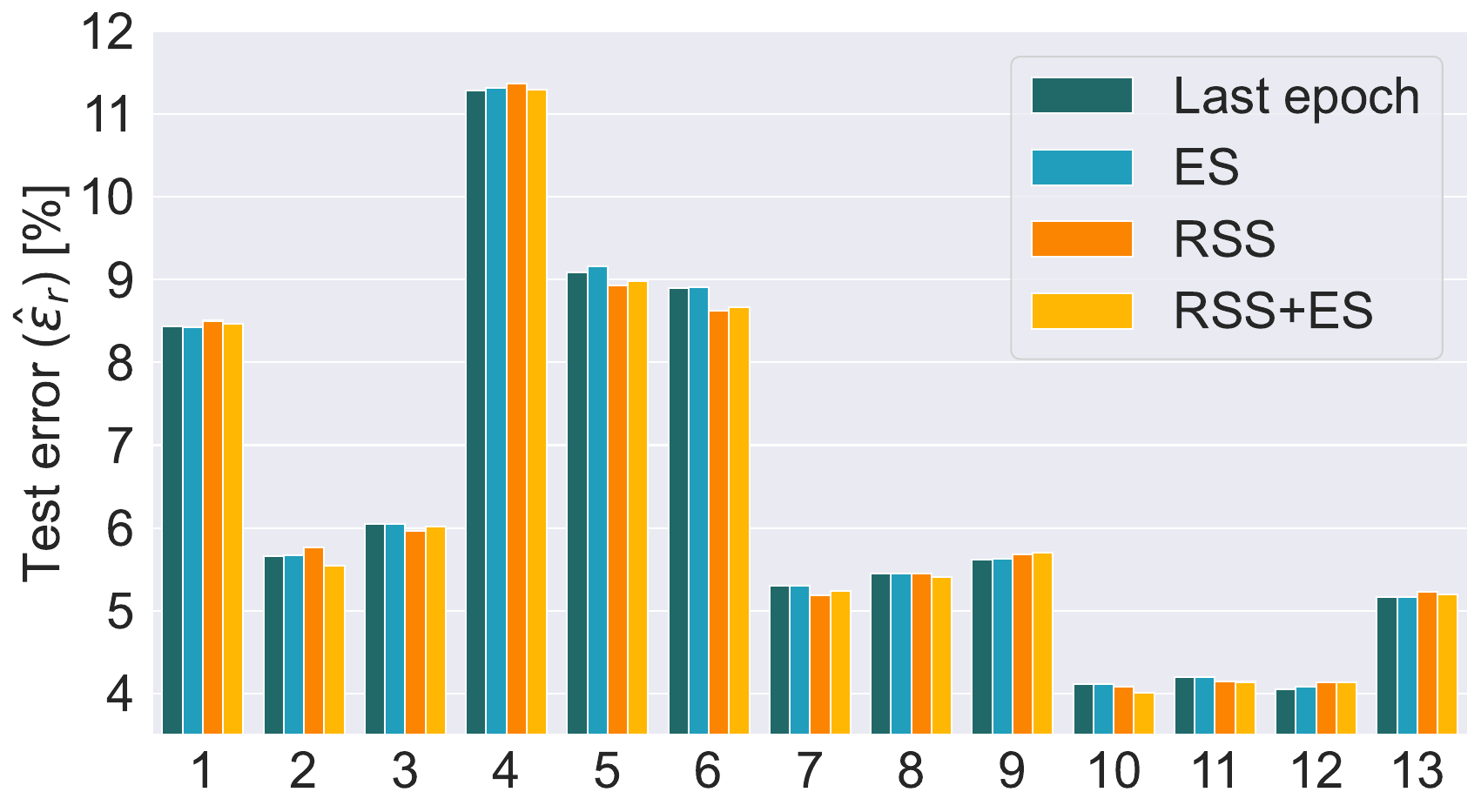} &
    \includegraphics[height=0.3\linewidth]{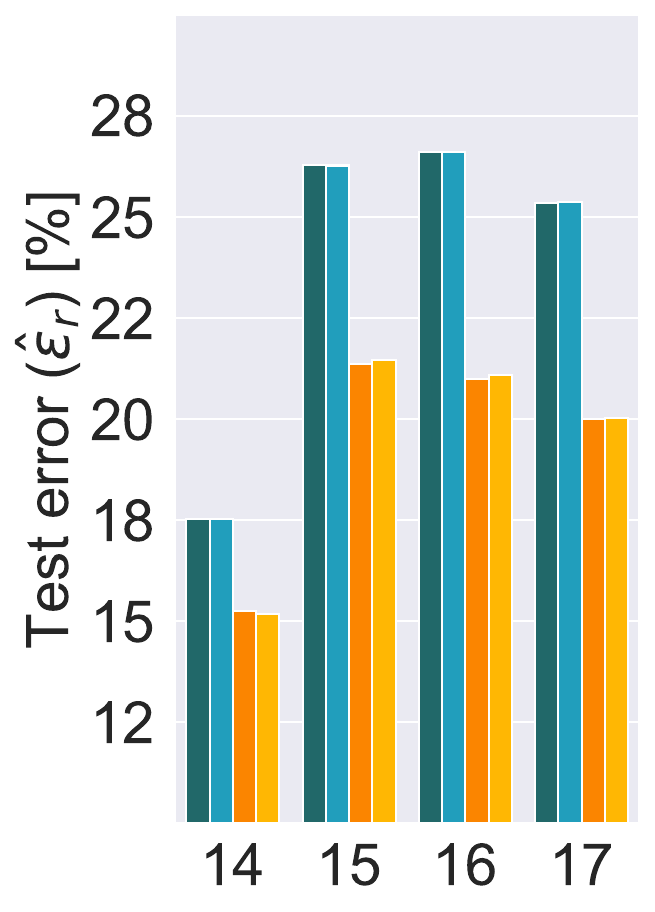} \\
    \rotatebox{90}{
        \quad\quad\quad\quad Train50K
    } &
    \includegraphics[height=0.3\linewidth]{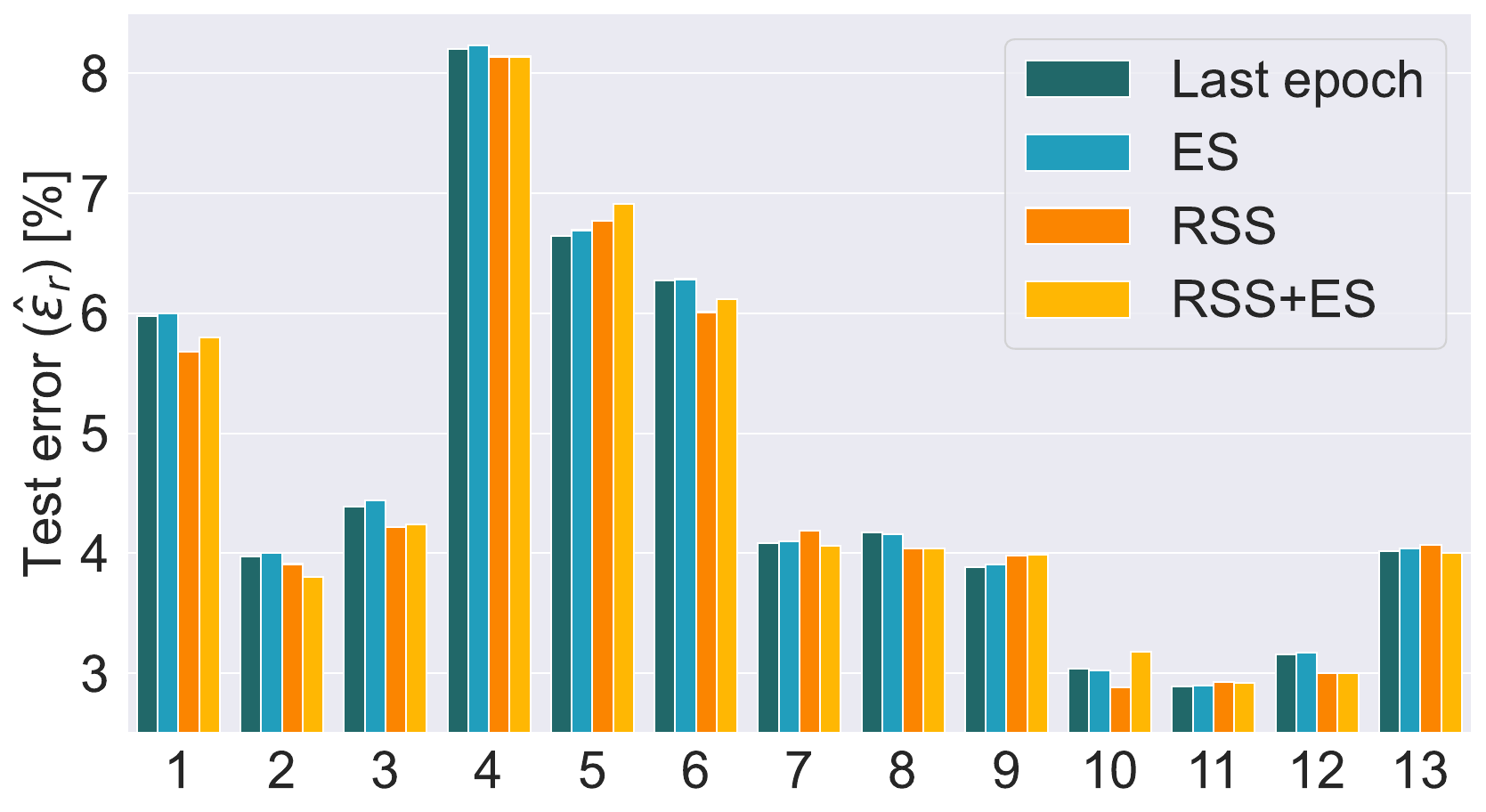} &
    \includegraphics[height=0.3\linewidth]{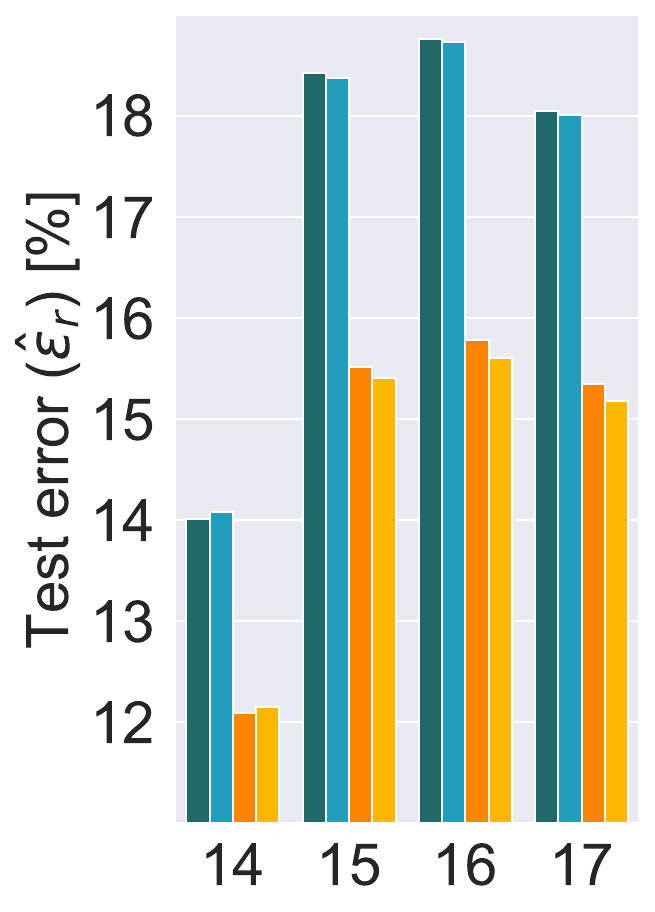} \\
\end{tabular}    


\caption{\textbf{ES and RSS for model selection (standard architectures) on all datasets:} Test errors of 17 architectures trained on each of the datasets. ``Last epoch'' and ES show the average error of the 10 models for each architecture. RSS and RSS+ES show the results of the selected model out of the 10 models.}
\label{fig:arch_es_rss}
\end{figure}

In addition to the results reported in~\ref{sec_es_rss_standard}, Figure~\ref{fig:arch_es_rss} shows results of ES, RSS and RSS+ES on all three datasets. 
RSS is beneficial for model selection in most cases, however the benefits decrease as the dataset size increases.


\newpage

\newpage

\section{Standard architecture convergence in training}
\label{sec:appendix_convergence_plots}
\begin{figure}[h]
\centering

\begin{subfigure}[b]{0.99\linewidth}
    \centering
    \includegraphics[height=0.25\linewidth, trim=160px 0 160px 0, clip=true]{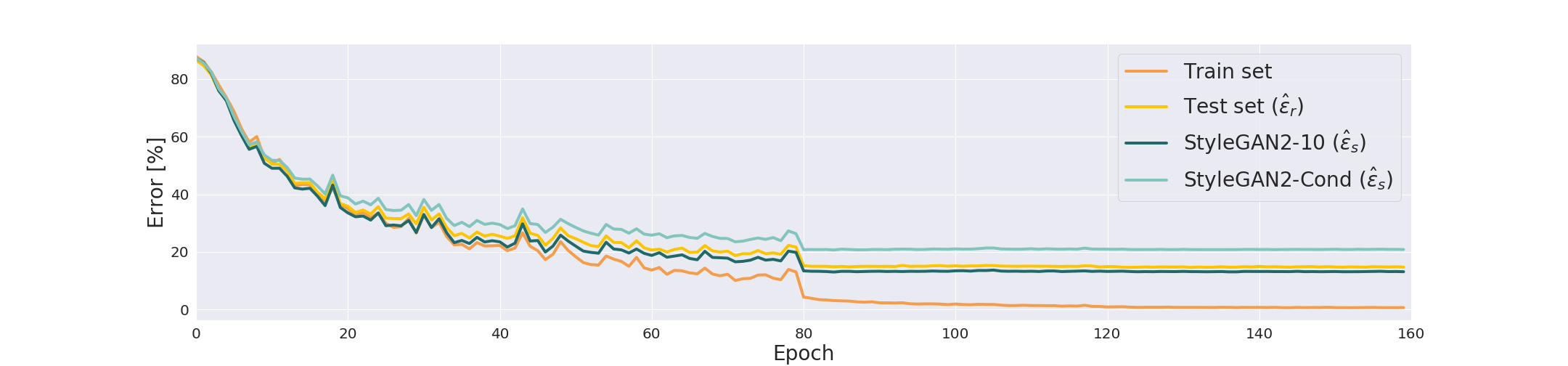}
    \caption{ResNet110 (Architecture 17 in Appendix~\ref{sec:appendix_models_description})}
    \label{fig:conv_resnet110}
\end{subfigure}
\begin{subfigure}[b]{0.99\linewidth}
    \centering
    \includegraphics[height=0.25\linewidth, trim=160px 0 160px 0, clip=true]{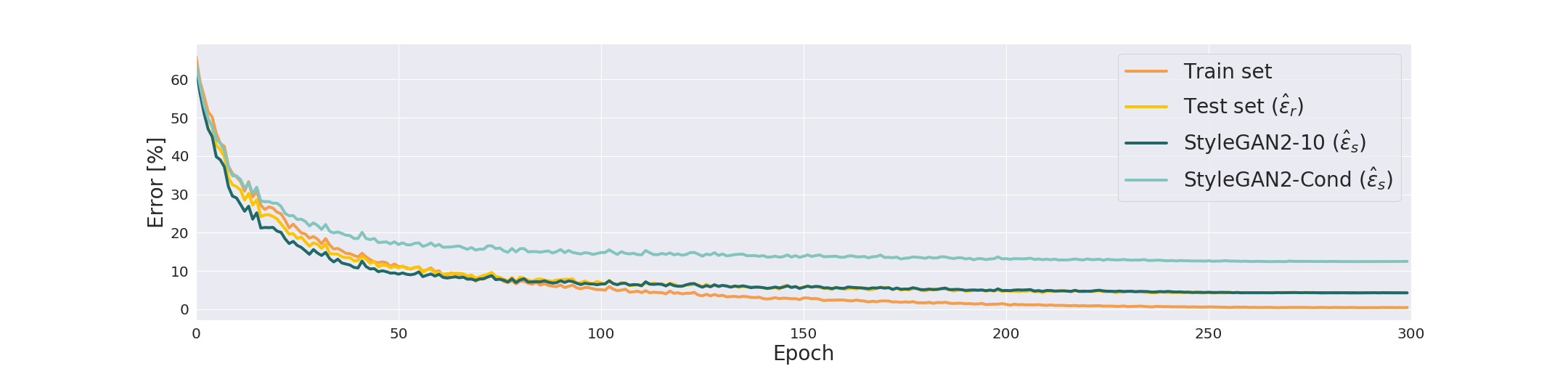} 
    \caption{Shake-Shake 64 + cutout (Architecture 10 in Appendix~\ref{sec:appendix_models_description})}
    \label{fig:conv_shake_shake_64}
\end{subfigure}



\caption{\textbf{Train, test and synthetic data error vs. epoch.}}
\label{fig:convergence_plots}
\end{figure}

Figure~\ref{fig:convergence_plots} shows two examples of the train, test and synthetic data errors vs. epoch index during training on the Train50K dataset. It can be observed that although the synthetic data error does not match the test error exactly, it follows the same trend as the test error.  
In the last epochs of training, where the learning rate has decreased there is very little change in the model's error.
This may explain why the early stopping experiments did not demonstrate any benefits.

\newpage

\section{Standard Architectures Description}
\label{sec:appendix_models_description}


Below is the list of architectures used in Sections~\ref{sec_rank_preservation} and~\ref{sec_es_rss_standard}:

\begin{enumerate}
    \item \textbf{DenseNet :} ~\citep{huang2017densely, huang2019convolutional} with batch size 32, initial learning rate 0.05, depth 100, block type ``bottleneck'', growth rate 12, compression rate 0.5.
    
    \item \textbf{PyramidNet 270:} ~\citep{DPRN} with depth 110, block type ``basic'', $\alpha = 270$.
    \item \textbf{PyramidNet 84:} ~\citep{DPRN} with depth 110, block type ``basic'', $\alpha = 84$.
    
    \item \textbf{SE-ResNet-preact:} ~\citep{hu2019squeezeandexcitation} with depth 110, se reduction=16.
    
    \item \textbf{ResNet-preact 110:} ~\citep{He2016} with depth 110, block type ``basic''.
    \item \textbf{ResNet-preact 164:} ~\citep{He2016} with depth 164, block type ``bottleneck''.
    
    \item \textbf{ResNext 4x64d:} ~\citep{Xie2016} with depth 29, cardinality 4, base channels 64, batch size 32 and initial learning rate 0.025.
    \item \textbf{ResNext 8x64d:} ~\citep{Xie2016} with depth 29, cardinality 8, base channels 64, batch size 64 and initial learning rate 0.05.

    \item \textbf{Shake-shake 32d:} ~\citep{Gastaldi17ShakeShake} with depth 26, base channels 32, S-S-I model.
    \item \textbf{Shake-shake 64d:} ~\citep{Gastaldi17ShakeShake} with depth 26, base channels 64, S-S-I model, batch size 64, base $lr=0.1$ .
    \item \textbf{Shake-shake 64d + cutout:} ~\citep{Gastaldi17ShakeShake} with depth 26, base channels 64, S-S-I model, batch size 64, $lr=0.1$, cosine scheduler, cutout ~\citep{devries2017improved} size 16.
    

    \item \textbf{Wide residual network + cutout:} ~\citep{Zagoruyko2016WRN} with depth 28, widening factor 10, base $lr=0.1$, batch size 64, cosine scheduler, cutout ~\citep{devries2017improved} size 16.
    \item \textbf{Wide residual network:} ~\citep{Zagoruyko2016WRN} with depth 28, widening factor 10.
    
    \item \textbf{ResNet 32:} ~\citep{He2016DeepRL} with depth 32, block type ``basic''.
    \item \textbf{ResNet 44:} ~\citep{He2016DeepRL} with depth 44, block type ``basic''.
    \item \textbf{ResNet 56:} ~\citep{He2016DeepRL} with depth 56, block type ``basic''.
    \item \textbf{ResNet 110:} ~\citep{He2016DeepRL} with depth 110, block type ``basic''.
    
\end{enumerate}



\newpage

\section{Synthetic Data Generation Details (CIFAR10)}
\label{sec:appendix_data_generation_description}
Our method for producing synthetic datasets is based on training GANs that in turn are used to generate the desired labeled data.
We consider two GAN frameworks for generating our synthetic datasets:
\begin{enumerate}
    \item StyleGAN2~\citep{Karras2019stylegan2} with non-leaking augmentation~\citep{karras2020training}. This framework is our best candidate for generating high quality synthetic datasets since it is the SOTA for generating CIFAR10 images. 
    \item WGAN-GP~\citep{NIPS2017_892c3b1c}. This framework generates lower quality images than StyleGAN2. 
    We consider it as a baseline to explore how the image quality impacts the datasets models selection capabilities. 
\end{enumerate}
For each GAN framework we consider two variants of training the GANs to generate labeled datasets:
\begin{enumerate}
    \item \textbf{Training 10 GANs (StyleGAN2-10/WGAN-GP-10):}
    For each of the 10 CIFAR10 classes, a different GAN was trained with just one class at a time (\eg, 5K images for Train50K, 3K images for Train30K and 1K images for Train10K). The generator instance with the best FID~\citep{heusel2017gans} score out of all instances obtained during training was selected to generate 10K images of its corresponding class.   
    \item \textbf{Training one Conditional GAN (StyleGAN2-Cond/WGAN-GP-Cond):}
    A single Conditional-GAN was trained, and best instance selected by FID score.
    Thereafter, 10K images were generated per class.
\end{enumerate}
Using the above methods we constructed 8 datasets (each with 100K labeled images): three ``StyleGAN2-10'' datasets and three ``StyleGAN2-Cond'' datasets (one per CIFAR10 subset), one ``WGAN-GP-10'' dataset and one ``WGAN-GP-Cond'' dataset (for the Train50K CIFAR10 subset).

Table~\ref{tab:table_of_synth_gen} shows the FID scores breakdown for our synthetic datasets. As expected, as the training dataset size decreases the FID score increases. 

\begin{figure*}
    \centering
    \begin{tabular}{ c  c  c  c }
     Real & Train50K & Train30K & Train10K \\
    \includegraphics[width=0.12\linewidth]{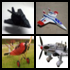} &
\includegraphics[width=0.12\linewidth]{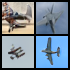} &
\includegraphics[width=0.12\linewidth]{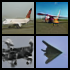} &
\includegraphics[width=0.12\linewidth]{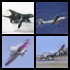} \\ 
    \includegraphics[width=0.12\linewidth]{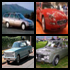} &
\includegraphics[width=0.12\linewidth]{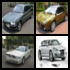} &
\includegraphics[width=0.12\linewidth]{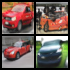} &
\includegraphics[width=0.12\linewidth]{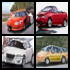} \\ 
    \includegraphics[width=0.12\linewidth]{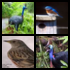} &
\includegraphics[width=0.12\linewidth]{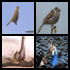} &
\includegraphics[width=0.12\linewidth]{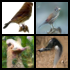} &
\includegraphics[width=0.12\linewidth]{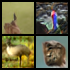} \\ 
    \includegraphics[width=0.12\linewidth]{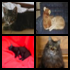} &
\includegraphics[width=0.12\linewidth]{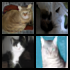} &
\includegraphics[width=0.12\linewidth]{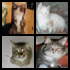} &
\includegraphics[width=0.12\linewidth]{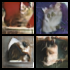} \\ 
    \includegraphics[width=0.12\linewidth]{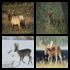} &
\includegraphics[width=0.12\linewidth]{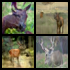} &
\includegraphics[width=0.12\linewidth]{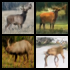} &
\includegraphics[width=0.12\linewidth]{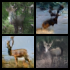} \\ 
    \includegraphics[width=0.12\linewidth]{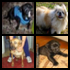} &
\includegraphics[width=0.12\linewidth]{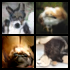} &
\includegraphics[width=0.12\linewidth]{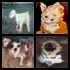} &
\includegraphics[width=0.12\linewidth]{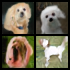} \\ 
    \includegraphics[width=0.12\linewidth]{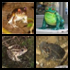} &
\includegraphics[width=0.12\linewidth]{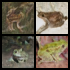} &
\includegraphics[width=0.12\linewidth]{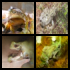} &
\includegraphics[width=0.12\linewidth]{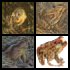} \\
    \includegraphics[width=0.12\linewidth]{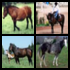} &
\includegraphics[width=0.12\linewidth]{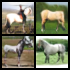} &
\includegraphics[width=0.12\linewidth]{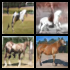} &
\includegraphics[width=0.12\linewidth]{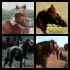} \\ 
    \includegraphics[width=0.12\linewidth]{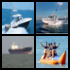} &
\includegraphics[width=0.12\linewidth]{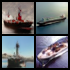} &
\includegraphics[width=0.12\linewidth]{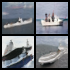} &
\includegraphics[width=0.12\linewidth]{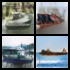} \\ 
    \includegraphics[width=0.12\linewidth]{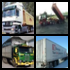} &
\includegraphics[width=0.12\linewidth]{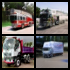} &
\includegraphics[width=0.12\linewidth]{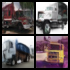} &
\includegraphics[width=0.12\linewidth]{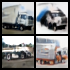} \\ 
  \end{tabular}
 
\caption{\textbf{Real images vs. StyleGAN2-10 datasets.}}
\label{fig:table_of_images_classes}
\end{figure*}

\begin{figure}
    \centering
    \begin{tabular}{ c  c  c  c }
     Real & Train50K & Train30K & Train10K \\
    \includegraphics[width=0.12\linewidth]{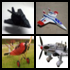} &
\includegraphics[width=0.12\linewidth]{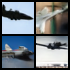} &
\includegraphics[width=0.12\linewidth]{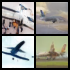} &
\includegraphics[width=0.12\linewidth]{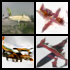} \\ 
    \includegraphics[width=0.12\linewidth]{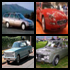} &
\includegraphics[width=0.12\linewidth]{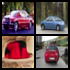} &
\includegraphics[width=0.12\linewidth]{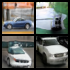} &
\includegraphics[width=0.12\linewidth]{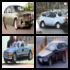} \\ 
    \includegraphics[width=0.12\linewidth]{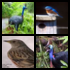} &
\includegraphics[width=0.12\linewidth]{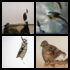} &
\includegraphics[width=0.12\linewidth]{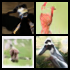} &
\includegraphics[width=0.12\linewidth]{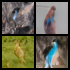} \\ 
    \includegraphics[width=0.12\linewidth]{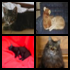} &
\includegraphics[width=0.12\linewidth]{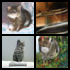} &
\includegraphics[width=0.12\linewidth]{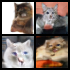} &
\includegraphics[width=0.12\linewidth]{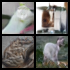} \\ 
    \includegraphics[width=0.12\linewidth]{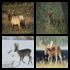} &
\includegraphics[width=0.12\linewidth]{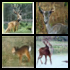} &
\includegraphics[width=0.12\linewidth]{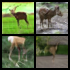} &
\includegraphics[width=0.12\linewidth]{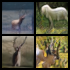} \\ 
    \includegraphics[width=0.12\linewidth]{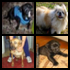} &
\includegraphics[width=0.12\linewidth]{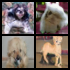} &
\includegraphics[width=0.12\linewidth]{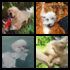} &
\includegraphics[width=0.12\linewidth]{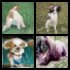} \\ 
    \includegraphics[width=0.12\linewidth]{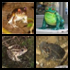} &
\includegraphics[width=0.12\linewidth]{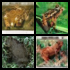} &
\includegraphics[width=0.12\linewidth]{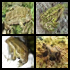} &
\includegraphics[width=0.12\linewidth]{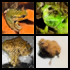} \\
    \includegraphics[width=0.12\linewidth]{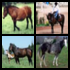} &
\includegraphics[width=0.12\linewidth]{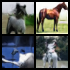} &
\includegraphics[width=0.12\linewidth]{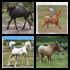} &
\includegraphics[width=0.12\linewidth]{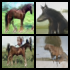} \\ 
    \includegraphics[width=0.12\linewidth]{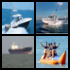} &
\includegraphics[width=0.12\linewidth]{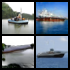} &
\includegraphics[width=0.12\linewidth]{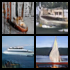} &
\includegraphics[width=0.12\linewidth]{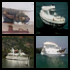} \\ 
    \includegraphics[width=0.12\linewidth]{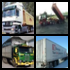} &
\includegraphics[width=0.12\linewidth]{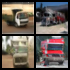} &
\includegraphics[width=0.12\linewidth]{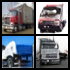} &
\includegraphics[width=0.12\linewidth]{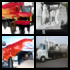} \\ 
  \end{tabular}
 
\caption{\textbf{Real images vs. StyleGAN2-Cond datasets.}}
\label{fig:table_of_images_cond}
\end{figure}

Figures~\ref{fig:table_of_images_classes} and~\ref{fig:table_of_images_cond} show samples of real CIFAR10 images and our synthetic StyleGAN2-based datasets for each of the CIAFR10 classes.

\begin{table}[h]
\caption{\textbf{FID scores ($\downarrow$) breakdown.}}
\vskip 0.15in
\label{tab:table_of_synth_gen}
\renewcommand{\arraystretch}{1.2}
\begin{center}
    \begin{tabular}{ l | c | c | c | c }
    Synth dataset & Class & Train50K & Train30K & Train10K \\
    \hline
    \multirow{10}{*}{StyleGAN2-10} 
    & 0 & 10.11 & 17.06 & 44.15 \\ \cline{2-5}
    & 1 & 6.05 & 9.41 & 29.91 \\ \cline{2-5}
    & 2 & 10.65 & 16.39 & 49.79 \\ \cline{2-5}
    & 3 & 12.04 & 18.58 & 56.67 \\ \cline{2-5}
    & 4 & 7.94 & 12.5 & 35.76\\ \cline{2-5}
    & 5 & 11.23 & 16.98 & 51.15\\ \cline{2-5}
    & 6 & 8.36 & 13.22 & 39.84\\ \cline{2-5}
    & 7 & 8.41 & 12.91 & 31.57\\ \cline{2-5}
    & 8 & 7.59 & 11.02 & 32.2 \\ \cline{2-5}
    & 9 & 6.25 & 10.26 & 28.33\\ \cline{2-5}
    & All & 4.4 & 4.86 & 14.15\\ \hline
    StyleGAN2-Cond & All & 4.4 & 6.25 & 11.72\\ \hline
    WGAN-GP-10 & All & 35.7 & N/A & N/A \\ \hline
    WGAN-GP-Cond & All & 27.3 & N/A & N/A \\
  \end{tabular}
\end{center}
\end{table}

\newpage

\section{Generated Synthetic ImageNet Samples}
\label{sec:appendix_imagenet_generated}
\begin{figure}[H]
    \centering
    \setlength{\tabcolsep}{2pt}
    \begin{tabular}{ c c  c  c  c  c }
     Real & BigGan & DiT (cfg=1) & DiT (cfg=2) & DiT (cfg=3) & DiT (cfg=4) \\
     \includegraphics[width=0.153\linewidth]{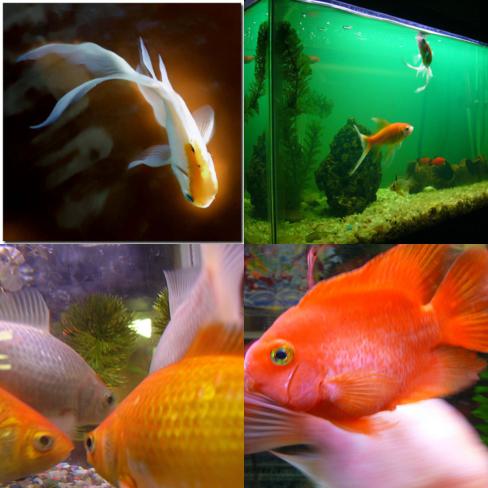}
     & \includegraphics[width=0.153\linewidth]{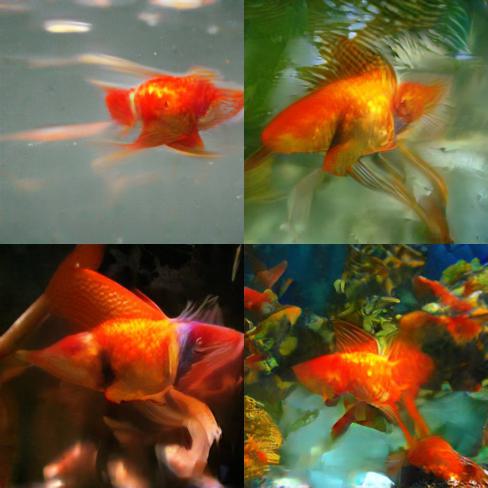}
     & \includegraphics[width=0.153\linewidth]{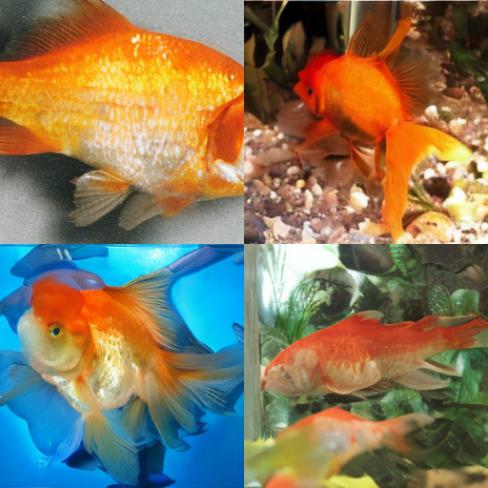}
     & \includegraphics[width=0.153\linewidth]{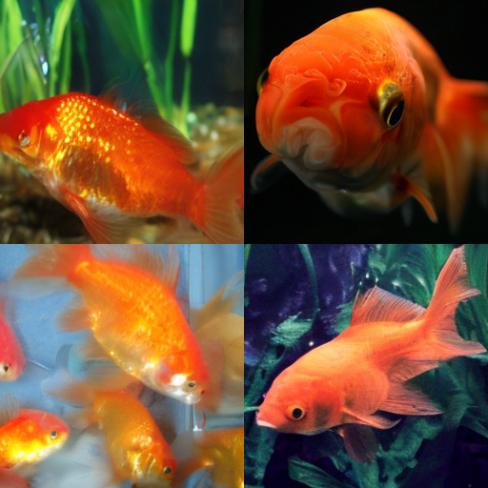}
     & \includegraphics[width=0.153\linewidth]{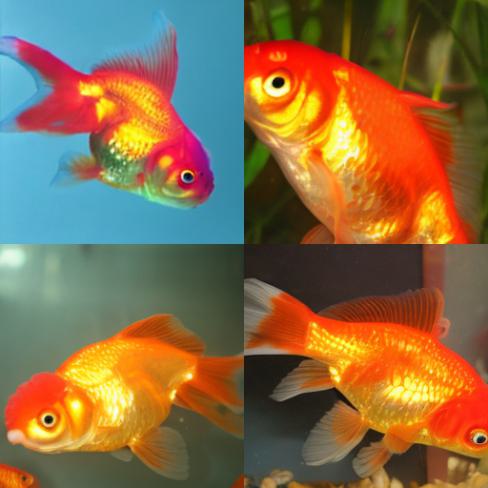}
     & \includegraphics[width=0.153\linewidth]{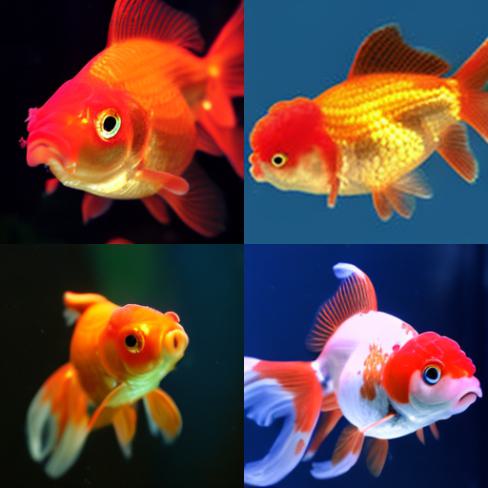} \\
     
     
     \includegraphics[width=0.153\linewidth]{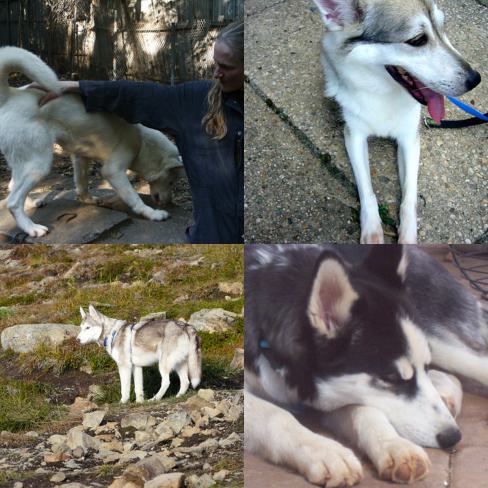}
     & \includegraphics[width=0.153\linewidth]{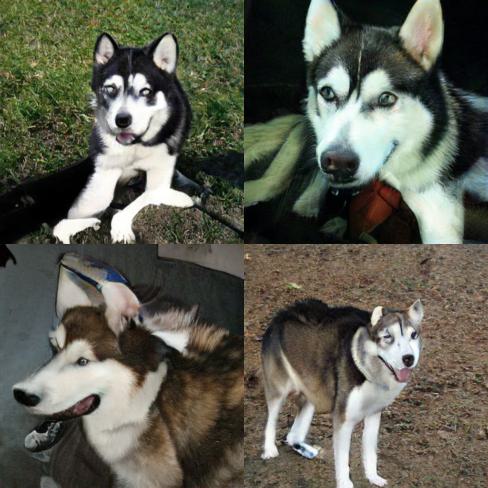}
     & \includegraphics[width=0.153\linewidth]{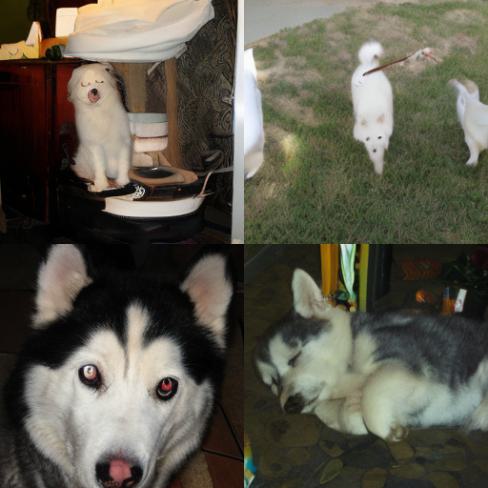}
     & \includegraphics[width=0.153\linewidth]{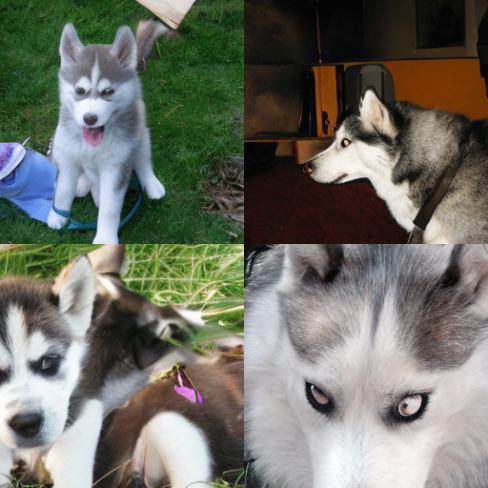}
     & \includegraphics[width=0.153\linewidth]{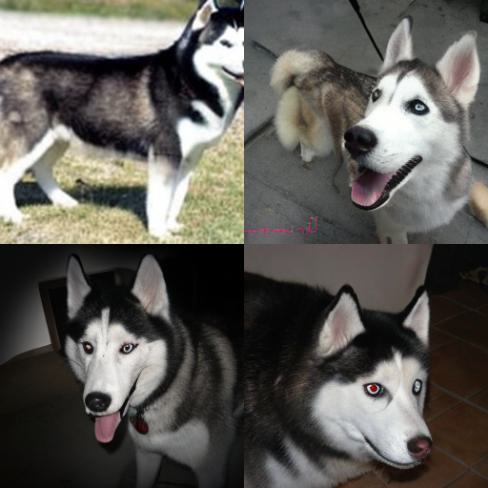}
     & \includegraphics[width=0.153\linewidth]{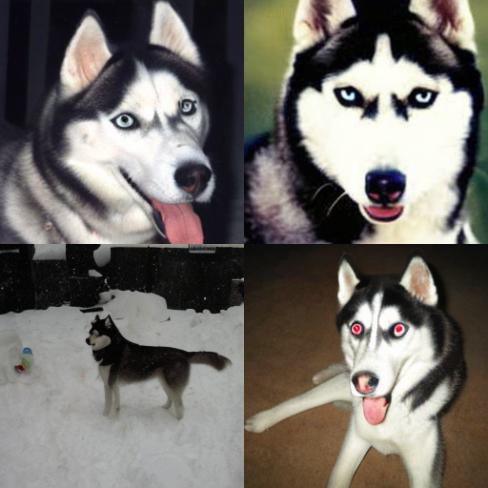} \\
     
     \includegraphics[width=0.153\linewidth]{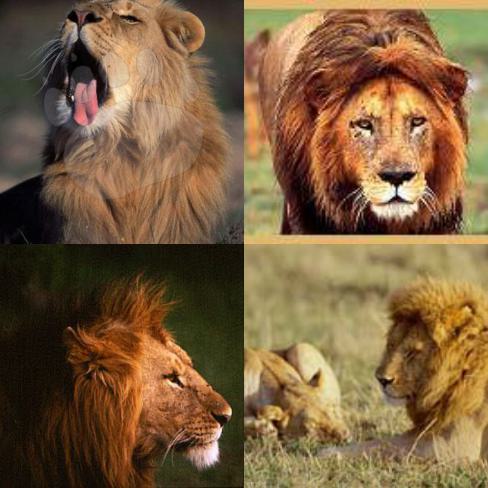}
     & \includegraphics[width=0.153\linewidth]{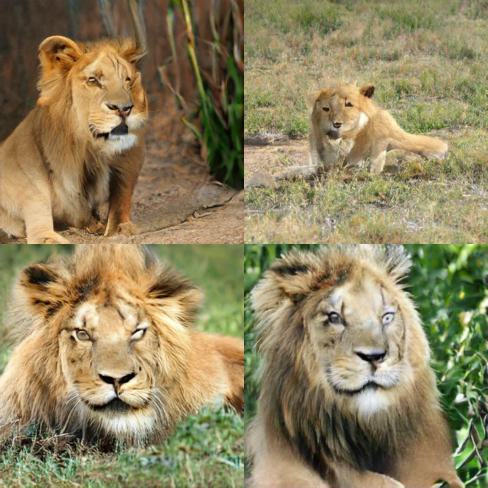}
     & \includegraphics[width=0.153\linewidth]{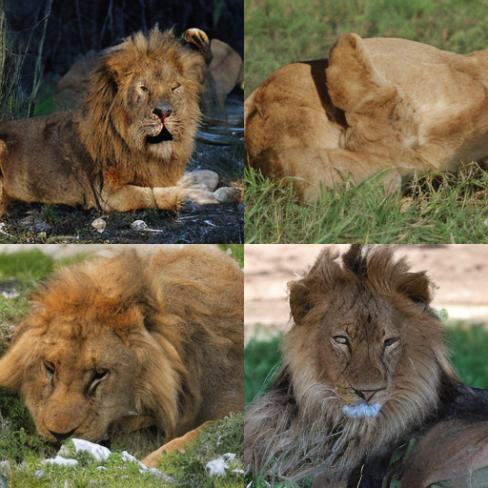}
     & \includegraphics[width=0.153\linewidth]{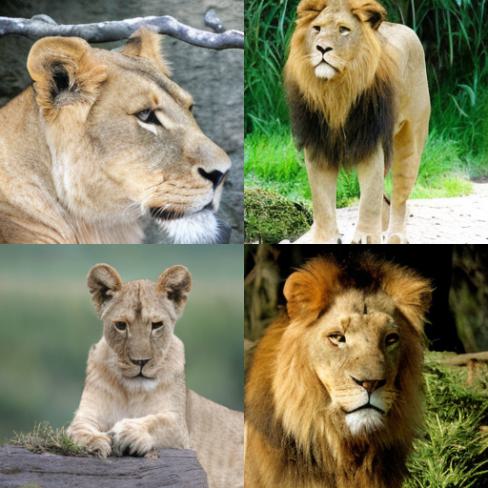}
     & \includegraphics[width=0.153\linewidth]{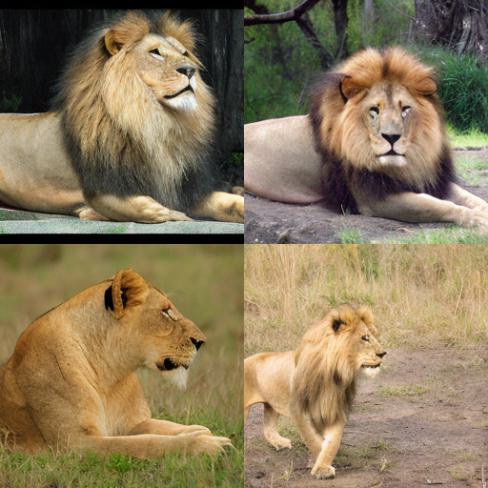}
     & \includegraphics[width=0.153\linewidth]{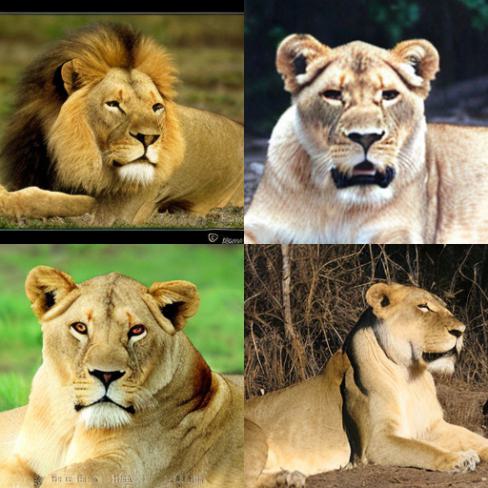} \\
     
     
     \includegraphics[width=0.153\linewidth]{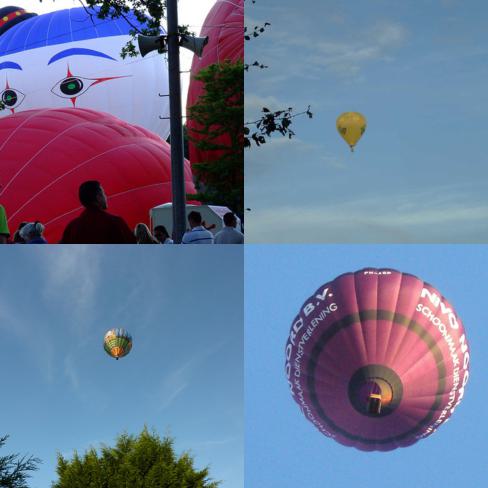}
     & \includegraphics[width=0.153\linewidth]{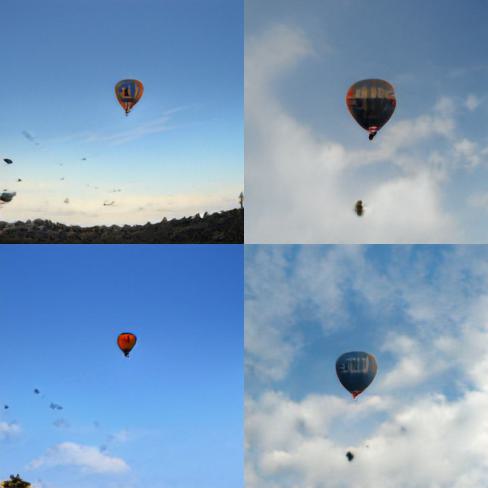}
     & \includegraphics[width=0.153\linewidth]{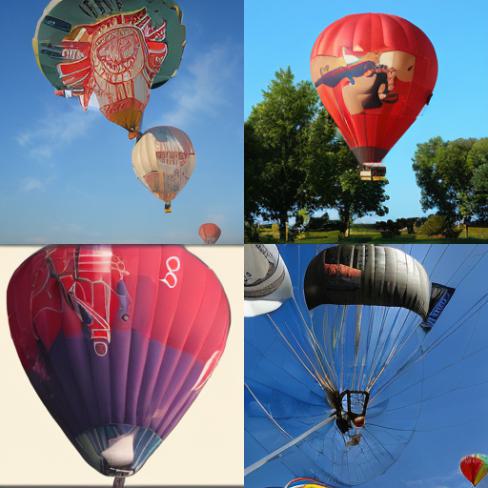}
     & \includegraphics[width=0.153\linewidth]{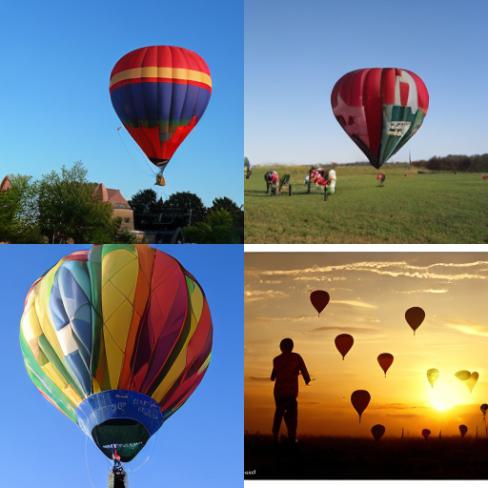}
     & \includegraphics[width=0.153\linewidth]{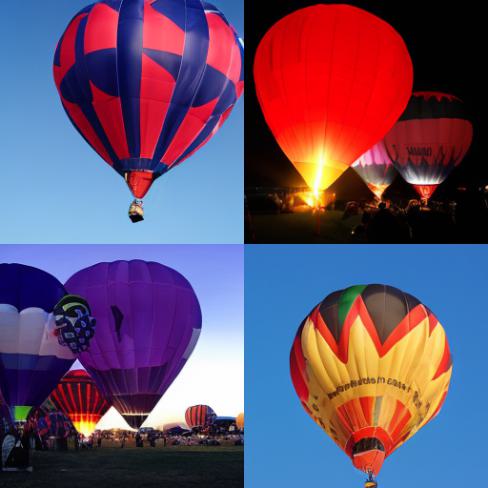}
     & \includegraphics[width=0.153\linewidth]{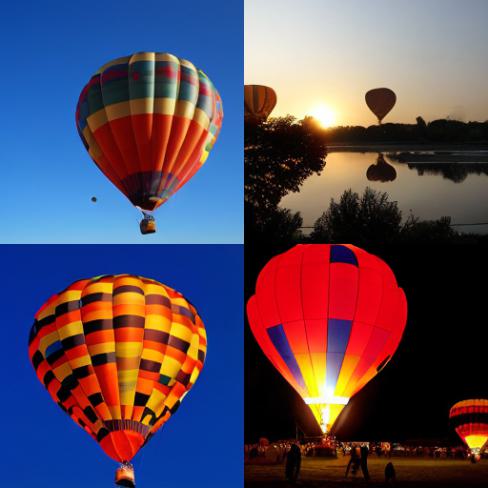} \\
     
     \includegraphics[width=0.153\linewidth]{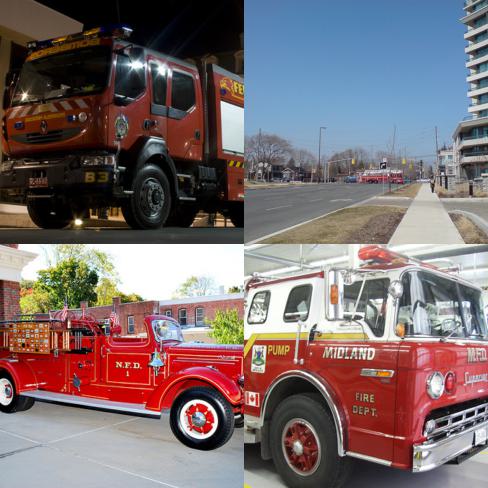}
     & \includegraphics[width=0.153\linewidth]{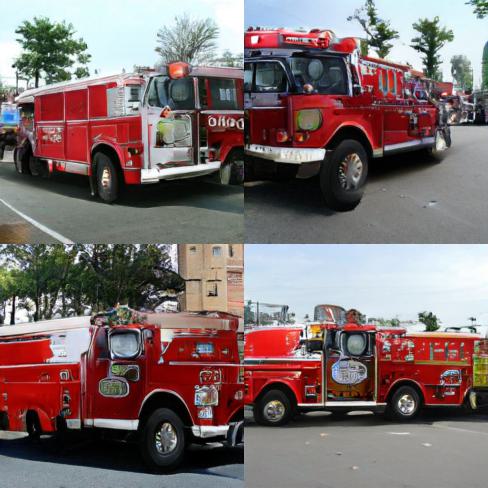}
     & \includegraphics[width=0.153\linewidth]{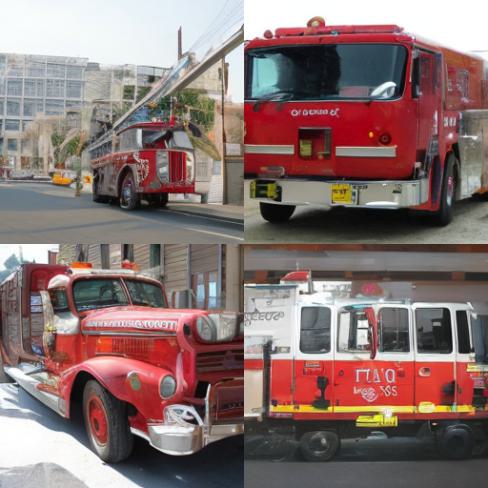}
     & \includegraphics[width=0.153\linewidth]{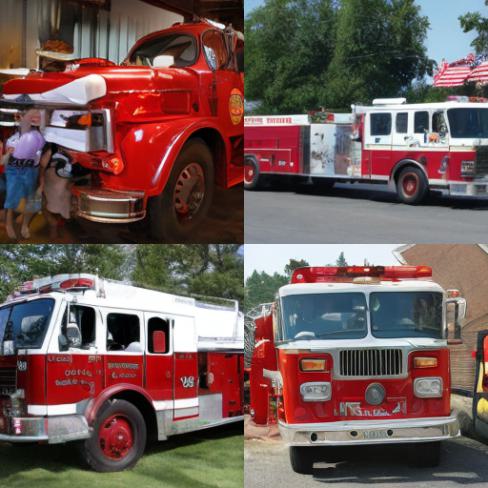}
     & \includegraphics[width=0.153\linewidth]{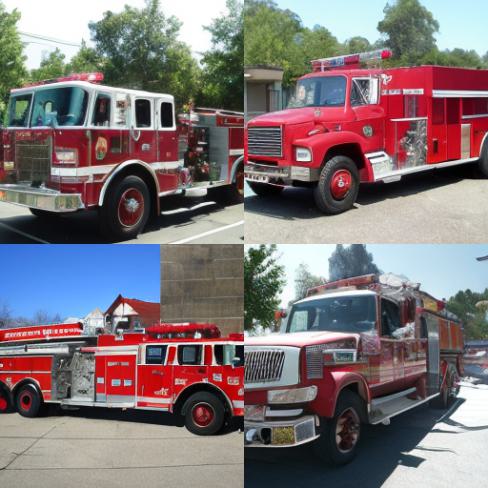}
     & \includegraphics[width=0.153\linewidth]{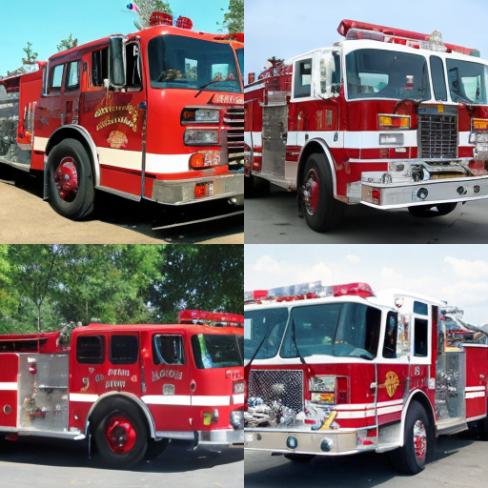} \\
     
     \includegraphics[width=0.153\linewidth]{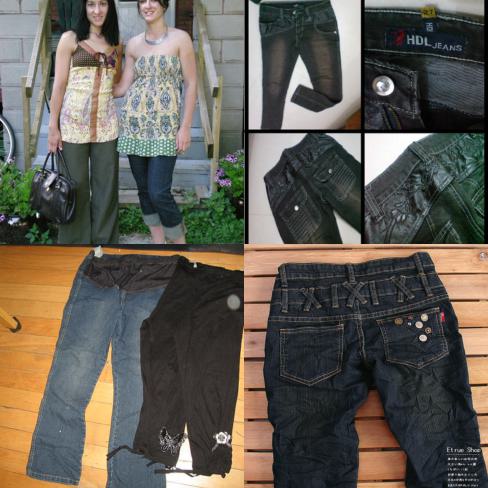}
     & \includegraphics[width=0.153\linewidth]{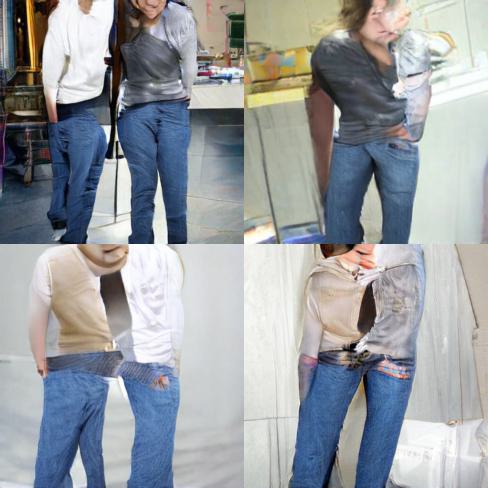}
     & \includegraphics[width=0.153\linewidth]{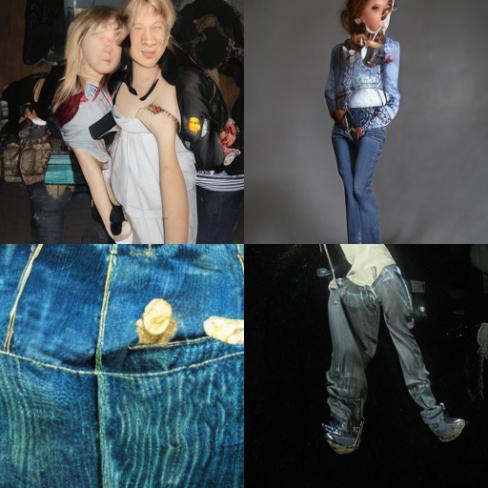}
     & \includegraphics[width=0.153\linewidth]{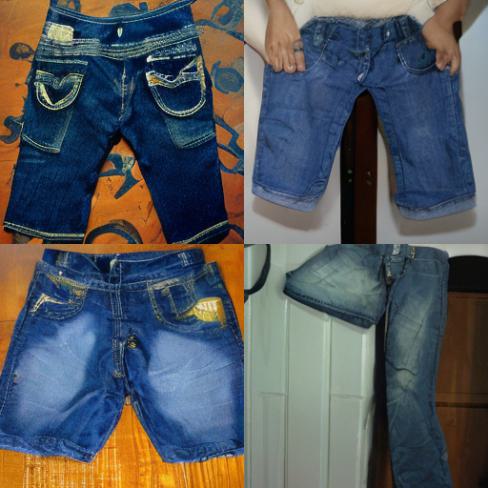}
     & \includegraphics[width=0.153\linewidth]{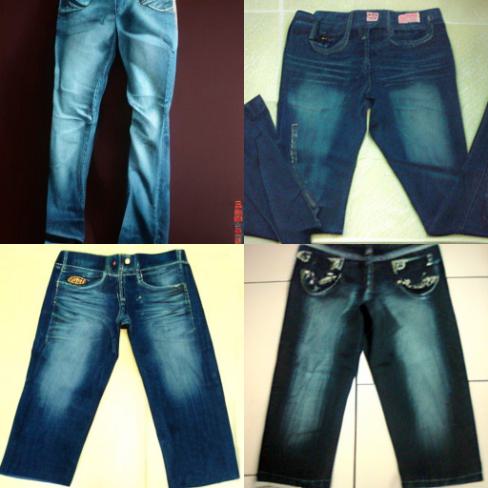}
     & \includegraphics[width=0.153\linewidth]{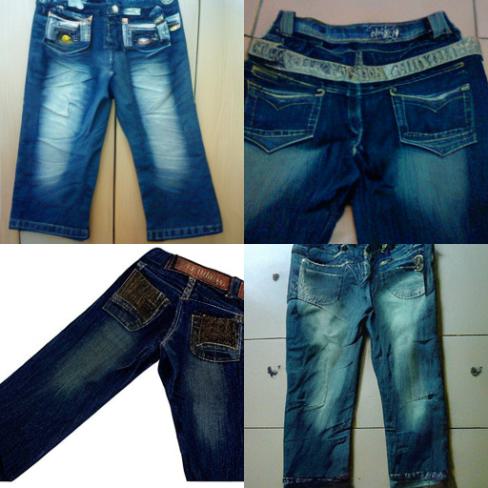} \\
     
     \includegraphics[width=0.153\linewidth]{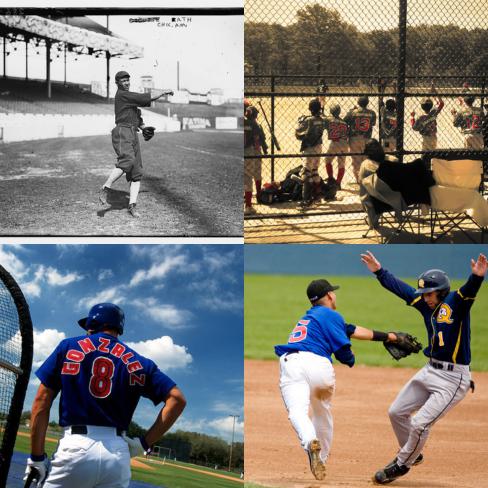}
     & \includegraphics[width=0.153\linewidth]{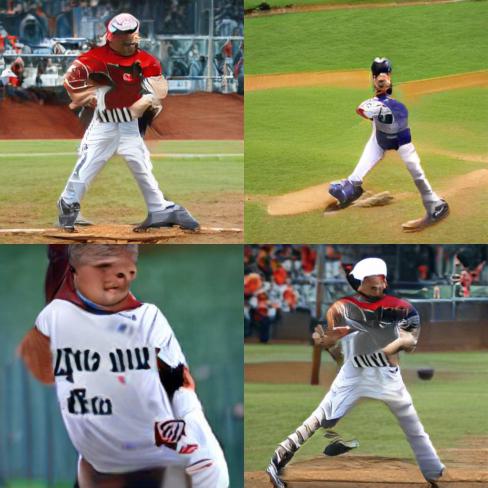}
     & \includegraphics[width=0.153\linewidth]{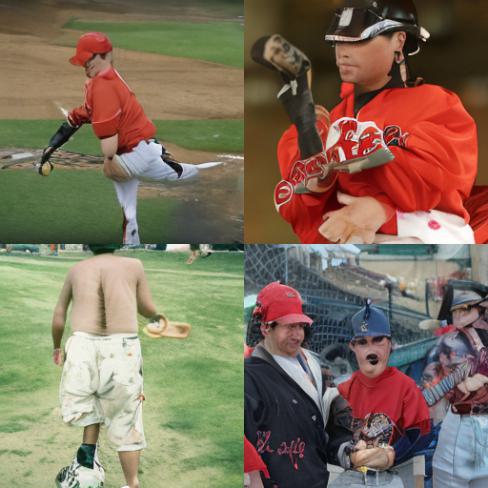}
     & \includegraphics[width=0.153\linewidth]{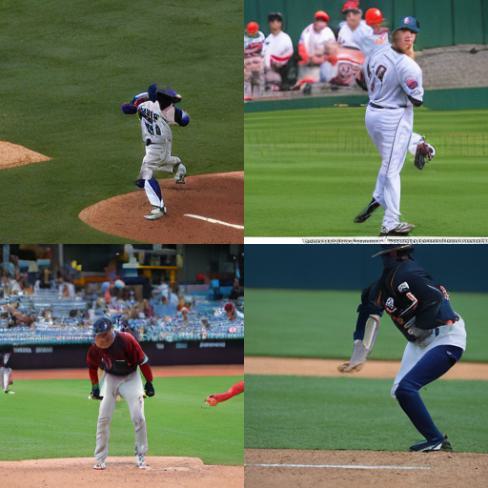}
     & \includegraphics[width=0.153\linewidth]{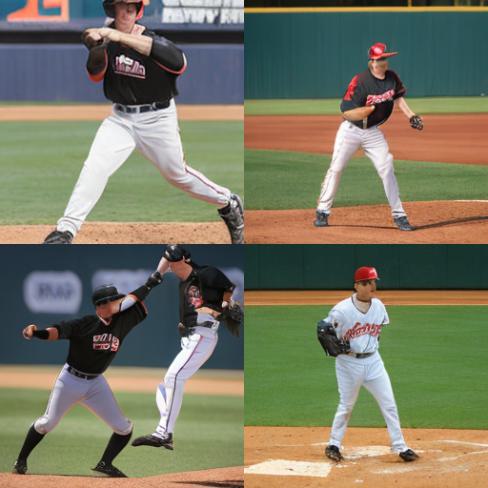}
     & \includegraphics[width=0.153\linewidth]{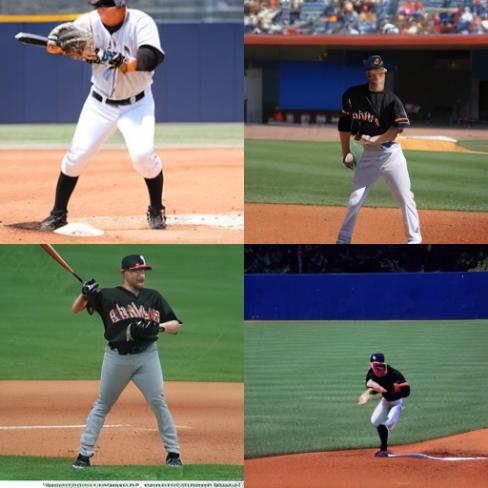} \\

    \end{tabular}
 
\caption{\textbf{ImageNet generated samples.} Rows show the following classes: Goldfish (1), 
Siberian husky (250), Lion (291), 
Balloon (555), Fire truck (555), Denim (608), Baseball player (981).}
\label{fig:fig_imagenet_samples}
\end{figure}

Figure~\ref{fig:fig_imagenet_samples} shows randomly sampled images from our five synthetic datasets (BigGAN, DiT with cfg $=1/2/3$ or $4$) for the classes: Goldfish (1), Siberian husky (250), Lion (291), Balloon (555), Fire truck (555), Denim (608), Baseball player (981).

\newpage

\section{Calibrated Synthetic ImageNet Samples}
\label{sec:appendix_imagenet_calibrated}
\begin{figure}[H]
    \footnotesize
    \centering
    \setlength{\tabcolsep}{2pt}
    \begin{tabular}{ c  c  c  c  c  c  c}
    \multicolumn{2}{c}{Highest positive $w$} & \multicolumn{2}{c}{Lowest negative $w$} & \multicolumn{2}{c}{$w=0.0$} & $w$ Distribution \\
    \midrule
    $w=0.048$ & $w=0.044$ & $w=-0.088$ & $w=-0.009$ & $w=0.0$ & $w=0.0$ & \\
    \rotatebox{90}{\hspace{2mm}\tiny{Class 21, DiT (cfg=4) \textcolor{white}{p}}}
    \includegraphics[width=0.133\linewidth]{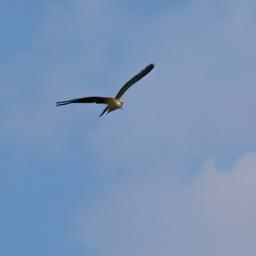} &
    \includegraphics[width=0.133\linewidth]{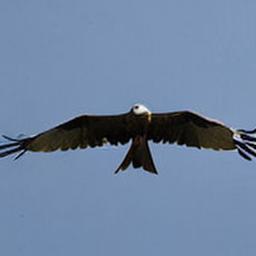} &
    \includegraphics[width=0.133\linewidth]{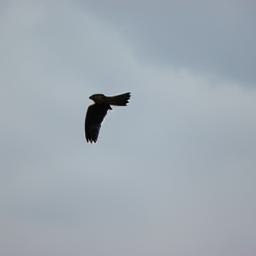} &
    \includegraphics[width=0.133\linewidth]{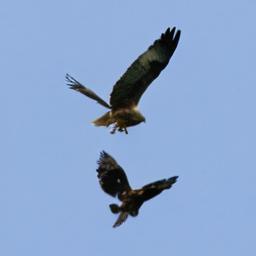} &
    \includegraphics[width=0.133\linewidth]{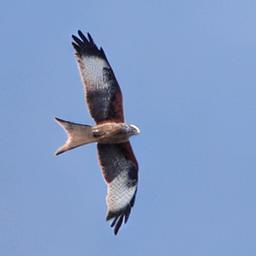} &
    \includegraphics[width=0.133\linewidth]{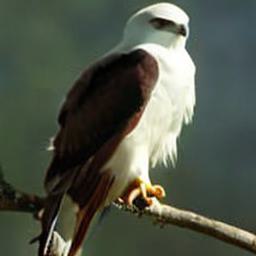} &
    \includegraphics[width=0.133\linewidth]{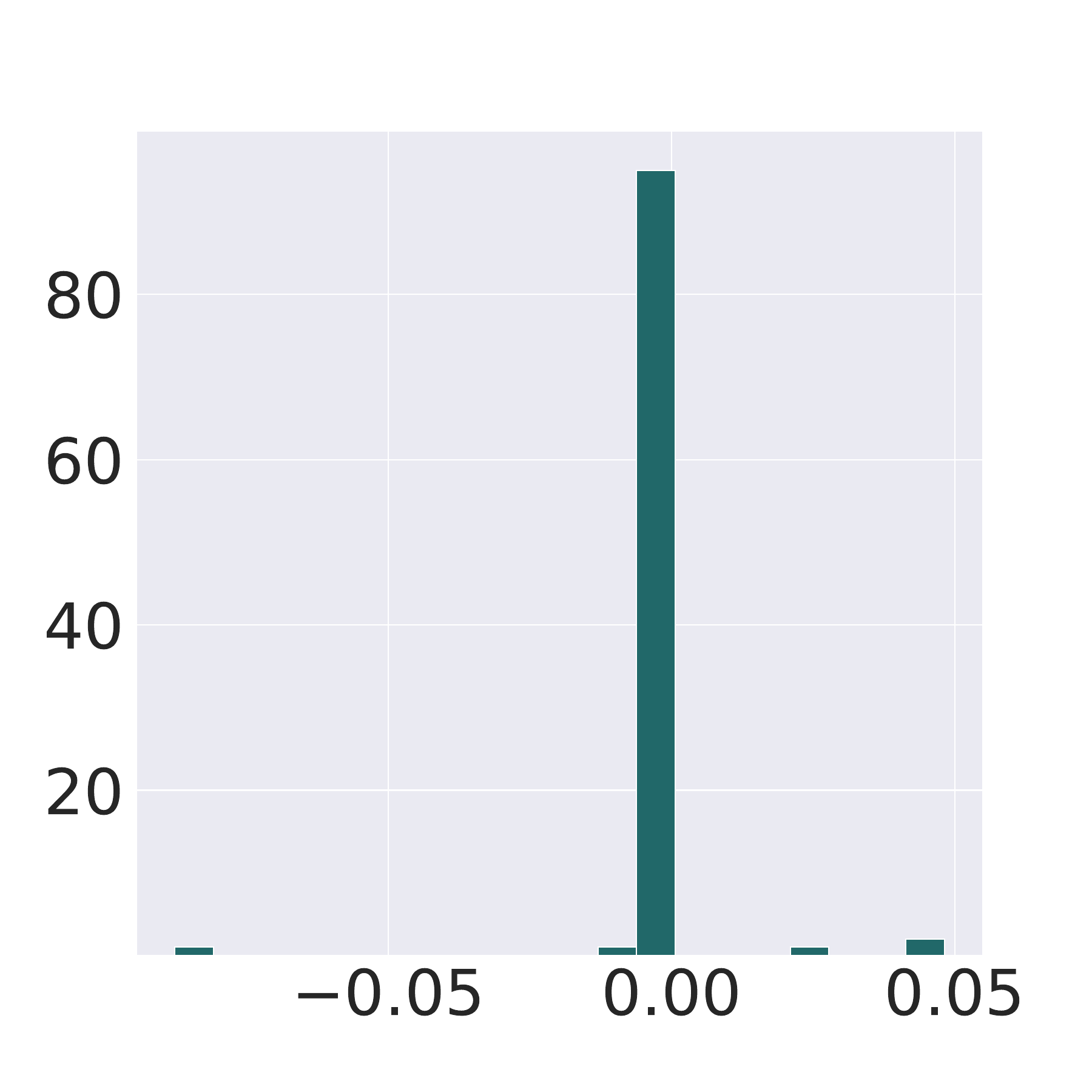} \\
    
    $w=0.050$ & $w=0.049$ & $w=-0.030$ & $w=-0.029$ & $w=0.0$ & $w=0.0$ & \\
    \rotatebox{90}{\hspace{2mm}\tiny{Class 21, DiT (cfg=1) \textcolor{white}{p}}}
    \includegraphics[width=0.133\linewidth]{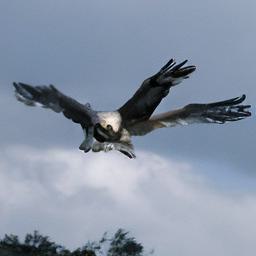} &
    \includegraphics[width=0.133\linewidth]{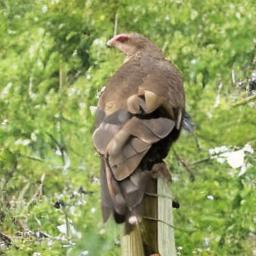} &
    \includegraphics[width=0.133\linewidth]{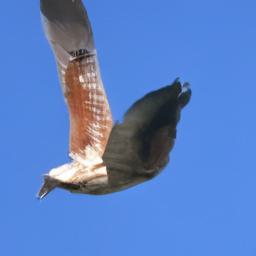} &
    \includegraphics[width=0.133\linewidth]{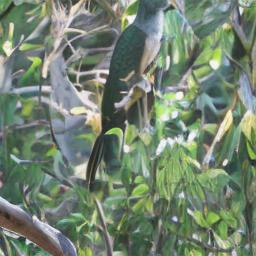} &
    \includegraphics[width=0.133\linewidth]{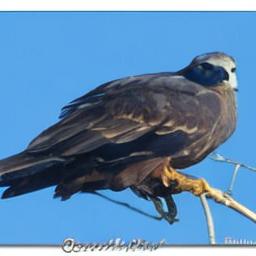} &
    \includegraphics[width=0.133\linewidth]{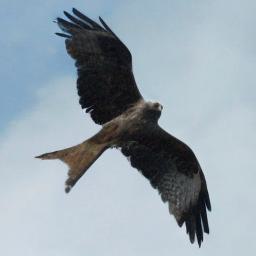} &
    \includegraphics[width=0.133\linewidth]{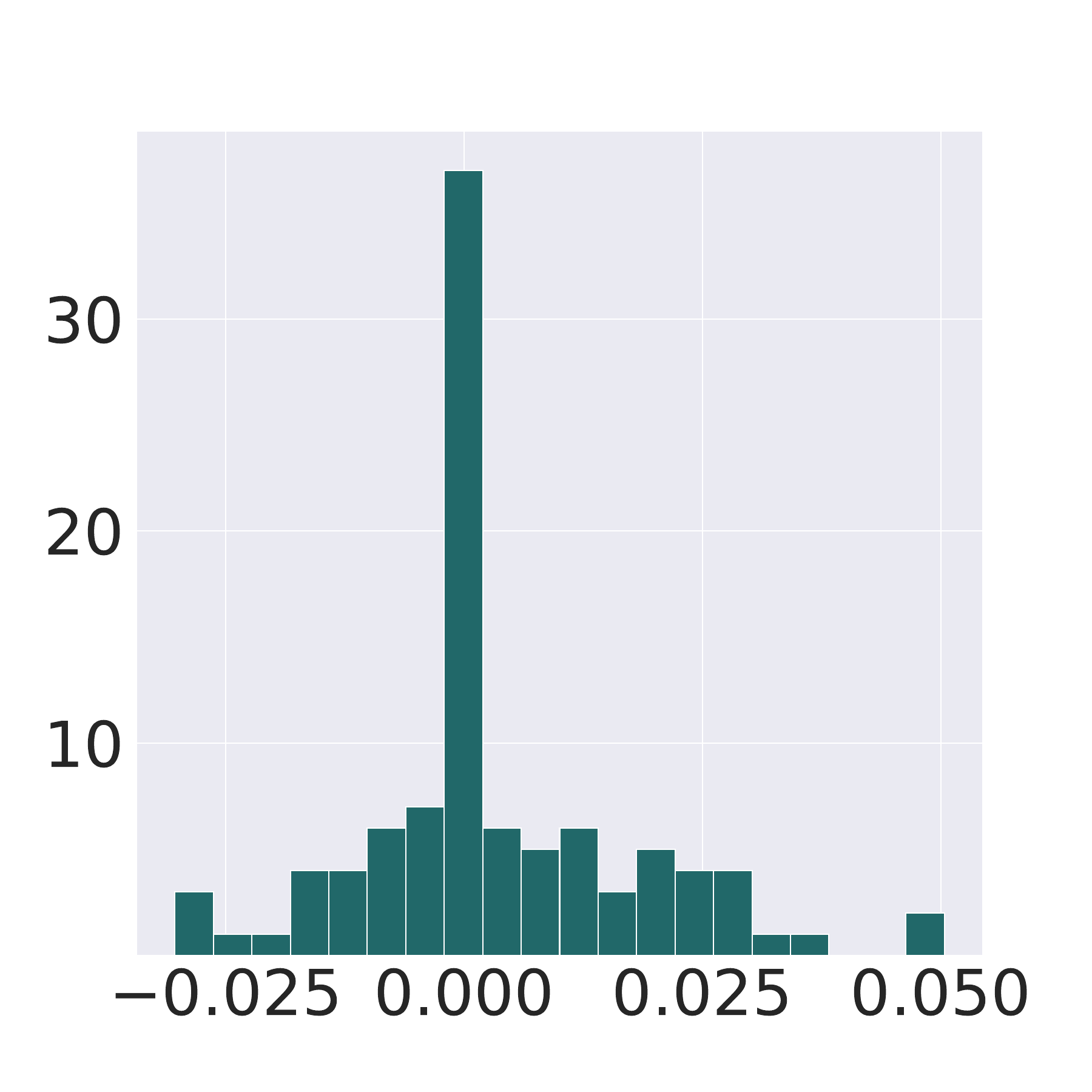} \\
    
    $w=0.040$ & $w=0.032$ & $w=-0.027$ & $w=-0.025$ & $w=0.0$ & $w=0.0$ & \\
    \rotatebox{90}{\hspace{2mm}\tiny{Class 21, BigGAN \textcolor{white}{p}}}
    \includegraphics[width=0.133\linewidth]{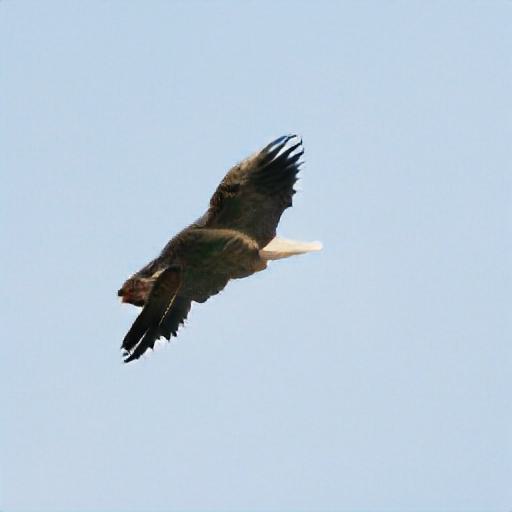} &
    \includegraphics[width=0.133\linewidth]{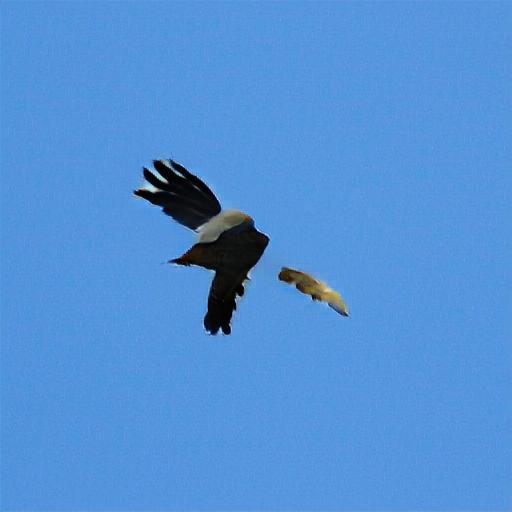} &
    \includegraphics[width=0.133\linewidth]{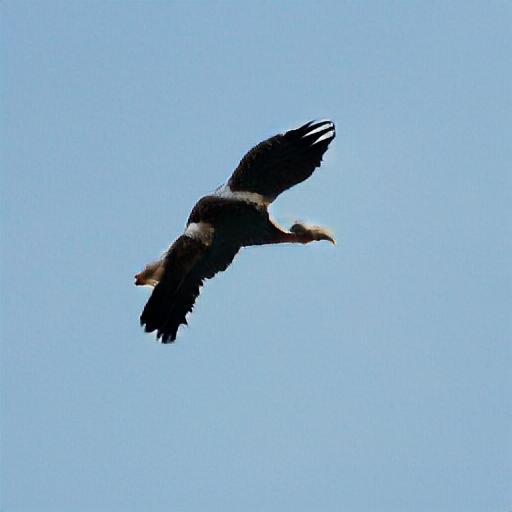} &
    \includegraphics[width=0.133\linewidth]{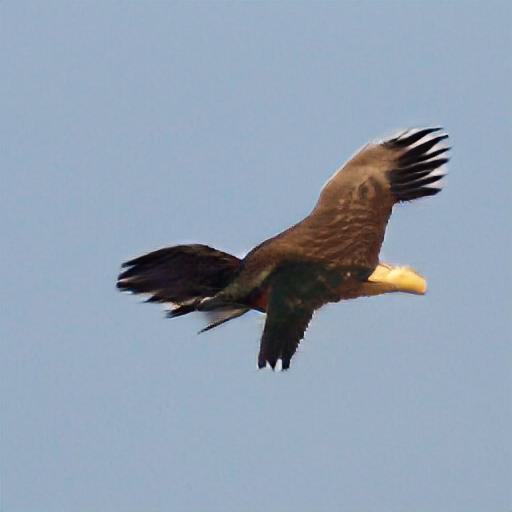} &
    \includegraphics[width=0.133\linewidth]{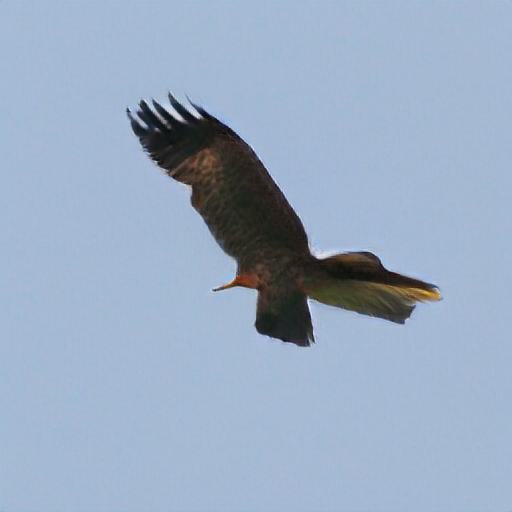} &
    \includegraphics[width=0.133\linewidth]{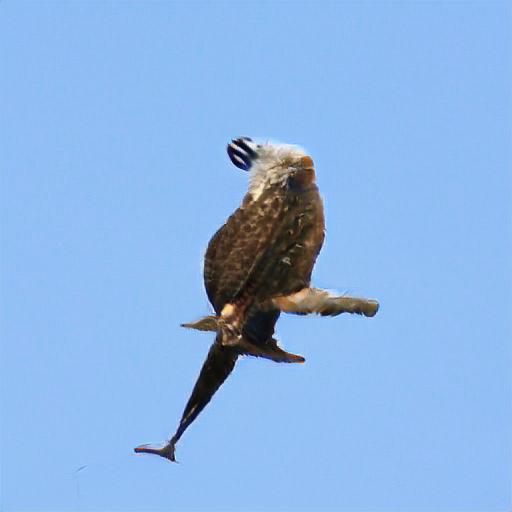} &
    \includegraphics[width=0.133\linewidth]{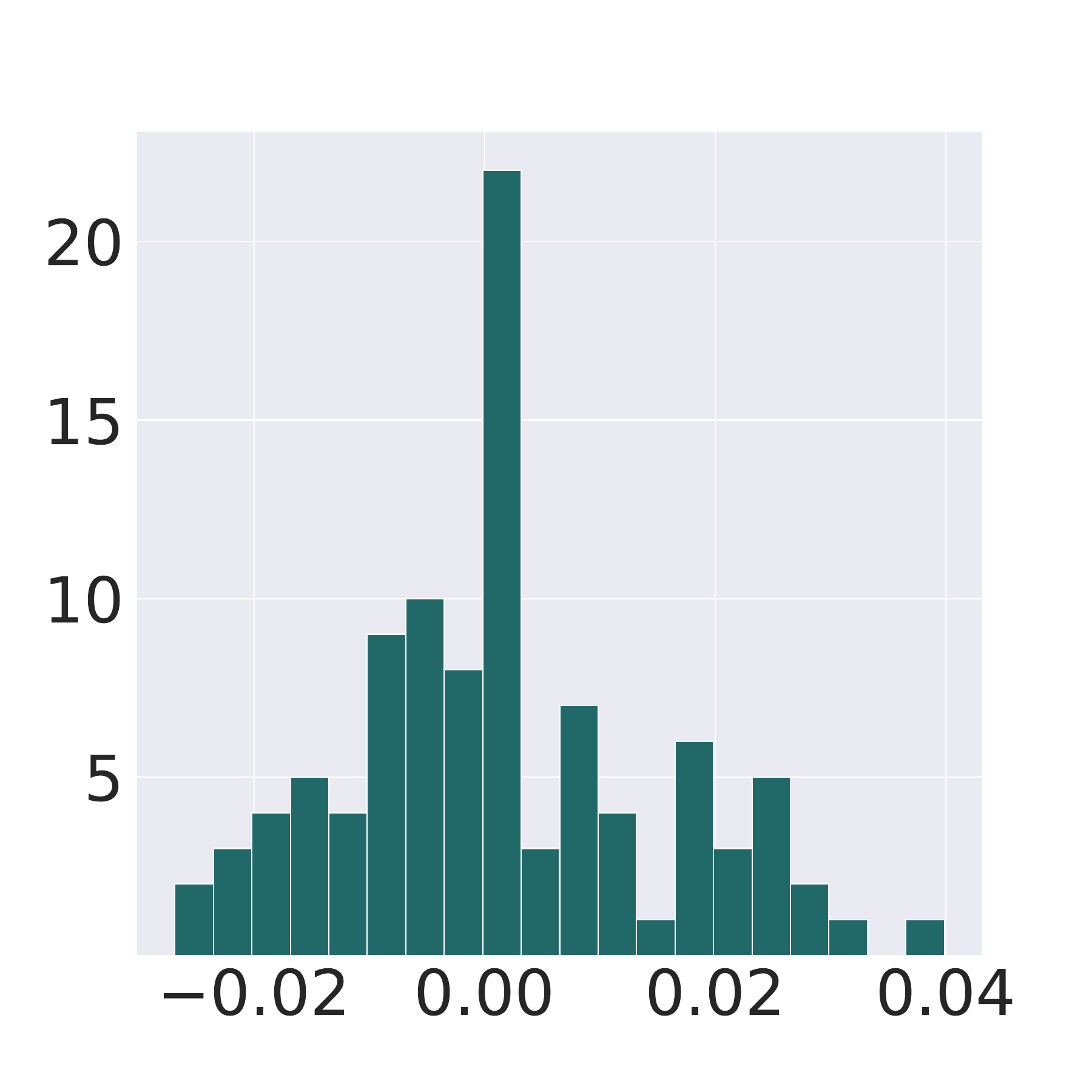} \\
    
    $w=0.088$ & $w=0.083$ & $w=-0.0289$ & $w=-0.024$ & $w=0.0$ & $w=0.0$ & \\
    \rotatebox{90}{\hspace{2mm}\tiny{Class 883, DiT (cfg=4) \textcolor{white}{p}}}
    \includegraphics[width=0.133\linewidth]{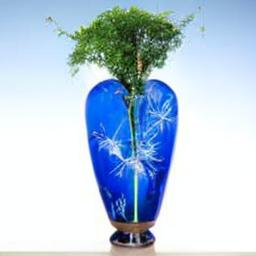} &
    \includegraphics[width=0.133\linewidth]{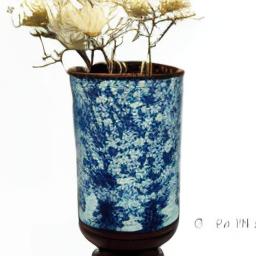} &
    \includegraphics[width=0.133\linewidth]{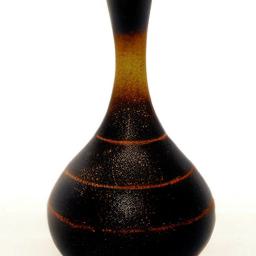} &
    \includegraphics[width=0.133\linewidth]{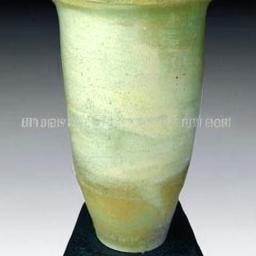} &
    \includegraphics[width=0.133\linewidth]{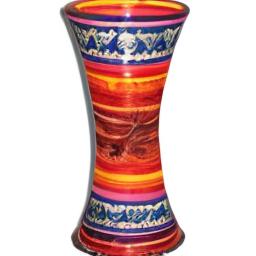} &
    \includegraphics[width=0.133\linewidth]{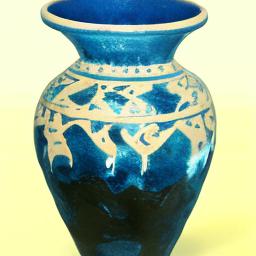} &
    \includegraphics[width=0.133\linewidth]{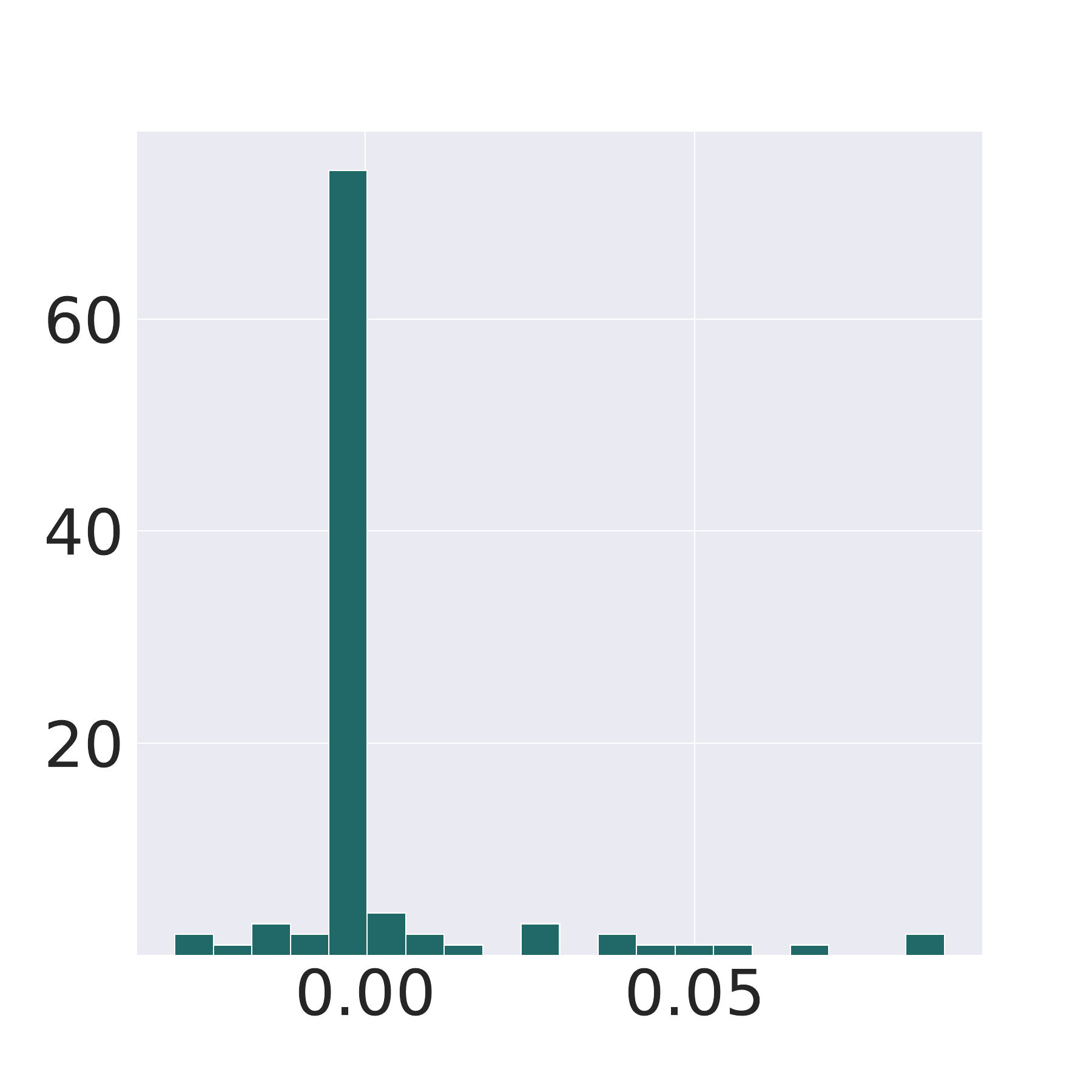} \\
    
    $w=0.041$ & $w=0.038$ & $w=-0.038$ & $w=-0.037$ & $w=0.0$ & $w=0.0$ & \\
    \rotatebox{90}{\hspace{2mm}\tiny{Class 883, DiT (cfg=1) \textcolor{white}{p}}}
    \includegraphics[width=0.133\linewidth]{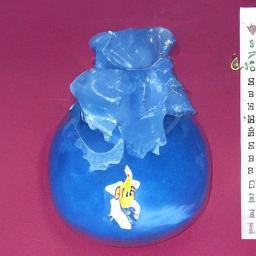} &
    \includegraphics[width=0.133\linewidth]{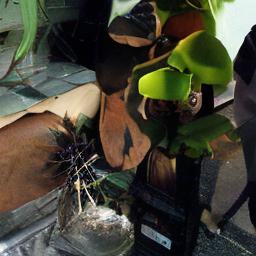} &
    \includegraphics[width=0.133\linewidth]{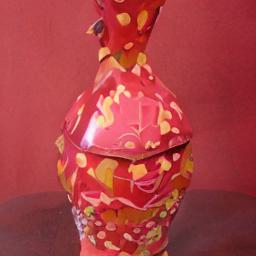} &
    \includegraphics[width=0.133\linewidth]{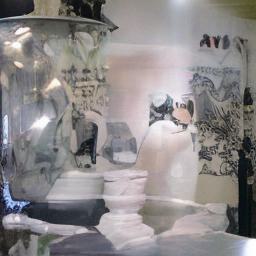} &
    \includegraphics[width=0.133\linewidth]{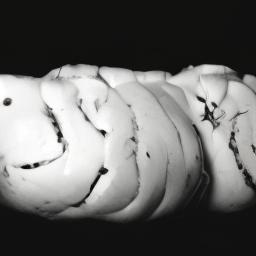} &
    \includegraphics[width=0.133\linewidth]{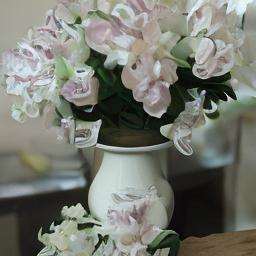} &
    \includegraphics[width=0.133\linewidth]{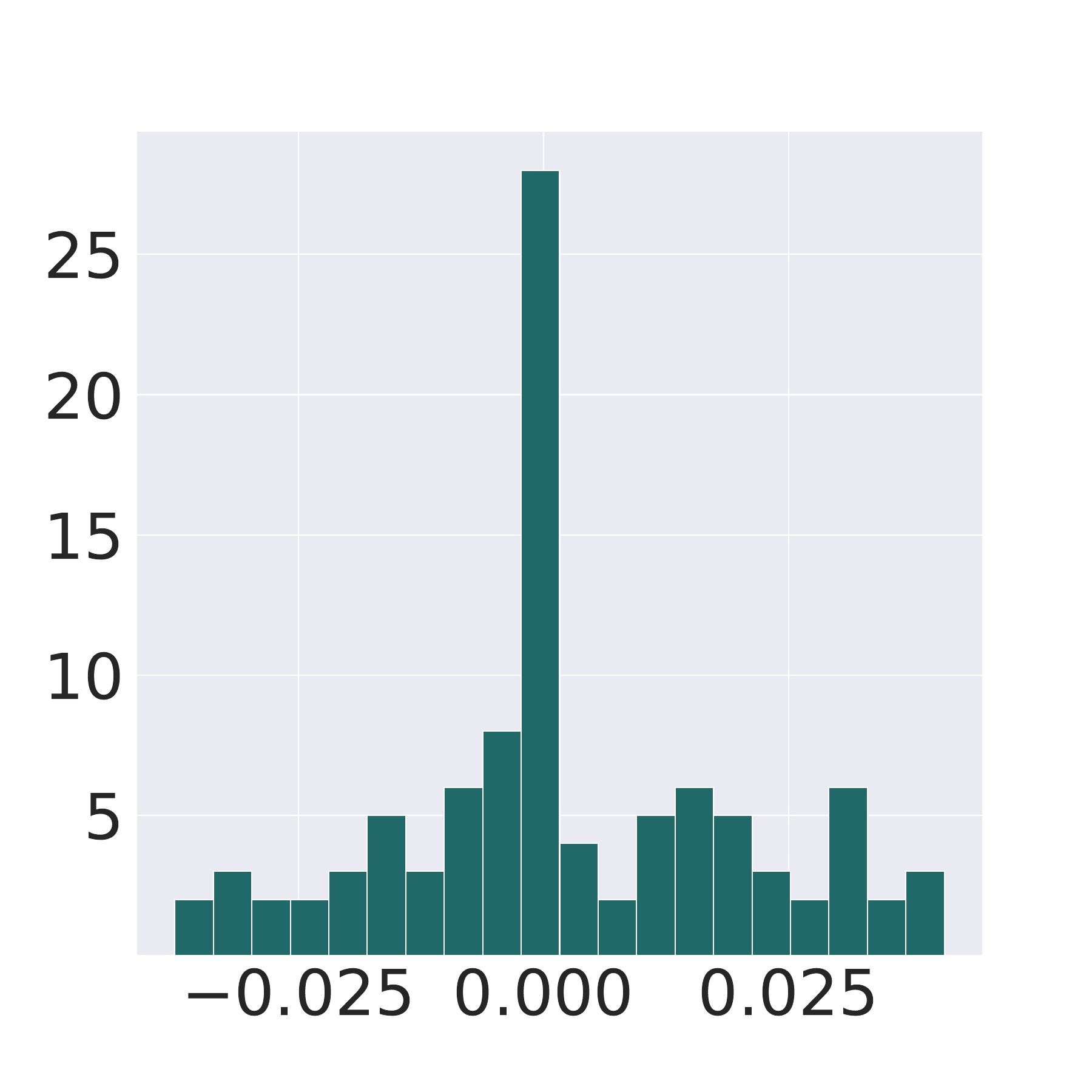} \\
    
    $w=0.082$ & $w=0.058$ & $w=-0.039$ & $w=-0.036$ & $w=0.0$ & $w=0.0$ & \\
    \rotatebox{90}{\hspace{2mm}\tiny{Class 883, BigGAN \textcolor{white}{p}}}
    \includegraphics[width=0.133\linewidth]{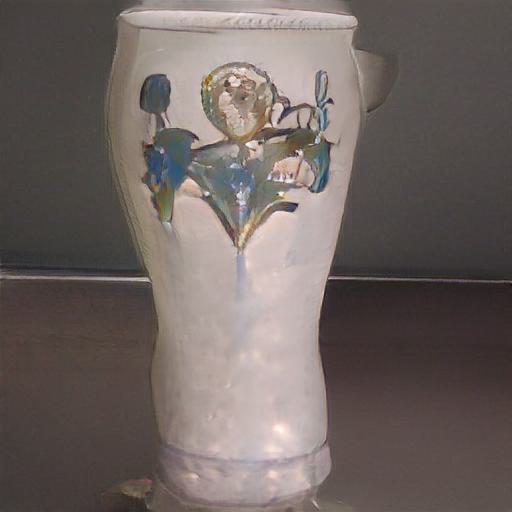} &
    \includegraphics[width=0.133\linewidth]{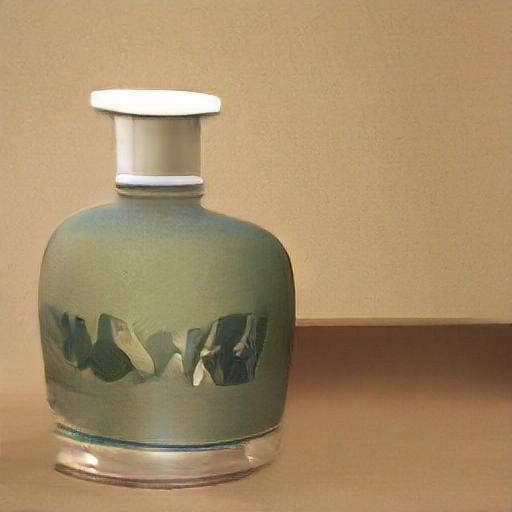} &
    \includegraphics[width=0.133\linewidth]{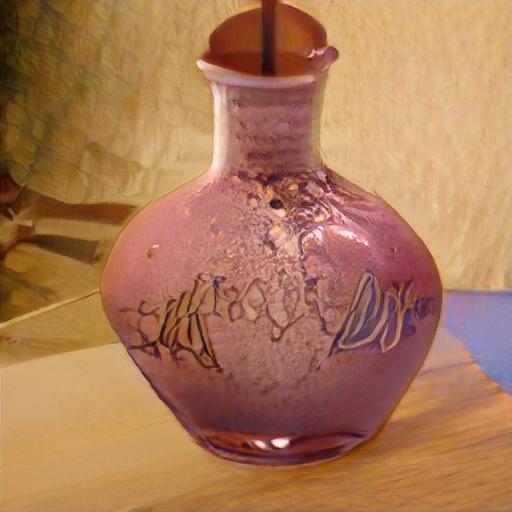} &
    \includegraphics[width=0.133\linewidth]{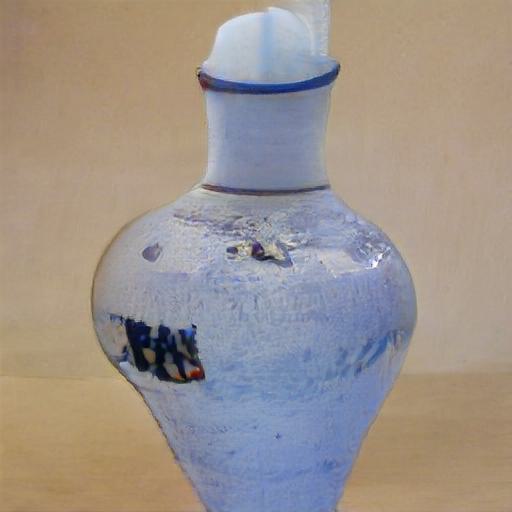} &
    \includegraphics[width=0.133\linewidth]{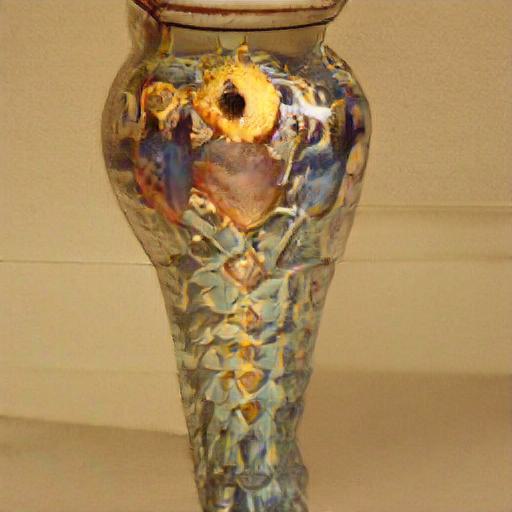} &
    \includegraphics[width=0.133\linewidth]{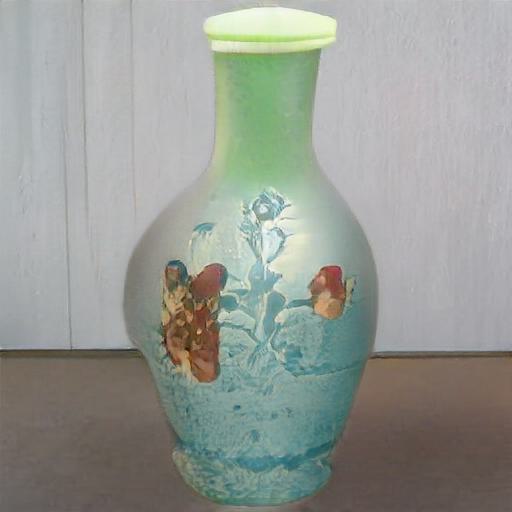} &
    \includegraphics[width=0.133\linewidth]{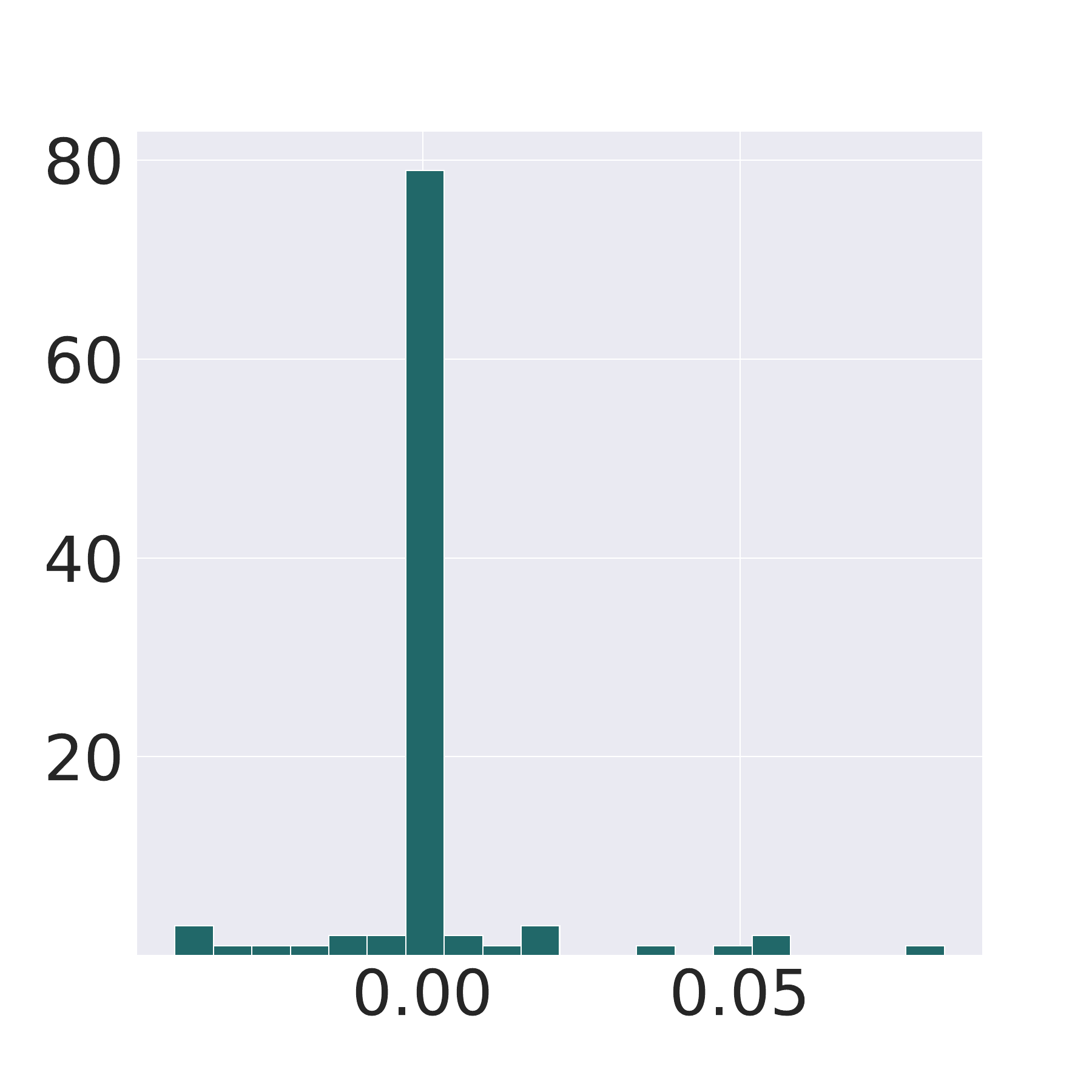} \\

    \end{tabular}
 
\caption{\textbf{Calibrated samples of class “Kite” (bird, 21), “Vase” (883).}}
\label{fig:fig_imagenet_w_samples}
\end{figure}

\begin{figure}[H]
    \footnotesize
    \centering
    \setlength{\tabcolsep}{2pt}
    \begin{tabular}{ c  c  c  c  c  c  c}
    \multicolumn{2}{c}{Highest positive $w$} & \multicolumn{2}{c}{Lowest negative $w$} & \multicolumn{2}{c}{$w=0.0$} & $w$ Distribution \\
    \midrule
    $w=0.055$ & $w=0.015$ & \multicolumn{2}{c}{$w=-0.011$} & $w=0.0$ & $w=0.0$ & \\
    \rotatebox{90}{\hspace{2mm}\tiny{Class 755, DiT (cfg=4) \textcolor{white}{p}}}
    \includegraphics[width=0.133\linewidth]{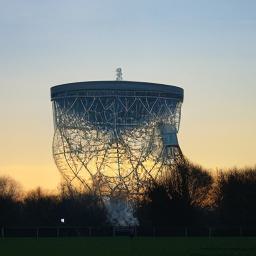} &
    \includegraphics[width=0.133\linewidth]{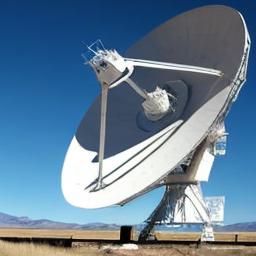} &
    \multicolumn{2}{c}{\includegraphics[width=0.133\linewidth]{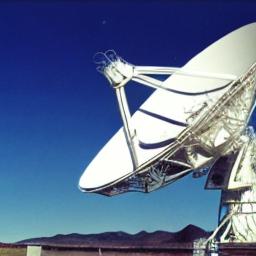}} &
    \includegraphics[width=0.133\linewidth]{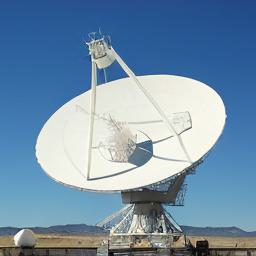} &
    \includegraphics[width=0.133\linewidth]{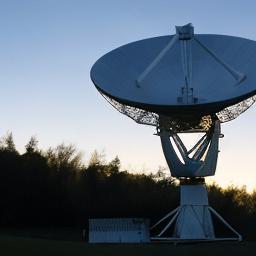} &
    \includegraphics[width=0.133\linewidth]{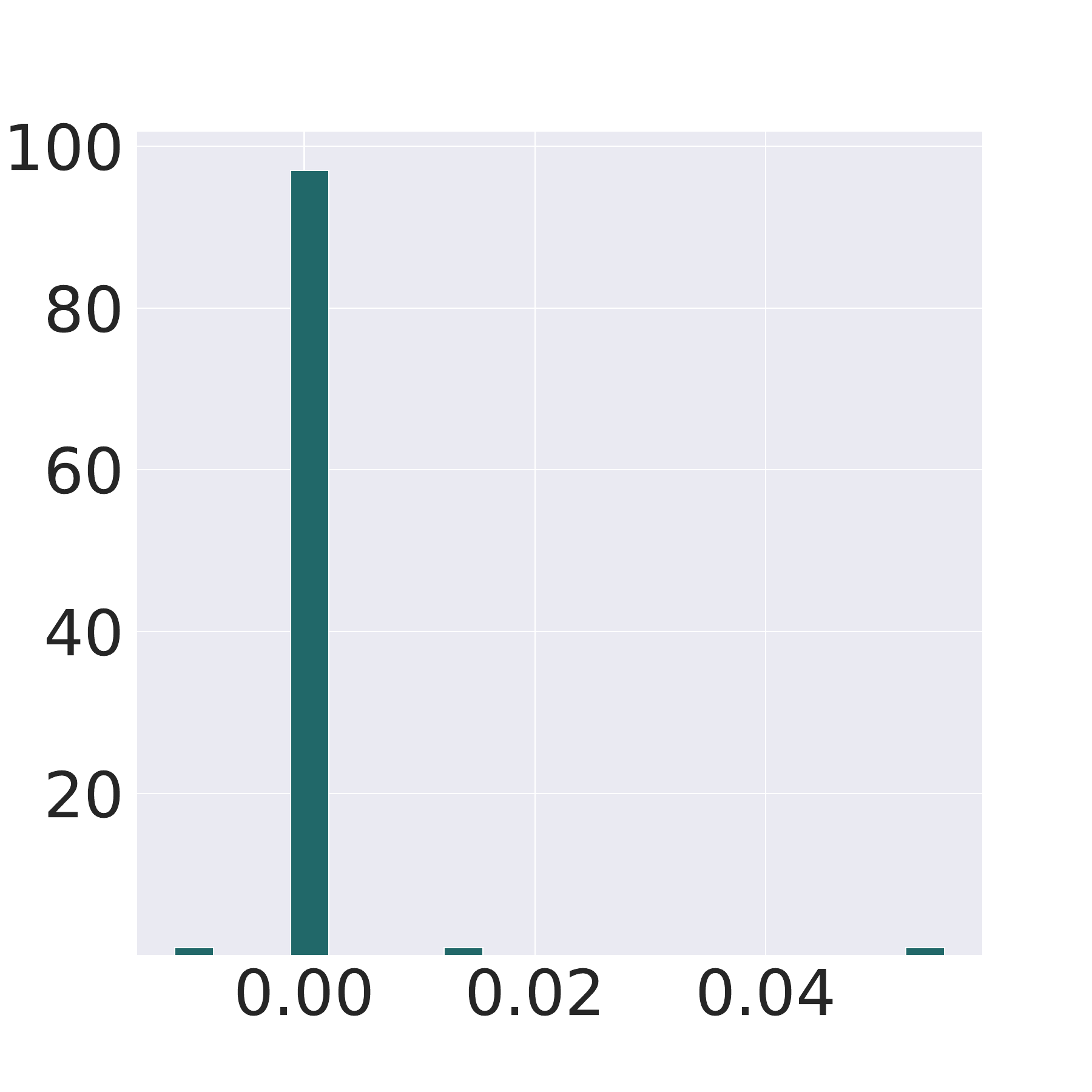} \\
    
    $w=0.016$ & $w=0.015$ & $w=-0.012$ & $w=-0.011$ & $w=0.0$ & $w=0.0$ & \\
    \rotatebox{90}{\hspace{2mm}\tiny{Class 755, DiT (cfg=1) \textcolor{white}{p}}}
    \includegraphics[width=0.133\linewidth]{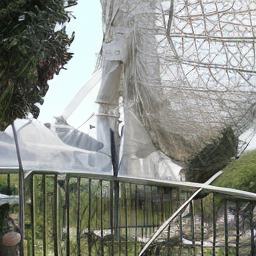} &
    \includegraphics[width=0.133\linewidth]{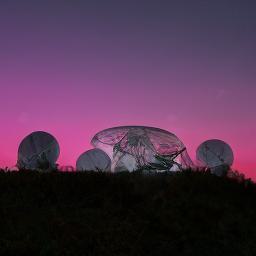} &
    \includegraphics[width=0.133\linewidth]{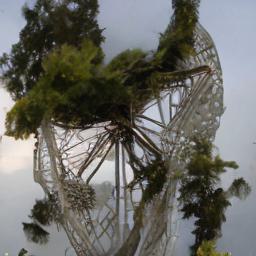} &
    \includegraphics[width=0.133\linewidth]{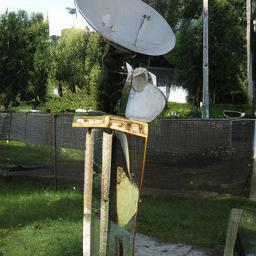} &
    \includegraphics[width=0.133\linewidth]{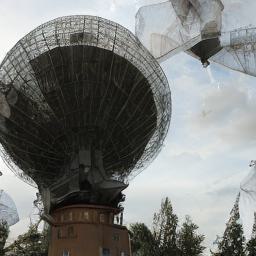} &
    \includegraphics[width=0.133\linewidth]{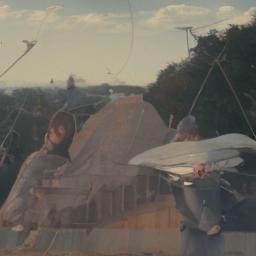} &
    \includegraphics[width=0.133\linewidth]{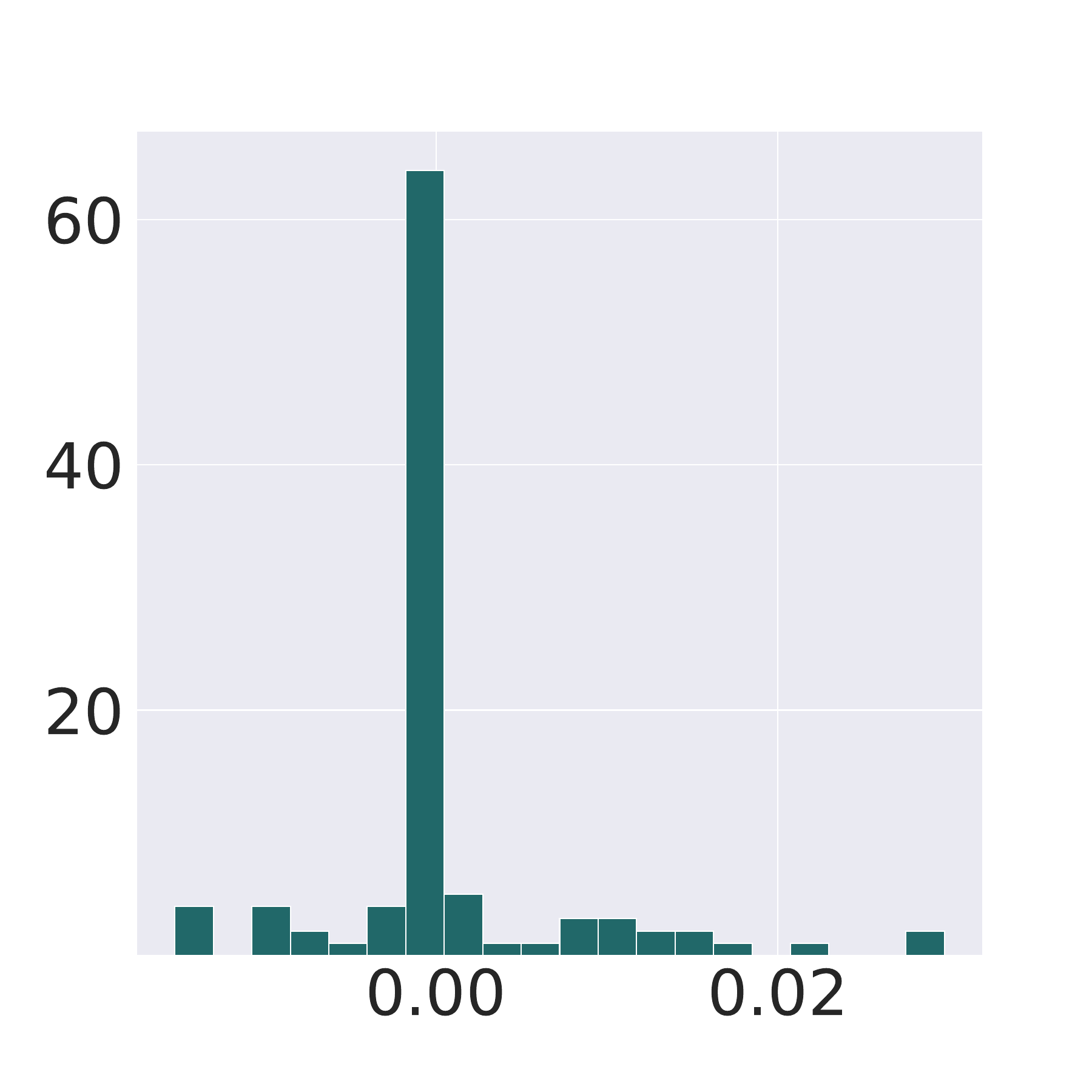} \\
    
    $w=0.040$ & $w=0.032$ & $w=-0.027$ & $w=-0.025$ & $w=0.0$ & $w=0.0$ & \\
    \rotatebox{90}{\hspace{2mm}\tiny{Class 755, BigGAN \textcolor{white}{p}}}
    \includegraphics[width=0.133\linewidth]{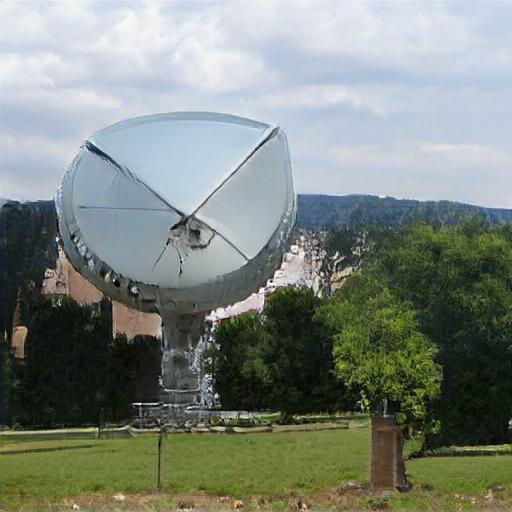} &
    \includegraphics[width=0.133\linewidth]{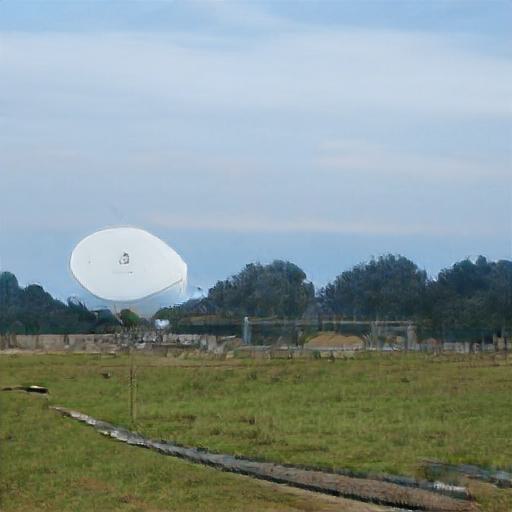} &
    \includegraphics[width=0.133\linewidth]{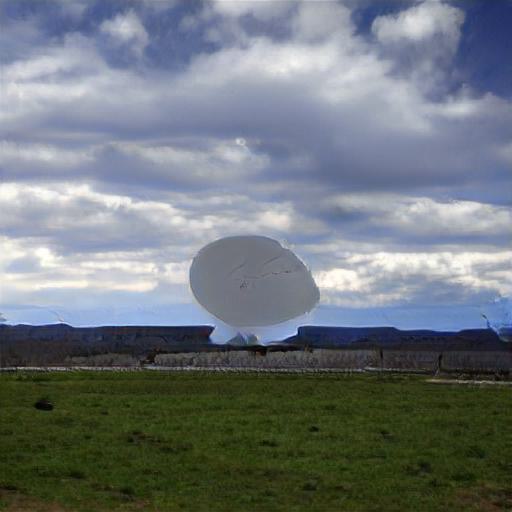} &
    \includegraphics[width=0.133\linewidth]{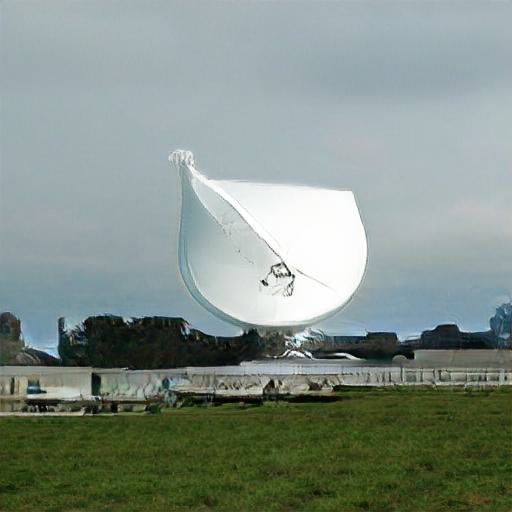} &
    \includegraphics[width=0.133\linewidth]{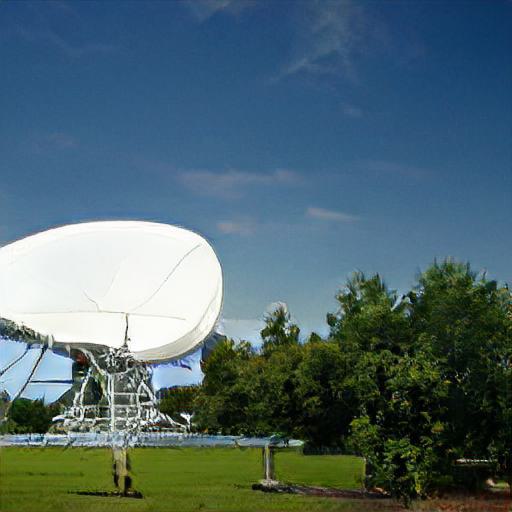} &
    \includegraphics[width=0.133\linewidth]{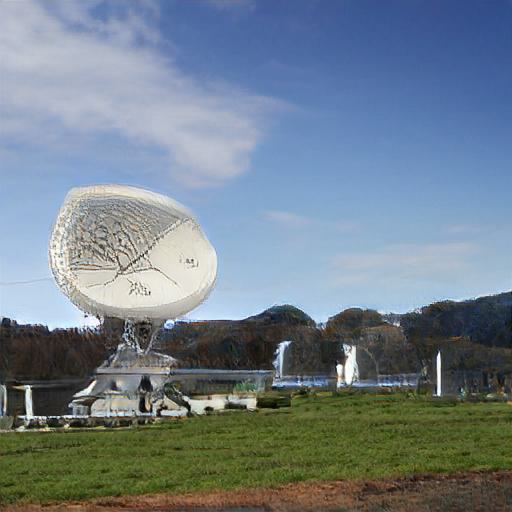} &
    \includegraphics[width=0.133\linewidth]{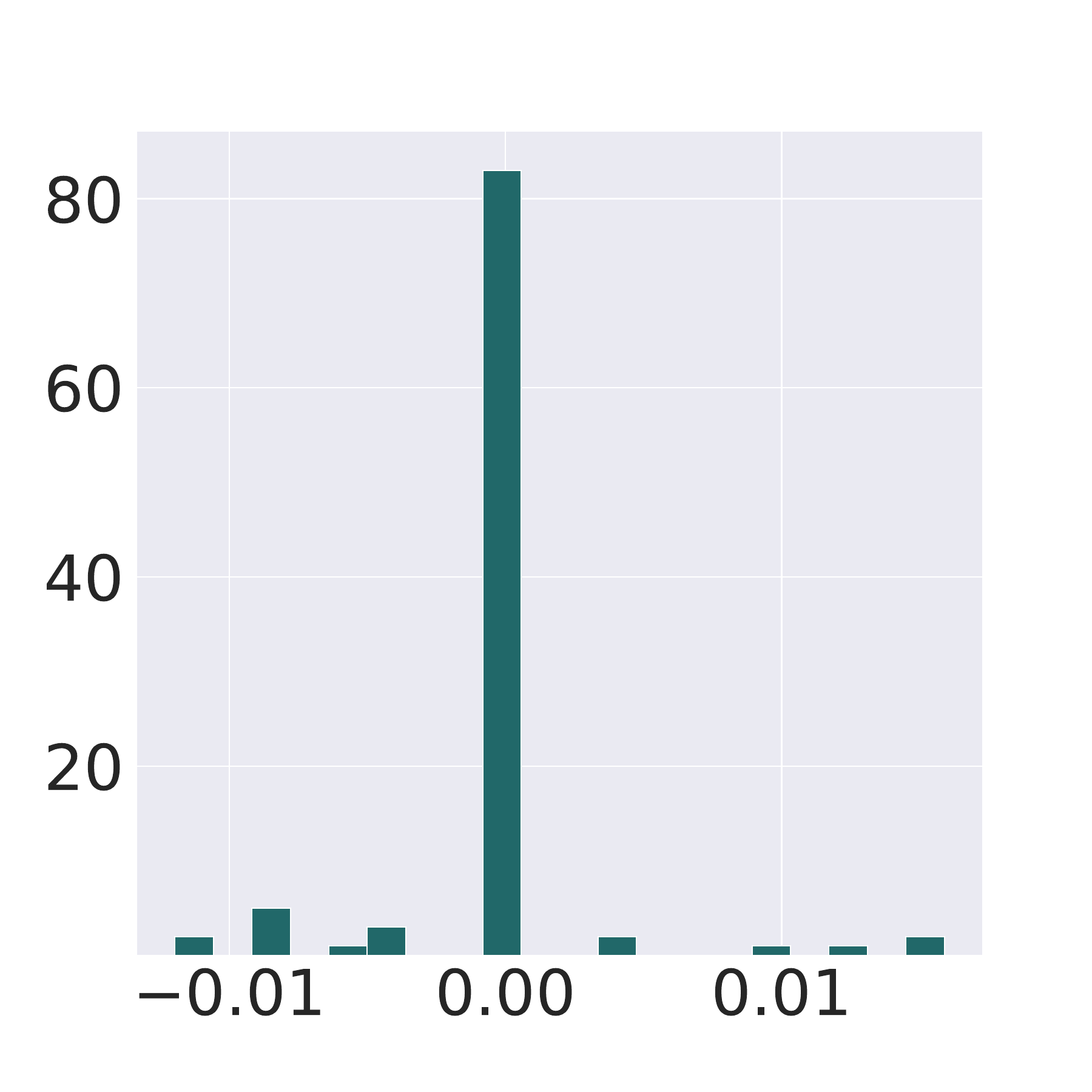} \\
    
    $w=0.052$ & $w=0.030$ & $w=-0.021$ & $w=-0.014$ & $w=0.0$ & $w=0.0$ & \\
    \rotatebox{90}{\hspace{2mm}\tiny{Class 2, DiT (cfg=4) \textcolor{white}{p}}}
    \includegraphics[width=0.133\linewidth]{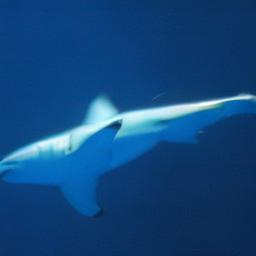} &
    \includegraphics[width=0.133\linewidth]{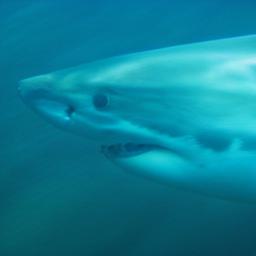} &
    \includegraphics[width=0.133\linewidth]{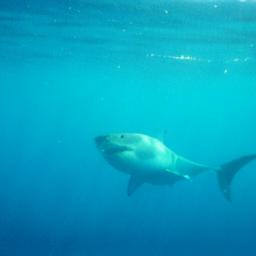} &
    \includegraphics[width=0.133\linewidth]{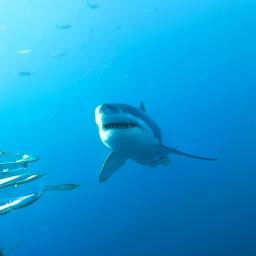} &
    \includegraphics[width=0.133\linewidth]{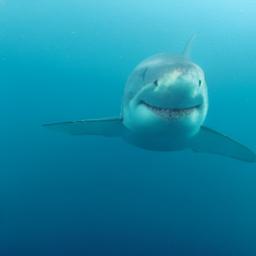} &
    \includegraphics[width=0.133\linewidth]{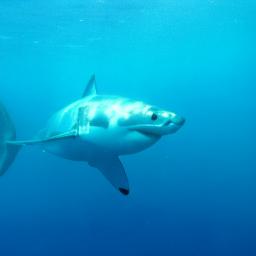} &
    \includegraphics[width=0.133\linewidth]{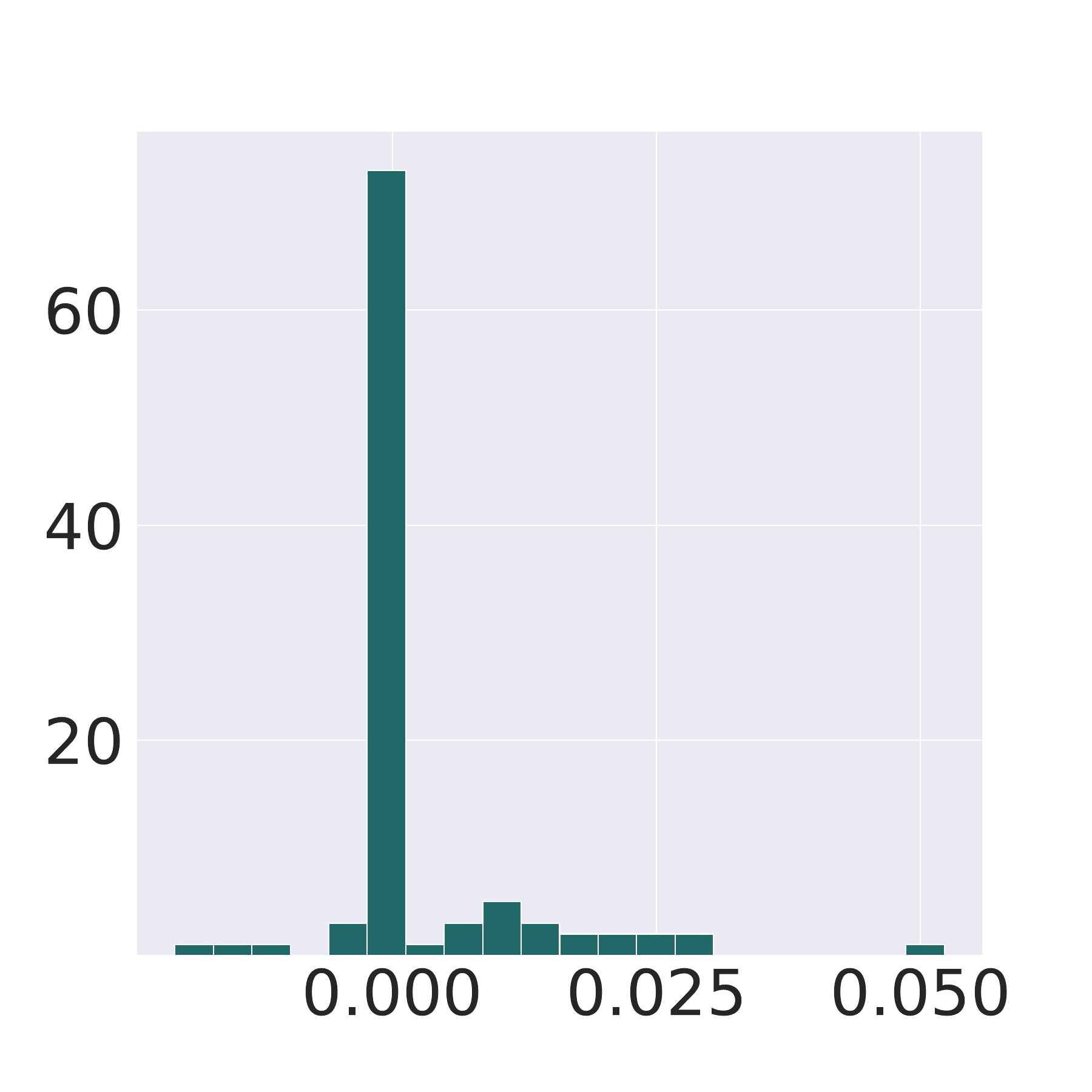} \\
    
    $w=0.039$ & $w=0.038$ & $w=-0.032$ & $w=-0.027$ & $w=0.0$ & $w=0.0$ & \\
    \rotatebox{90}{\hspace{2mm}\tiny{Class 2, DiT (cfg=1) \textcolor{white}{p}}}
    \includegraphics[width=0.133\linewidth]{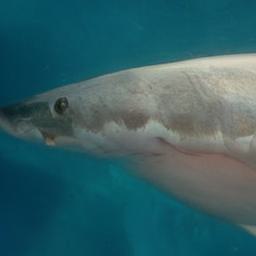} &
    \includegraphics[width=0.133\linewidth]{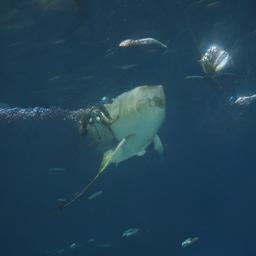} &
    \includegraphics[width=0.133\linewidth]{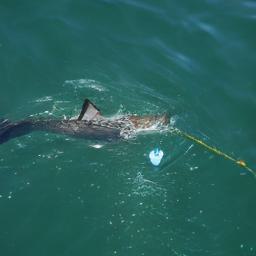} &
    \includegraphics[width=0.133\linewidth]{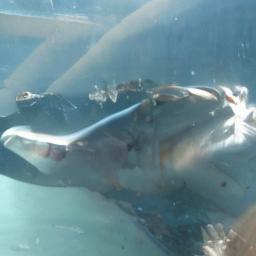} &
    \includegraphics[width=0.133\linewidth]{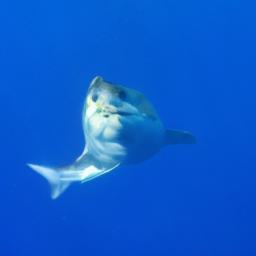} &
    \includegraphics[width=0.133\linewidth]{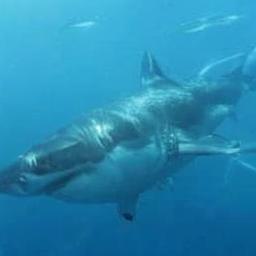} &
    \includegraphics[width=0.133\linewidth]{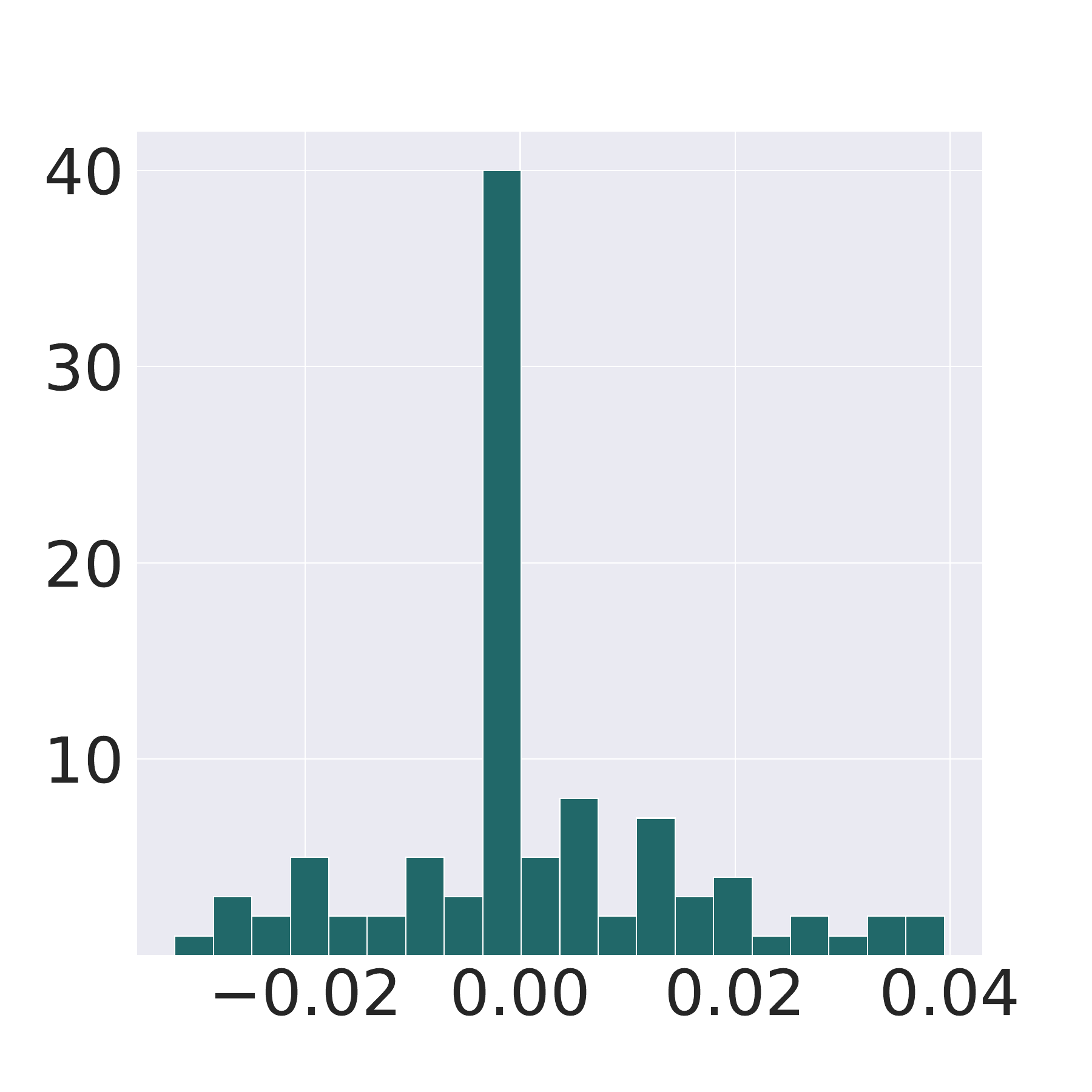} \\
    
    $w=0.050$ & $w=0.043$ & $w=-0.045$ & $w=-0.040$ & $w=0.0$ & $w=0.0$ & \\
    \rotatebox{90}{\hspace{2mm}\tiny{Class 2, BigGAN \textcolor{white}{p}}}
    \includegraphics[width=0.133\linewidth]{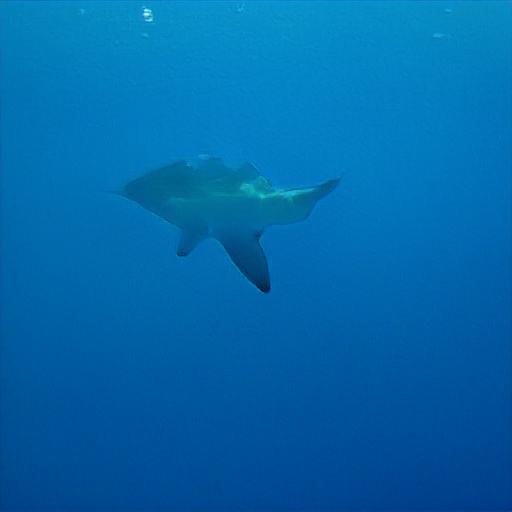} &
    \includegraphics[width=0.133\linewidth]{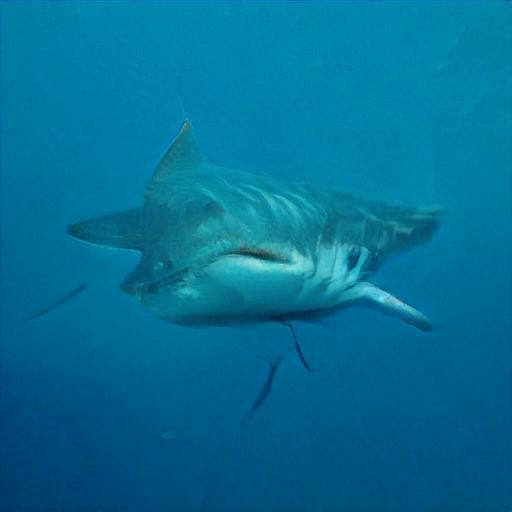} &
    \includegraphics[width=0.133\linewidth]{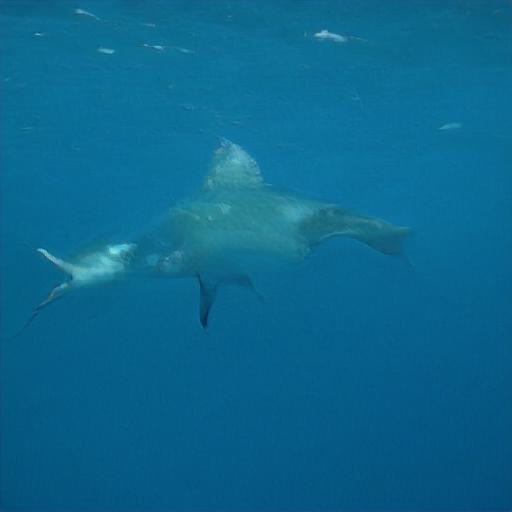} &
    \includegraphics[width=0.133\linewidth]{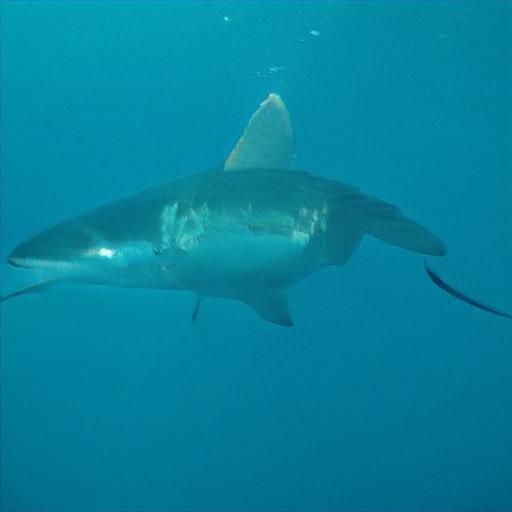} &
    \includegraphics[width=0.133\linewidth]{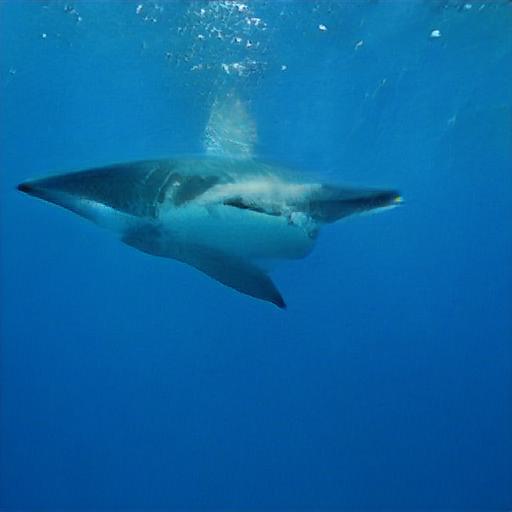} &
    \includegraphics[width=0.133\linewidth]{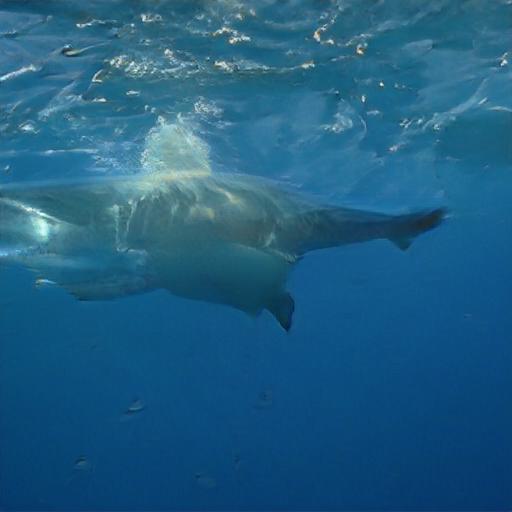} &
    \includegraphics[width=0.133\linewidth]{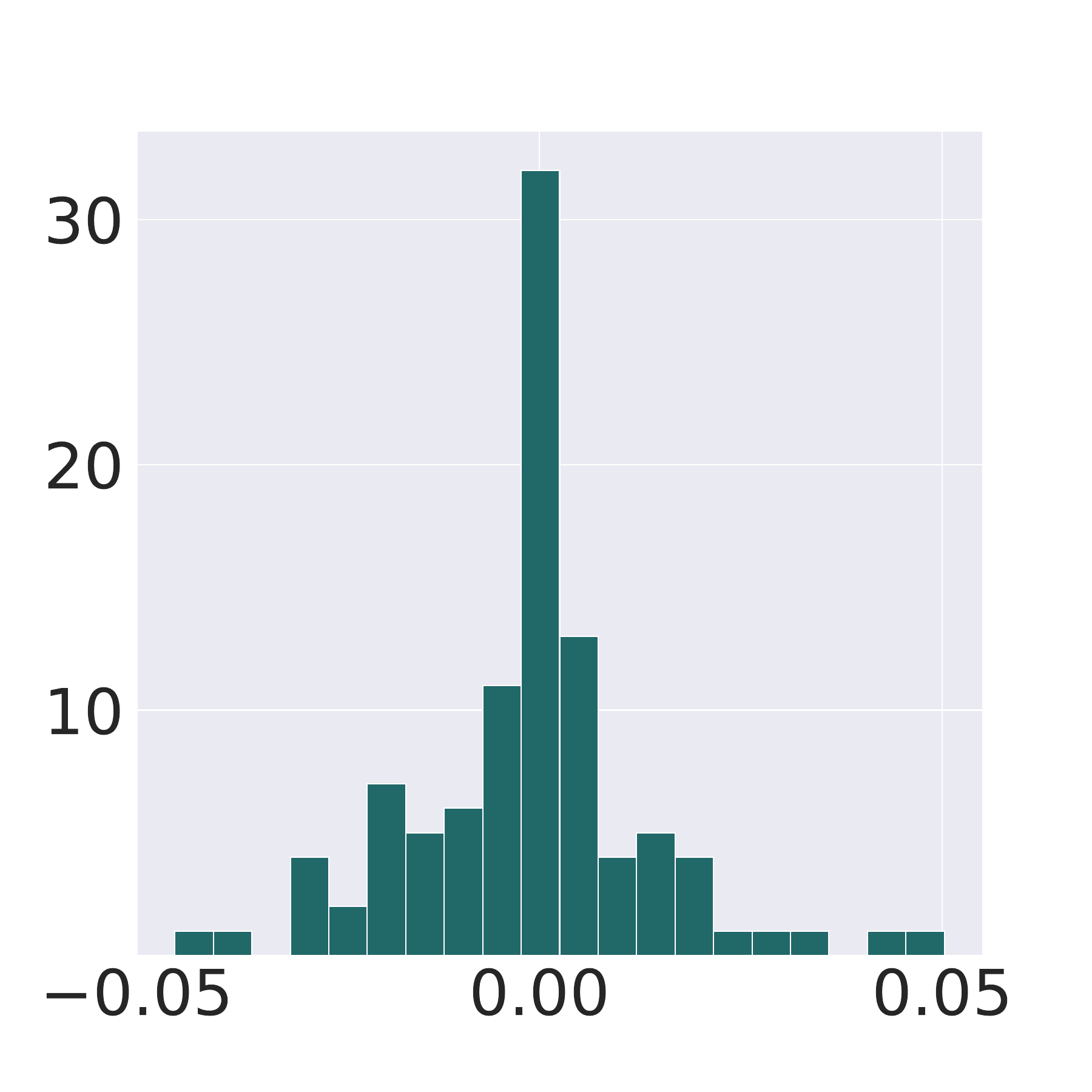} \\

    \end{tabular}
 
\caption{\textbf{Calibrated samples of classes “Radio telescope” (755), “White shark” (2).} For DiT (cfg $=4$) only one image is presented for the “Radio telescope” class as only it has received a negative $w$.
}
\label{fig:fig_imagenet_w_samples_2}
\end{figure}

In this section we analyze the contribution of images from calibrated synthetic datasets to model ranking.
In Figures~\ref{fig:fig_imagenet_w_samples} and~\ref{fig:fig_imagenet_w_samples_2} we show the images, $x_i$, that received the highest and lowest coefficients, $\mathbf{w}_c[i]$, in their corresponding classes, during the calibration process.
Each row shows images from a different synthetic dataset.
The first four columns show the images that received the highest positive and lowest negative $w$ coefficient values during calibration.
Columns five and six show samples that are not useful for ranking and received $w=0$.
The last column shows the distribution of $w$ for all 100 images in the class.
For sake of presentation clarity we use $w$ as a shorthand notation for $\mathbf{w}_c[i]$ next to an image to denote the value it received in its matching vector $\mathbf{w}_c$.
Images with positive $w$ coefficients increase the likelihood of models labeling them correctly to be ranked higher. 
In contrast, images with negative values of $w$ indicate that calibration models labeling them ``incorectly'' in general perform well on that class.
The improvement in ranking of unseen models indicates that in general images with non-zero coefficient are valuable for ranking.
Images with $w=0$ are those which were unanimously classified either correctly or incorrectly by all the models in the calibration set.

\end{document}